\documentclass[10pt,twocolumn,letterpaper]{article}

\usepackage{textcomp}
\usepackage{cvpr}              % To produce the CAMERA-READY version
\usepackage{bm}
\usepackage{epsfig}
\usepackage{siunitx}
\usepackage{gensymb}
\usepackage{mathtools}
\usepackage{caption}
\usepackage{subcaption}
\usepackage{comment}
\usepackage[dvipsnames]{xcolor}
\usepackage{colortbl}
\usepackage{subcaption}
\usepackage{physics}
\usepackage{accents}
\usepackage[outline]{contour}
\usepackage{bbding} % For checkmark

\renewcommand{\paragraph}[1]{\vspace{1ex}\noindent\textbf{#1}\quad}
\newcommand{\mae}[1]{\textcolor{white}{\contour{black}{{\tiny MAE:} #1}}}

\newcommand{\redstart}{\color{black}}

\newcommand{\albedo}{\ensuremath{\rho}\xspace}
\newcommand{\sourcepower}{\ensuremath{s}\xspace}
\newcommand{\normal}{\ensuremath{\bm{n}}\xspace}

\renewcommand{\time}{\ensuremath{t}\xspace}
\newcommand{\period}{\ensuremath{T}\xspace}
\newcommand{\timelastevent}{\ensuremath{\tau_{\mbox{\scriptsize $-$}}(\time)}\xspace}
\newcommand{\timenextevent}{\ensuremath{\tau_{\mbox{\scriptsize $+$}}(\time)}\xspace}
\newcommand{\lightdirection}[1][\time]{\ensuremath{\bm{l}({#1})}\xspace}
\newcommand{\derivlightdirection}[1][\time]{\ensuremath{\bm{l}'({#1})}\xspace}
\newcommand{\fixedlightdirection}{\ensuremath{\bm{l}}\xspace}
\newcommand{\radiance}[1][\time]{\ensuremath{L({#1})}\xspace}
\newcommand{\qeff}{\ensuremath{q}\xspace}
\newcommand{\ampsig}[1][\time]{\ensuremath{s_p({#1})}\xspace}
\newcommand{\diffsig}{\ensuremath{s_d(\time)}\xspace}
\newcommand{\compsig}{\ensuremath{s_c(\time)}\xspace}
\newcommand{\thresh}{\ensuremath{h}\xspace}
\newcommand{\anythresh}{\ensuremath{\thresh_t}\xspace}
\newcommand{\posthresh}{\ensuremath{\thresh_p}\xspace}
\newcommand{\negthresh}{\ensuremath{\thresh_n}\xspace}
\newcommand{\profile}[1][\time|\normal, \lightdirection]{\ensuremath{p({#1})}\xspace}

\newcommand{\offsetlight}{\ensuremath{L_\varepsilon}\xspace}
\newcommand{\numposevents}{\ensuremath{N_p}\xspace}
\newcommand{\numobs}{\ensuremath{K}\xspace}
\newcommand{\heatmap}{\ensuremath{\Delta I}\xspace}
\newcommand{\costthresh}{\ensuremath{c}\xspace}

\newcommand{\nummasks}{\ensuremath{M}\xspace}
\newcommand{\maskattached}[1][\time|\normal]{\ensuremath{m_a(#1)}\xspace}
\newcommand{\maskgeneral}[1][\time]{\ensuremath{m_e(#1)}\xspace}
\newcommand{\maskglossiness}[1][\time]{\ensuremath{m_s(#1)}\xspace}
\newcommand{\marginglossiness}{\ensuremath{\Delta t_s}\xspace}
\newcommand{\maskcast}[1][\time]{\ensuremath{m_c(#1)}\xspace}
\newcommand{\margincast}{\ensuremath{\Delta t_c}\xspace}
\newcommand{\timetoppeak}{\ensuremath{t_t}\xspace}
\newcommand{\timebottompeak}{\ensuremath{t_b}\xspace}
\newcommand{\durationtopbottom}{\ensuremath{t_{tb}}\xspace}

\newcommand{\mat}[1]{\textsl{#1}\xspace}
\newcommand{\diffusesphere}{\mat{Diffuse}}
\newcommand{\glossysphere}{\mat{Glossy}}
\newcommand{\shadowsphere}{\mat{Pole}}
\newcommand{\twocolorsphere}{\mat{Colors}}
\newcommand{\multirefsphere}{\mat{MulRefs}}
\newcommand{\bunny}{\mat{Bunny}}
\newcommand{\caesar}{\mat{Caesar}}
\newcommand{\baboon}{\mat{Baboon}}
\newcommand{\cube}{\mat{Cube}}
\newcommand{\cv}{\mat{CV}}
\newcommand{\brick}{\mat{Brick}}
\newcommand{\floral}{\mat{Floral}}
\newcommand{\twocolorglossysphere}{\mat{Glossy2C}}
\newcommand{\diffuseyr}{\mat{DiffuseYR}}
\newcommand{\diffusebg}{\mat{DiffuseBG}}
\newcommand{\blacksphere}{\mat{Black}}
\newcommand{\semiglossy}{\mat{SemiGlossy}}
\newcommand{\twopoles}{\mat{2Poles}}
\newcommand{\glossycv}{\mat{GlossyCV}}
\newcommand{\multirefsbunny}{\mat{RefsBunny}}
\newcommand{\bear}{\mat{Bear}}

\newcommand{\cactusA}{\mat{CactusA}}
\newcommand{\cactusB}{\mat{CactusB}}
\newcommand{\cactusC}{\mat{CactusC}}
\newcommand{\elmo}{\mat{Elmo}}
\newcommand{\medalcat}{\mat{MedalCat}}
\newcommand{\medalfrog}{\mat{MedalFrog}}
\newcommand{\miffy}{\mat{Miffy}}
\newcommand{\owl}{\mat{Owl}}
\newcommand{\panda}{\mat{Panda}}
\newcommand{\retriever}{\mat{Retriever}}
\newcommand{\rody}{\mat{Rody}}
\newcommand{\soccerfrog}{\mat{SoccerFrog}}
\newcommand{\tenniscat}{\mat{TennisCat}}
\newcommand{\whitedog}{\mat{WDogA}}
\newcommand{\satwhitedog}{\mat{WDogB}}

\newcommand{\um}{\micro\metre}
\newcommand{\mm}{\milli\metre}

\definecolor{cvprblue}{rgb}{0.21,0.49,0.74}
\usepackage[pagebackref,breaklinks,colorlinks,citecolor=cvprblue]{hyperref}

\def\paperID{1981} % *** Enter the Paper ID here
\def\confName{CVPR}
\def\confYear{2025}

\title{PS-EIP: Robust Photometric Stereo Based on Event Interval Profile\footnotemark[1]}
\author{Kazuma Kitazawa$^{1}$ \quad Takahito Aoto$^{2}$ \quad Satoshi Ikehata$^{3}$ \quad Tsuyoshi Takatani$^{1}$\\
\vspace{-0.4cm}\\
{ $^{1}$University of Tsukuba} \quad
{ $^{2}$Optech Innovation, LLC.} \quad
{ $^{3}$National Institute of Informatics (NII)}\\
{\tt\small kitazawa.kazuma.qy@alumni.tsukuba.ac.jp} \quad
{\tt\small aoto@optechinnovation.com} \quad \\
{\tt\small sikehata@nii.ac.jp} \quad
{\tt\small takatani@iit.tsukuba.ac.jp} \\
\vspace{-4ex}
}

\begin{document}
\maketitle

\renewcommand{\thefootnote}{\fnsymbol{footnote}}
\footnotetext[1]{This work was supported by JSPS KAKENHI Grant Numbers JP23K20382 and JP24K02966.}
\renewcommand{\thefootnote}{\arabic{footnote}}
\setcounter{footnote}{0}

\begin{abstract}
Recently, the energy-efficient photometric stereo method using an event camera (EventPS~\cite{EventPS}) has been proposed to recover surface normals from events triggered by changes in logarithmic Lambertian reflections under a moving directional light source. However, EventPS treats each event interval independently, making it sensitive to noise, shadows, and non-Lambertian reflections. This paper proposes Photometric Stereo based on Event Interval Profile (PS-EIP), a robust method that recovers pixelwise surface normals from a time-series profile of event intervals. By exploiting the continuity of the profile and introducing an outlier detection method based on profile shape, our approach enhances robustness against outliers from shadows and specular reflections. Experiments using real event data from 3D-printed objects demonstrate that PS-EIP significantly improves robustness to outliers compared to EventPS's deep-learning variant, EventPS-FCN, without relying on deep learning.

\end{abstract}

\section{Introduction}
\label{sec:intro}

Photometric stereo~\cite{Woodham1980,Shi2010,Shi2012a,Ikehata2014a,Ikehata2022,Ikehata2023,Ikehata2024} is a technique for reconstructing surface normals from images captured under varying lighting conditions. Despite its high fidelity, photometric stereo methods often suffer from significant information redundancy and substantial computational costs. Processing tens of high-resolution color images requires considerable computational resources and often depends on large-scale deep learning models accelerated by GPUs for extracting high-level image features.

Recently, event cameras have emerged as a promising alternative for energy-efficient computer vision applications. Unlike conventional RGB cameras, event cameras independently and asynchronously detect changes in intensity at each pixel, recording the timestamp of each change as an event. Compared with conventional cameras, event cameras provide superior features such as high temporal resolution, high dynamic range (HDR), and low power consumption.

Building upon these advantages, the Lambertian event-based photometric stereo, namely EventPS~\cite{EventPS}, has recently been proposed. EventPS leverages the fact that, under a continuously moving light source, events are triggered by differences in logarithmic Lambertian reflections. In this context, the surface normal can be uniquely determined as the null space of the logarithmic intensity ratio scaled by the event signal threshold. However, this method assumes that events are triggered at all adjacent event occurrence times. Consequently, reconstruction fails when events occur at different timings due to observation noise, shadows, or outliers resulting from non-Lambertian reflection components. Furthermore, despite assuming the continuity of the moving light source, it does not consider the relationships between event intervals. This omission makes it more challenging to remove outliers caused by non-Lambertian effects.
\begin{figure}[t]
    \centering
    \includegraphics[width=\linewidth]{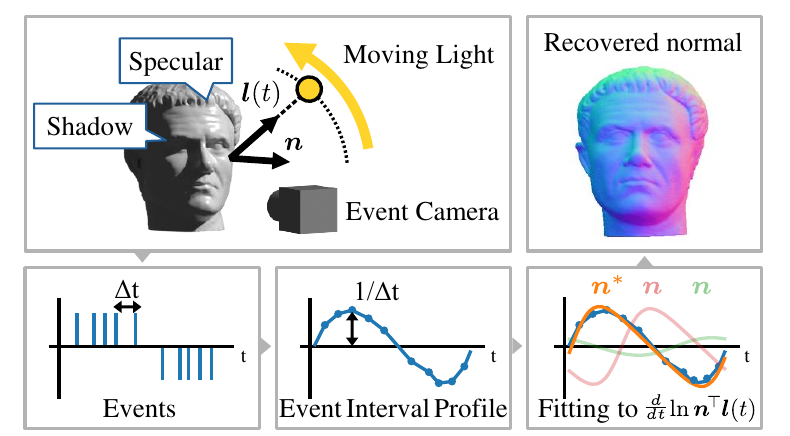 }
    \caption{Procedure of the proposed method: Events are recorded under moving light conditions. The profile is reconstructed from the inverse of the event intervals. A surface normal is determined through curve fitting for each pixel. The proposed method is robust to non-Lambertian effects such as specularity and shadows.}
    \label{fig:teaser}
\end{figure}

In this paper, we propose a robust method for Lambertian event-based photometric stereo, namely {\it PS-EIP} (Photometric Stereo based on Event Interval Profile) as illustrated in~\cref{fig:teaser}. Unlike EventPS, which treats each event interval independently, our approach considers a time-series profile formed by event intervals, taking into account the relationships between adjacent intervals. This profile, consisting of the inverses of event intervals, can be regarded as the time derivative of logarithmic intensity. Since the shape of this profile is uniquely determined by the surface normal at each pixel under known movement of the directional light source of constant power, we estimate the surface normals by minimizing the distance between this theoretical profile and the profile obtained from actual observations. By considering the continuity of the profile, our method becomes more robust against outliers; however, it can still be affected by factors such as shadows and specular reflections. To address this issue, we propose a method that identifies outliers based on the shape of the profile and excludes them from the profile fitting process. 

We conducted numerous quantitative and qualitative evaluations of the proposed method using real event data obtained from 3D printed objects whose shapes are known a priori. Surprisingly, despite not employing resource-consuming deep learning and estimating normals entirely at the pixel level, our method is significantly more robust to outliers than a deep-learning variant of EventPS~\cite{EventPS} (\ie, EventPS-FCN). 

The major contributions are summarized as follows:
\begin{itemize} 
\item We propose a novel method called PS-EIP for robust Lambertian event-based photometric stereo. Unlike the existing EventPS~\cite{EventPS}, which treats each event interval independently, PS-EIP considers the time-series profile formed by event intervals, taking into account the relationships between adjacent intervals. 
\item We introduce a rule-based approach to identify and exclude outliers based on the shape of the event interval profile. This further improves the accuracy of surface normal estimation by reducing the influence of non-Lambertian reflections and observation noise. 
\item We develop a prototype system with off-the-shelf components and calibration techniques to validate the proposed method and conducted a quantitative evaluation using $21$ 3D printed objects with various shapes, achieving a mean angular error of \SI{8.12}{\degree} in the presence of non-Lambertian effects. 
\end{itemize}

\section{Related Work}
\label{sec:related work}

\subsection{Photometric Stereo}
Since Woodham~\cite{Woodham1980} proposed the Lambertian photometric stereo method in 1980s, numerous methods have been developed to address more complex phenomena. A simple yet effective approach involves removing or ignoring non-Lambertian observations as outliers~\cite{miyazaki2010median,chandraker2007shadowcuts,takatani2013enhanced,ikehata2012,Mukaigawa2007,Wu2010}, while some works explicitly modeled non-Lambertian reflectance by introducing non-Lambertian BRDFs (Bidirectional Reflectance Distribution Functions)~\cite{Goldman2005,Ikehata2014b,Shi2012a,chen2019microfacet}. As an alternative approach, example-based methods use observations of known objects recorded under identical conditions to the target scene~\cite{Silver1980,Hertzmann2005,Hui2017}. 

Recently, learning-based methods~\cite{Santo2017, Ikehata2018, Li2019, Zheng2019, Logothetis2021, Chen2018, Yakun2021, Yao2020, Liu2021, Ikehata2021, Chen2019, Chen2020, Kaya2021, Tiwari2022, Ikehata2022, Taniai2018, Li2022a, Li2022b} have emerged to address challenges faced by traditional physics-based approaches. Notably, the introduction of universal photometric stereo methods~\cite{Ikehata2022, Ikehata2023, Ikehata2024} has enabled the handling of unknown, spatially varying lighting in a purely data-driven framework. It is also worth considering the use of different sensors in photometric stereo, such as thermal cameras~\cite{tanaka2019time}, time-of-flight sensors~\cite{ti2015simultaneous}, and polarimetric sensors~\cite{ATKINSON2017158}, to enhance reconstruction quality.

\subsection{Event Camera in Computer Vision}
Event cameras detect intensity changes asynchronously and independently at each pixel. They are widely used for tasks like tracking moving objects~\cite{Cohen2017,Gehrig2018a,Mitrokhin2018}, visual odometry~\cite{zhu2017,Bryner2019}, 3D scene reconstruction~\cite{Schraml2015,Andreopoulos2018,Zhu2018a}, and simultaneous localization and mapping (SLAM)~\cite{kim2016}. Since event cameras do not record the radiance directly, reconstructing relative intensity images~\cite{Bardow2016,Munda2018} and videos~\cite{Pan2019,Rebecq2019} from event streams is essential for applying them to traditional computer vision tasks like object classification~\cite{Sironi2018}. This reconstruction leverages the unique features of event cameras, offering benefits such as high dynamic range (HDR) and high frame rates~\cite{Wang2019a}.

Recent work has increasingly employed event cameras for active sensing applications. Combining an event camera with a scanning laser sheet enables 3D shape reconstruction of highly dynamic objects under strong ambient illumination~\cite{Matsuda2015}. Temporally modulated bispectral illumination allows direct readout of differential absorption between two wavelengths from event streams~\cite{takatani2021event}. Projector-camera systems with event cameras have been used to capture 3D scenes~\cite{muglikar2021esl}, and exponentially modulated illumination leads to accurate radiance recovery~\cite{Han2023CVPR}. Passive encoding methods, such as shape reconstruction using a rotating polarizer, have also been combined with event cameras~\cite{Muglikar2023CVPR}. As has been discussed, EventPS~\cite{EventPS} has firstly proposed event-based photometric stereo by leveraging event intervals of Lambertian observations under the continuous directional light source. We extend this work by employing the time-series profile formed by event intervals for improving the robustness of the reconstruction.

\section{Preliminaries}
\label{sec:principle}

\subsection{Events Under a Moving Directional Light}
Assuming that a single directional light (\ie, a distant point light source) illuminates an object, and the light moves temporally along a known trajectory on a unit sphere, the radiance at a surface point at a specific time is expressed as
\begin{equation}
    \radiance = \sourcepower \albedo \normal^\top \lightdirection,
    \label{eq:radiance}
\end{equation}
where \sourcepower is the constant power of the light source, \albedo and \normal are the albedo and normal direction of the surface point, and \lightdirection is the direction of light at a time \time, respectively. Here, the reflectance of the object is assumed to be Lambertian and attached shadows are ignored, but the both are revisited later.
When the light moves temporally, changes in radiance trigger events in an event camera, which consists of three circuits: a photoreceptor, a differentiator, and comparators~\cite{Lichtsteiner2008, posch2011}. The photoreceptor converts the radiance into an electricity signal and then logarithmically amplifies it as follows
\begin{equation}
    \ampsig = \ln{\qeff \radiance},
    \label{eq:ampsig}
\end{equation}
where \qeff is the quantum efficiency of the photoreceptor. Here, we assume that the light-to-electricity conversion is linear and the logarithmic amplifier can be mathematically modeled as a logarithmic function. In practice, it is known that the characteristics of logarithmic amplifiers do not follow a logarithmic function in the low input range~\cite{zumbahlen2007basic}. However, EventPS~\cite{EventPS} did not account for these artifacts, which causes problems for the low-intensity observations (\eg, around attached shadows). To address this, we propose adding an offset light alongside a distant point light source, as will be explained in \cref{sec:implementation}.

The differentiator calculates a temporal difference of the amplified signal based on a time, \timelastevent, when the previous event is triggered before the current time. The output from the differentiator can be written as below;
\begin{align}
    \diffsig &= \ampsig - \ampsig[\timelastevent] \\
    &= \ln\left({\frac{\qeff \radiance}{\qeff \radiance[\timelastevent]}}\right) = \ln\left({\frac{\sourcepower \albedo \normal^\top \lightdirection}{\sourcepower \albedo \normal^\top \lightdirection[\timelastevent]}}\right) \\
    &= \ln\left({\frac{\normal^\top \lightdirection}{\normal^\top \lightdirection[\timelastevent]}}\right).
\end{align}
As have been discussed in~\cite{EventPS}, the quantum efficiency \qeff, the power of the light source \sourcepower, and the albedo \albedo are canceled, and thus \diffsig does not depend on those.

The final circuit consists of two comparators to trigger events with two polarities; positive and negative. Each of the comparators generally has an individual threshold and triggers an event when \diffsig goes over the threshold. This output can be explained as below;
\begin{equation}
    \compsig = 
    \begin{cases}
        +1  & \text{if $\diffsig \geq \posthresh$,} \\
        0   & \text{if $\negthresh < \diffsig < \posthresh$,} \\
        -1  & \text{if $\diffsig \leq \negthresh$,} \\
    \end{cases}
    \label{eq:compsig}
\end{equation}
where \posthresh and \negthresh are the thresholds for positive and negative events, respectively. Again, the triggered events are unaffected by the albedo of the surface point because \diffsig is independent to that. 
More importantly, an event is recorded with a single timestamp; however, since the temporal resolution is quite high, the interval from one event to another can be considered as continuous information.
\subsection{Revisiting EventPS~\cite{EventPS}}
Based on the Lambertian event generation model discussed in the last section, Yu~\etal~\cite{EventPS} proposed EventPS under the assumption that the movement of a single directional light is continuous and the event occurs only at $s_d(t)=h_p$ or $s_d(t)=h_n$ as 
\vspace{-0.2ex}
\begin{equation}
\ln\left( \frac{\normal^\top \lightdirection}{\normal^\top \lightdirection[\timelastevent]} \right) = \anythresh \in
\begin{cases}
        \posthresh  & \text{if $\diffsig = \posthresh$,} \\
        \negthresh  & \text{if $\diffsig = \negthresh$.} \\
    \end{cases}
\end{equation}
\vspace{-0.3ex}
Under this assumption, the surface normal is recovered by solving following the linear systems
\vspace{-0.2ex}
\begin{equation}
\normal^\top \left( \lightdirection - e^{\anythresh} \lightdirection[\timelastevent] \right) = 0.
\end{equation}

\vspace{-0.5ex}
Since this method is valid only in the absence of non-Lambertian effects, such as cast shadow and specularities, EventPS~\cite{EventPS} aimed to overcome this limitation by thresholding out event signals where the event intervals are shorter than a specified time threshold, as these corruptions abruptly change the radiance. 
% However, this n\"aive strategy is only valid where the observation is 
% ここまで直した
\redstart
However, this na\"ive strategy relies only on each of the event intervals, limiting detection to boundaries with non-Lambertian events while our method identifies outliers as a region.
\color{black}

\section{PS-EIP}
\subsection{Event Interval Profile (EIP)}
\begin{figure}[t]
    \centering
    \begin{subfigure}[t]{0.45\linewidth}
        \centering
        \includegraphics[width=\linewidth]{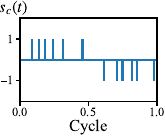}
        \caption{}
    \end{subfigure}
    \begin{subfigure}[t]{0.45\linewidth}
        \centering
        \includegraphics[width=\linewidth]{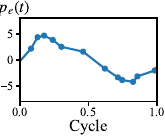}
        \caption{}
    \end{subfigure}
    % \begin{subfigure}[t]{0.325\linewidth}
    %     \centering
    %     \includegraphics[width=\linewidth]{figs/profgen/prof_resample.pdf}
    %     \caption{}
    % \end{subfigure}
    \caption{Reconstruction of a profile from a series of events of a single cycle light movement. (a)~Recorded events. (b)~Reconstructed event interval profile (EIP).} 
    % (c)~Linearly interpolated profile.
    \label{fig:profile reconstruction}
\end{figure}

% \aoto{ここでEvent intervalの話を持ってくる}
%A set of event information includes a pixel location, a timestamp, and a polarity. 
%To recover the surface normal, a temporal series of events with respect to the moving light source is used. 
%In this paper, we define the time derivative of the amplified signal as a temporal profile. 
We assume that the event interval is sufficiently small due to the small event threshold ($h_t$). Then, we here define the time derivative of the amplified signal as a time-series profile formed by event intervals, as follows
\vspace{-0.3ex}
\begin{equation}
    \profile = \dv{\ampsig}{\time} = \frac{\normal^\top \derivlightdirection}{\normal^\top \lightdirection},
    \label{eq:ideal profile}
\end{equation}
\vspace{-0.3ex}
where \derivlightdirection denotes the time derivative of \lightdirection and we call $\profile$ as {\it Event Interval Profile (EIP)} which is the function of time ($t$). Since we assume a small interval threshold, the profile is approximated using the central finite difference of event intervals from the recorded events, as shown below:
\vspace{-0.3ex}
\begin{equation}
p_e(\time|h_t) = \frac{2 \anythresh}{\timenextevent - \timelastevent}, \label{eq
reconstruction}
\end{equation}
\vspace{-0.3ex}
where \timenextevent is the time when the next event is triggered after \time. This represents the central finite difference in a region spanning three recorded event intervals. Similar to EventPS~\cite{EventPS}, we also assume that an event is ideally triggered when $\diffsig = \thresh_p$ or $\diffsig = \thresh_n$. The profile reconstruction is illustrated in~\cref{fig:profile reconstruction}. 
%Since the recorded events are temporally non-uniform, \obsprofile is resampled using linear interpolation

\subsection{Normal Recovery from EIP}

Given events and their triggered times, we first recover the event interval profile from them. Under the condition where the event thresholds and the trajectory of a single directional light source are known, we can calculate the ideal EIP as well. Then, we optimize the pixel-wise surface normal by minimizing residuals between reconstructed and ideal EIPs as:
\begin{equation}
    \min_{\normal}{\frac{1}{\nummasks} \sum_{\time=0}^{\period} \maskattached \left\| \profile[\time|\normal,\lightdirection] - p_e(\time|h_t) \right\|_{2}^{2}},
    \label{eq:cost}
\end{equation}
where \period is the duration of the trajectory of the light source, and \nummasks is the sum of \maskattached over the duration. \maskattached is an attached shadow mask function accounting for attached shadows, which is defined below:
\begin{equation}
    \maskattached =
    \begin{cases}
        1 & \text{if $\normal^\top \lightdirection > 0$}, \\
        0 & \text{otherwise}. \\
    \end{cases}
    \label{eq:mask attached}
\end{equation}

Note that we assume that the trajectory of the light source and all the surface points are not coplanar as was also assumed in~\cite{EventPS}. Otherwise, \profile is ambiguous with respect to \normal in \cref{eq:ideal profile}.

\subsection{Masks for Non-Lambertian Effects}
To improve our method's robustness to non-Lambertian observations, we propose removing temporal regions of the EIP distorted by non-Lambertian effects such as specular reflections and cast shadow. For instance, when a glossy surface point reflects the moving light source directly into a camera pixel, the pixel observes a specular highlight. This results in many more positive events in a shorter time compared to the Lambertian case. Additionally, when the light source moves away, many negative events occur. Similarly, when a surface enters a cast shadow, the radiance decreases sharply, causing many events to be triggered. As a result, the reconstructed profiles may exhibit unexpected strong peaks. Analyzing these strong peaks allows us to identify such distorted regions in the profile. In this section, we propose methods to design masks to remove regions of the profile distorted by non-Lambertian effects.

\begin{figure}[t]
    \centering
    %%%%%%%%%%%%%%%%%%%%%%%%%%%%%%
    \begin{subfigure}[t]{0.95\linewidth}
        \centering
        \includegraphics[width=\linewidth]{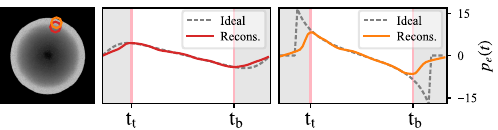}
        \put(-\linewidth + 4ex, 0.5ex){\heatmap}
        \vspace{-0.5ex}
        \caption{}
        \label{fig:mask diffuse}
    \end{subfigure} \\
    %%%%%%%%%%%%%%%%%%%%%%%%%%%%%%
    \vspace{-0.5ex}
    
    %%%%%%%%%%%%%%%%%%%%%%%%%%%%%%
    \begin{subfigure}[t]{0.95\linewidth}
        \centering
        \includegraphics[width=\linewidth]{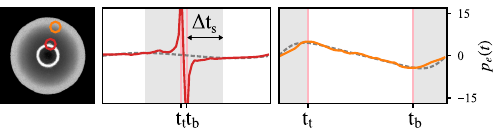}
        \put(-\linewidth + 4ex, 0.5ex){\heatmap}
        \vspace{-0.5ex}
        \caption{}
        \label{fig:mask glossy}
    \end{subfigure} \\
    %%%%%%%%%%%%%%%%%%%%%%%%%%%%%%
    \vspace{-0.5ex}

    %%%%%%%%%%%%%%%%%%%%%%%%%%%%%%
    \begin{subfigure}[t]{0.95\linewidth}
        \centering
        \includegraphics[width=\linewidth]{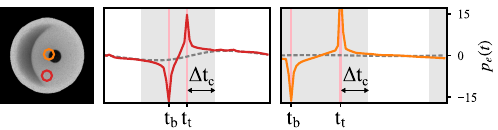}
        \put(-\linewidth + 4ex, 0.5ex){\heatmap}
        \vspace{-0.5ex}
        \caption{}
        \label{fig:mask shadow}
    \end{subfigure} \\
    %%%%%%%%%%%%%%%%%%%%%%%%%%%%%%
    
    \caption{Designs of the masks for non-Lambertian effects. 
    The left: intensity-based image from events, \heatmap.
    The middle and right: reconstructed profiles at the red and orange circle points in \heatmap, respectively.
    The gray dashed line represents the corresponding ideal profile. The gray fill indicates the masked regions to be removed.}
    
    % The images on the left denote an accumulation of all the recorded events, in which black to white represents zero to high.
    % The colors of circles correspond to those of the profiles on the right. 
    % The gray dash lines and the gray areas are the ideal profiles and masked regions, respectively.}
    \label{fig:practical masks}
\end{figure}

\paragraph{Mask for the collapsed events.}
In our system, a diffuse sphere, \diffusesphere in \cref{fig:eval spheres}, is measured, and profiles reconstructed at two surface points are analyzed. In \cref{fig:mask diffuse}, the red and orange points on the intensity-based image are selected, and the profiles at those points are shown next. 
The reconstructed profile (red) is very close to the corresponding ideal profile (gray dashed). However, we found that regions before the top peak and after the bottom peak appear distorted. This could be due to collapsed events. Therefore, we design the mask as follows
\begin{equation}
    \maskgeneral =
    \begin{cases}
        1 & \text{if $\time \in [\timetoppeak, \timebottompeak]$,} \\
        0 & \text{otherwise,} \\
    \end{cases}
\end{equation}
where \timetoppeak and \timebottompeak are the times of the top and bottom peaks, respectively. This mask is visualized as the gray fill, where $\maskgeneral = 0$. Another finding is that the same mask can remove the effect of attached shadows (orange). When the attached shadows occur, the intensity change is too precipitous to trigger events properly, as shown in \cref{fig:mask diffuse} (right). 
However, the region between the peaks seems to fit to the ideal profile. Practically, \maskgeneral is used for all points, because \maskgeneral works better than \maskattached.

\paragraph{Mask for the specularity.}
A glossy sphere, \glossysphere in \cref{fig:eval spheres}, is measured, and the reconstructed profile at a highlight point is shown in red in \cref{fig:mask glossy}. We found that the reconstructed profile at a highlight point generally includes a pair of positive and negative peaks, and the profile around these peaks deviates from the ideal one. To remove the deviated region, a mask for specularity is defined as follows
\begin{equation}
    \maskglossiness =
    \begin{cases}
            0 & \text{if $\time \in [\timetoppeak - \marginglossiness, \timebottompeak + \marginglossiness]$,} \\
        1 & \text{otherwise,} \\
    \end{cases}
\end{equation}
where \marginglossiness is a margin to expand the size of the mask, which is required for the specular lobe.
Note that highlight points are limited even on the glossy material, as can be seen in the intensity-based image. When selecting a non-highlight point (orange), the reconstructed profile is similar to that of the diffuse sphere.

\paragraph{Mask for cast shadow.}
The effect of cast shadow can be handled using a similar approach to that for specularity. A sphere with a pole, as shown as \shadowsphere in \cref{fig:eval spheres}, is measured. Since the pole blocks light, cast shadow occur, which can be seen as a brighter gray area in the intensity-based image in \cref{fig:mask shadow}. When a surface point (red) suddenly moves in and out of the shadows, the reconstructed profile includes a pair of peaks. Unlike the highlights, the peaks are first negative and then positive. Thus, the mask for cast shadow is defined as
\begin{equation}
    \maskcast =
    \begin{cases}
        0 & \text{if $\time \in [\timebottompeak - \margincast, \timetoppeak + \margincast]$,} \\
        1 & \text{otherwise,} \\
    \end{cases}
\end{equation}
where \margincast is a margin. Importantly, this mask works well for a different point (orange), even though the effect of cast shadow varies with respect to the geometry.

\paragraph{Mask process.}
First, in our implementation, \maskgeneral is applied to all surface points, instead of \maskattached as in \cref{eq:cost}.
Second, if the cost after the minimization is higher than a given threshold, \maskglossiness or \maskcast is applied instead of \maskgeneral. 
The second mask is determined based on the duration between \timetoppeak and \timebottompeak. If the duration is smaller than a given threshold, \maskglossiness is used, as it is most likely due to the highlights.
Note that we apply a single mask per outlier segment, regardless of the number of outlier types.

\section{Implementation}
\label{sec:implementation}

\subsection{Prototype System}
\label{sec:prototype}

\begin{figure}[b]
    \centering
    \begin{subfigure}[t]{0.65\linewidth}
        \includegraphics[width=\linewidth]{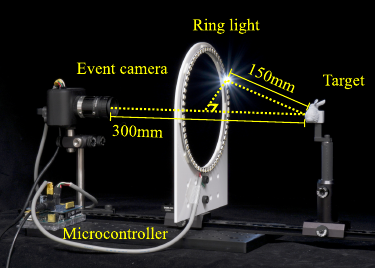}
        \caption{}    
        \label{fig:prototype}
    \end{subfigure}
    \begin{subfigure}[t]{0.27\linewidth}
        \includegraphics[width=\linewidth]{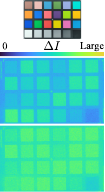}
        \caption{}    
        \label{fig:offset light}
    \end{subfigure}
    % \caption{Implementation. (a)~Prototype system. (b)~Correction by the offset light. From top to bottom: A color checker, the event accumulation images in a color map with $10$ and \SI{50}{\%} offset light to the maximum power, respectively. With the offset, the effect of albedos can be reduced.}
    \caption{Implementation. (a)~Prototype system. (b)~Top: Color checker. Middle and bottom: Event accumulation images under a light source modulated by a triangular wave. The power range of the light source is from $10$ to \SI{20}{\%} for the middle and from $50$ to \SI{100}{\%} for the bottom ($k=2$).}
    \label{fig:impl}
\end{figure}

We implemented a prototype system using an event camera (Prophesee EVK) equipped with a \SI{16}{\mm} fixed-focus lens (Fujinon HF16XA5M) and a ring light module (Adafruit NeoPixel Ring 60 RGBW) controlled by a microcontroller (Arduino UNO R3), as shown in \cref{fig:prototype}. We selected a circular motion for the trajectory due to its straightforward implementation and calibration. The ring light is positioned so that its center aligns with the optical axis of the camera, and its circular plane is perpendicular to this axis. Sync signals for every cycle are sent from the microcontroller to the event camera.

In principle, the light source is assumed to move continuously; however, the ring light, composed of multiple LEDs, can simulate continuous light motion through a smooth transition of intensity between adjacent LEDs. Since the LED intensity is controlled via pulse width modulation, the low-pass filter within the event camera is carefully adjusted to prevent events triggered by the pulse modulation. By employing a ring light instead of a physically moving light source, uniform offset lighting can be applied.
In the prototype, the light can be considered distant due to the small size of the target objects, \eg, \SI{25}{\milli\metre}. The camera model is assumed to be orthographic.

% The light is assumed to be distant. Thus, the ring light is located as each LED is around \SI{150}{\milli\metre} far from a target object whose size is roughly \SI{25}{\milli\metre}.
% The camera model in the principle is assumed to be orthographic. Thus, in the prototype system, the distance from the camera to an object is set to far enough for the assumption. However, the lens distortion is calibrated like the conventional cameras because it is not completely orthographic.

\subsection{Calibration}

The prototype system requires calibration of two factors: event thresholds and light trajectory. 

\paragraph{Event thresholds.}
The spatial characteristics among pixels in event cameras are not calibrated as they are in conventional cameras. Generally, the event thresholds, \posthresh and \negthresh, vary across all pixels. Therefore, it is necessary to estimate the thresholds for each pixel individually.

We explain here how to estimate the positive threshold, nothing that the negative one can be estimated as well. If a monotonically increasing function of \time is applied to the power of light sources, denoted as $\sourcepower(\time)$, at fixed positions, the radiance is given by
\begin{equation}
    \radiance = \sourcepower(\time) \albedo \normal^\top \fixedlightdirection.
\end{equation}
Assuming \numposevents positive events occur over the duration from $\time_1$ to $\time_2$, the threshold can be approximately calculated as follows
\begin{equation}
    \posthresh \simeq \frac{1}{\numposevents} \ln{\frac{\sourcepower(\time_2)}{\sourcepower(\time_1)}}.
\end{equation}
This approximation holds because event generation closely follows $\diffsig = \posthresh$ in \cref{eq:compsig} when the brightness changes gradually. However, in practice, event timestamps are often delayed, and the threshold itself can fluctuate due to heat variations.

If $\sourcepower(\time_2)$ is scaled by a factor of $k$ relative to $\sourcepower(\time_1)$, then
\begin{equation}
    \posthresh \simeq \frac{\ln{k}}{\numposevents}.
    \label{eq:hNk}
\end{equation}
Note that $\sourcepower(\time_1)$ should not be too small due to the characteristics of the logarithmic amplifier.
In our case, we used $k = 6$ under uniform illumination, resulting in approximately $20$ recorded events per pixel.

\paragraph{Light trajectory}
In the proposed method, the trajectory of the light source must be known. The commercial ring light used in the prototype system is designed with a precise circular shape. Therefore, we estimate the light directions for four out of the $60$ LEDs and then fit a circle to extrapolate the remaining directions.
For these estimations, Santo's method~\cite{Santo2020light} is applied using three pins and $10$ poses per LED. 
To detect feature points and pin shadows on the calibration board, it is necessary to reconstruct event accumulation images. 
Events are first recorded under each of the four LEDs, modulated by a triangular wave. 
Assuming that \numposevents positive events are recorded, the event accumulation image is defined as
\begin{equation}
    \heatmap = \posthresh \numposevents.
\end{equation}
Finally, the event accumulation images from all poses are used to estimate the LED direction as a distant light source.

\subsection{Offset Light}
We found that adding an offset to the illumination is crucial for event-based photometric analysis. 
The input-output characteristic of logarithmic amplifiers do not always adhere to an ideal logarithmic function. 
In practice, when the input voltage falls below a certain threshold, the output is floored.
This means that the desired series of events cannot be recorded when \radiance is low. 
Consequently, fewer events are captured when albedo is lower, even though albedo should not inherently affect the number of events.
The middle and bottom panels of \cref{fig:offset light} show event accumulation images for a color checker under a modulated light source, illustrating that fewer events are recorded when \radiance is too low.

To prevent \radiance from becoming too small, we add an offset light in all experiments by dimly turning on all the LEDs. 
This modifies \cref{eq:radiance} as follows
\begin{equation}
    \radiance = \sourcepower \albedo \max(\normal^\top \lightdirection, 0) + \offsetlight,
\end{equation}
where \offsetlight represents the offset light. The time derivative is also modified from \cref{eq:ideal profile} to
\begin{equation}
    \profile[\time|\normal, \lightdirection, \offsetlight] = \frac{\normal^\top \derivlightdirection}{\max(\normal^\top \lightdirection, 0) + \offsetlight}.
\end{equation}
In this work, \offsetlight is assumed to be known.

\section{Experiments}
\label{sec:experiments}

\subsection{Fabrication of Objects with Known-Shapes}

%\todo{Additonally, a different event camera (Prophesee EVK4) is used for validating that the proposed technique is available for a different camera.} 
% The rotational frequency of the ring light is set to \SI{1}{Hz} and measurements for $30$ periods are averaged as an input.

For quantitative analyses and evaluations, we fabricate objects with known shapes using a stereolithography 3D printer whith a resolution of \SI{25}{\um} (Formlabs Form3 with Grey Resin V4). The shapes are categorized into two types: sphere-based and complicated. Photographs of the objects can be seen in \cref{fig:eval spheres,fig:eval complicated}. 
All the printed objects are spray-painted with a gray surfacer to make them diffuse. Colors for different albedos, such as white and yellow, are sprayed over the surfacer. A clear top coat spray is used to make them glossy, if necessary.

\subsection{The Number of Measurements for the Average}

A simple and effective noise model for event cameras is the statistical threshold model~\cite{Rebecq2018}, where event thresholds fluctuate according to a normal distribution due to various camera noises. Assuming this threshold, averaging multiple measurements can reduce noise. In the experiments, a measurement refers to observing events during one cycle of the ring light. A profile is reconstructed from these events, and the averaged profile for \numobs measurements is calculated.
% A simple and effective noise model for event cameras is the statistical threshold model~\cite{Rebecq2018}, in which event thresholds fluctuate according to a normal distribution due to various noises in the event camera. Assuming the statistical threshold, averaging multiple measurements can reduce the noise. In the experiments, a measurement refers to observing events during one cycle of the ring light. A profile is reconstructed from the events recorded during the measurement, and the averaged profile for \numobs measurements is then calculated.

\begin{figure}[t]
    \centering
    \begin{subfigure}[t]{0.495\linewidth}
        \centering
        \includegraphics[width=\linewidth]{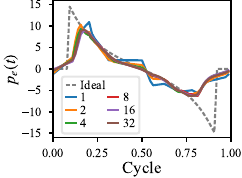}
        \caption{}
        \label{fig:avg profile}
    \end{subfigure}
    \begin{subfigure}[t]{0.495\linewidth}
        \centering
        \includegraphics[width=\linewidth]{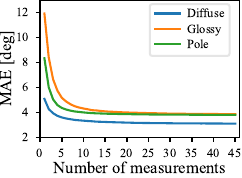}
        \caption{}
        \label{fig:avg angular error}
    \end{subfigure}
    
    \caption{Analysis on the number of measurements for the average. 
    (a)~Averaged profiles for the different numbers of measurements.
    (b)~Mean angular errors w.r.t. the number of measurements when targeting the three sphere-based objects.}
    \label{fig:avg}
\end{figure}

We analyze the performance of the proposed method for different numbers of measurements. \Cref{fig:avg profile} shows the reconstructed profiles at a surface point of \diffusesphere for $\numobs = \{1, 2, 4, 8, 16, 32\}$. The gray dashed line denotes the corresponding ideal profile from \cref{eq:ideal profile}. As \numobs increases, the shape of reconstructed profile better matches that of the ideal profile. \Cref{fig:avg angular error} shows the mean angular errors (MAEs) for \diffusesphere, \glossysphere, and \shadowsphere, which decrease as \numobs increases. After $30$ measurements, the MAE changes very little, so we use $30$ measurements for averaging in subsequent experiments.
%Here, we analyze the performance of the proposed method for different numbers of measurements. First, we compare the quality of the reconstructed profile with respect to the number of measurements. \Cref{fig:avg profile} shows the reconstructed profiles at a surface point of \diffusesphere for $\numobs = \{1, 2, 4, 8, 16, 32\}$. The gray dashed line denotes the corresponding ideal profile from \cref{eq:ideal profile}. As \numobs increases, the shape of reconstructed profile better matches that of the ideal profile. Second, we investigate the mean angular error (MAE) of the recovered normal map. \Cref{fig:avg angular error} shows the MAEs with respect to different \numobs for \diffusesphere, \glossysphere, and \shadowsphere. As observed, the MAE decreases as \numobs increases for all cases. After $30$ measurements, the MAE changes very little. For this reason, we use $30$ measurements for averaging in the experiments presented hereafter.

\subsection{Rotational Frequency of the Ring Light}

Next, we evaluate the performance relative to the rotational frequency of the ring light. The ring light is controlled by a microcomputer with a fixed-speed data stream, which limits the update rate of the illumination pattern. Consequently, the maximum achievable frequency for the rotating illumination pattern is determined to be \SI{2}{Hz}.

%When targeting \diffusesphere, the MAE is evaluated for a rotational frequency in the range of $0.1$ to \SI{2}{Hz}. It is discovered that the MAE is larger with increasing the rotational frequency, and also that can be large in a low frequency range, as shown in \cref{fig:freq without correction}. This is because of accumulating more noises in a longer period.We found that the reconstructed profiles are shifted along the time axis and the amount of the shift is related to the rotational frequency, as shown in \cref{fig:freq shifted profiles}. To analyze more, the recovered normal maps are separated into the zenith and azimuth maps, as shown in \cref{fig:freq azimuth zenith maps}. Note that the angles are indicated by discrete color codes. Compared with the reference, while the zenith maps seem similar to each other, the azimuth maps seem rotated linearly with respect to the frequency. We define the amount of ``rotation'' in the azimuth maps as an azimuth offset. The azimuth offset with respect to the rotational frequency is plotted in \cref{fig:freq azimuth offset} as the pink line, and the purple dash line is a fitted line in the range of $0.5$ to \SI{2}{Hz}. The fitted line allows to correct the azimuth offset for a rotational frequency in the range, as shown in \cref{fig:freq with correction}. Finally, \SI{1}{Hz} is determined the best rotational frequency in the prototype.

\begin{figure}[t]
    \centering
    \begin{subfigure}[t]{0.47\linewidth}
        \centering
        \includegraphics[width=\linewidth]{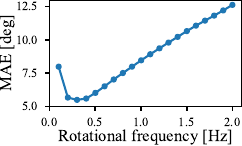}
        \caption{}
        \label{fig:freq without correction}
    \end{subfigure}
    \begin{subfigure}[t]{0.47\linewidth}
        \centering
        \includegraphics[width=\linewidth]{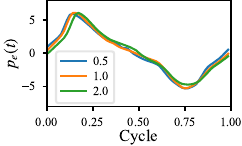}
        \caption{}
        \label{fig:freq shifted profiles}
    \end{subfigure}
    % \vspace{0.2em}
    
    \begin{subfigure}[t]{0.85\linewidth}
        \centering
        \includegraphics[width=\linewidth]{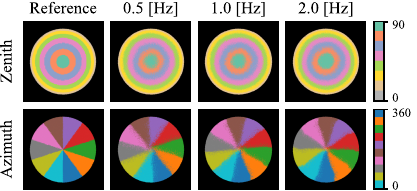}
        % \vspace{-1ex}
        \caption{}
        \label{fig:freq azimuth zenith maps}
    \end{subfigure}
    % \vspace{0.2em}
    
    \begin{subfigure}[t]{0.47\linewidth}
        \centering
        \includegraphics[width=\linewidth]{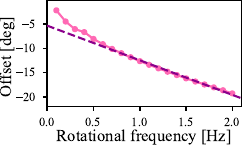}
        \caption{}
        \label{fig:freq azimuth offset}
    \end{subfigure}
    \begin{subfigure}[t]{0.47\linewidth}
        \centering
        \includegraphics[width=\linewidth]{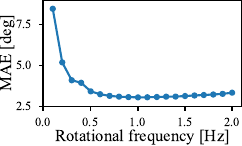}
        \caption{}
        \label{fig:freq with correction}
    \end{subfigure}
    
    \caption{Analysis of the rotational frequency of the ring light. (a)~Mean angular errors with respect to different rotational frequencies. (b)~Reconstructed profiles for frequencies of $0.5$, $1$, and \SI{2}{Hz}, with profiles shifted relative to the frequency. (c)~Zenith and azimuth maps extracted from a recovered normal map. (d)~Azimuth offset. (e)~Mean angular errors after applying azimuth offset correction.}
    \label{fig:freq}
\end{figure}

For \diffusesphere, the MAE is evaluated across rotational frequencies from $0.1$ to \SI{2}{Hz}. The MAE increases with higher frequencies and can also be large at very low frequencies due to noise accumulation over longer periods (\cref{fig:freq without correction}).
We observed that reconstructed profiles shift along the time axis, with the shift amount dependent on the frequency (\cref{fig:freq shifted profiles}).
Further analysis of the recovered normal maps, split into zenith and azimuth components (\cref{fig:freq azimuth zenith maps}), shows similar zenith maps but azimuth maps that rotate linearly with frequency. 
% This rotation, defined as the azimuth offset, is plotted in \cref{fig:freq azimuth offset}, with a fitted line for frequencies between $0.5$ and \SI{2}{Hz} aiding in offset correction (\cref{fig:freq with correction}).Ultimately, \SI{1}{Hz} is selected as the optimal frequency for the prototype.
\redstart
This rotation, defined as the azimuth offset, is plotted in \cref{fig:freq azimuth offset}. A linear fit between $0.5$ and \SI{2}{Hz} aids in offset correction, resulting in a nearly constant MAE within this frequency range (\cref{fig:freq with correction}). Finally, 1 Hz is selected as the optimal frequency.
\color{black}

%There is a hardware limit to speed of recording events, which is called the event rate.
%Another limit is derived from the ring light.

% \section{Experiments}
% \label{sec:experiments}

% We conduct three steps of experiments.
% First, a quantitative evaluation using known, simple shapes is done with 3D printed sphere-based objects. 
% Second, another quantitative evaluation using known, complicated shapes is done with 3D printed objects whose geometries are more realistic and practical.
% Third, a qualitative evaluation using unknown shapes is done with commercially available objects.

\subsection{Quantitative Evaluations}

To quantitatively evaluate performance, the 3D printed objects are measured. The ground truth normal map is computed from the 3D model and aligned to the observation using the event accumulation image through MeshLab's mutual information registration filter~\cite{Meshlab}, following the method in~\cite{Shi2016}. MAE is calculated only within an image-mask where an object exists, and the image-masks for all objects are manually made based on \heatmap.

\paragraph{Sphere-based shapes.}

\begin{figure}[t]
    % Settings
    \newcommand{\cellwidth}{0.134\linewidth} % 11.390
    \newcommand{\colorbarwidth}{0.03\linewidth} % 0.185 / 16
    \newcommand{\cellheight}{\linewidth}
    \setlength\tabcolsep{0pt}
    % \footnotesize
    \scriptsize
    \centering
        
    \begin{tabular*}{\linewidth}{@{\extracolsep{\fill}}ccccccccc}
        %%%%%%%%%%%%%%%%%%%%%%%%%%%%%%%%%%%%%%%%%%%%%%%%%%
        & Photo & \heatmap & GT & \multicolumn{2}{c}{EventPS-FCN~\cite{EventPS}} & \multicolumn{2}{c}{Ours} \\
        %%%%%%%%%%%%%%%%%%%%%%%%%%%%%%%%%%%%%%%%%%%%%%%%%%
    
        %%%%%%%%%%%%%%%%%%%%%%%%%%%%%%%%%%%%%%%%%%%%%%%%%%
        \rotatebox[origin=l]{90}{\diffusesphere} &
        \begin{subfigure}[t]{\cellwidth}
            \centering
            \includegraphics[height=\cellheight]{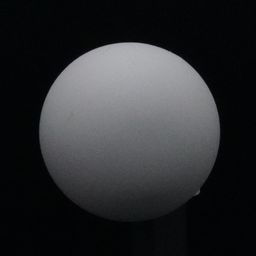}
        \end{subfigure} &
        \begin{subfigure}[t]{\cellwidth}
            \centering
            \includegraphics[height=\cellheight]{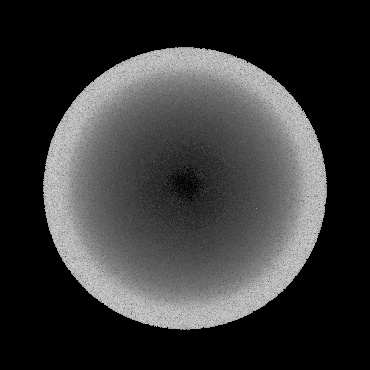}
        \end{subfigure} &
        \begin{subfigure}[t]{\cellwidth}
            \centering
            \includegraphics[height=\cellheight]{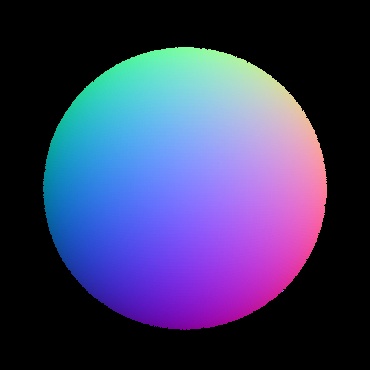}
        \end{subfigure} &
        \begin{subfigure}[t]{\cellwidth}
            \centering
            \includegraphics[height=\cellheight]{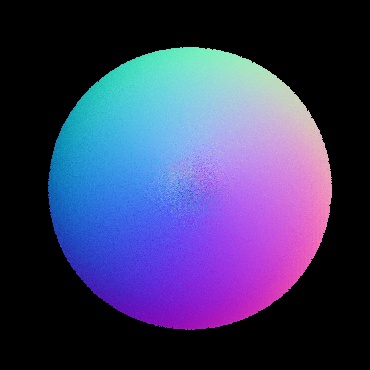}
        \end{subfigure} &
        \begin{subfigure}[t]{\cellwidth}
            \begin{picture}(\cellwidth, \cellheight)
                \put(0,0){\includegraphics[height=\cellheight]{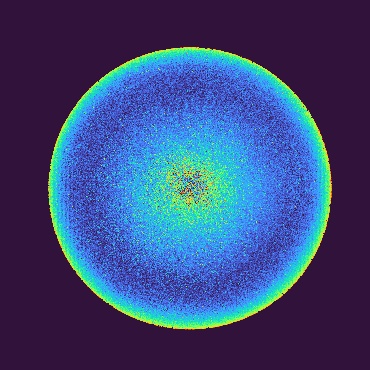}}
                \put(1,2){\mae{7.96}}
            \end{picture}
        \end{subfigure} &
        \begin{subfigure}[t]{\cellwidth}
            \centering
            \includegraphics[height=\cellheight]{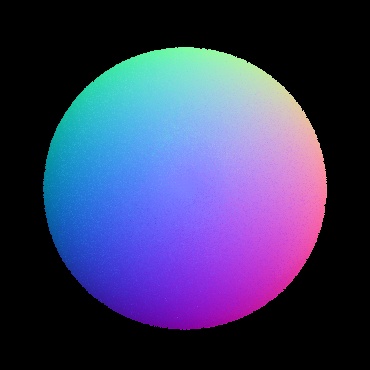}
        \end{subfigure} &
        \begin{subfigure}[t]{\cellwidth}
            \begin{picture}(\cellwidth, \cellheight)
                \put(0,0){\includegraphics[height=\cellheight]{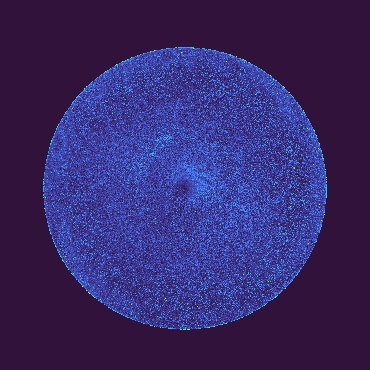}}
                \put(1,2){\mae{2.91}}
            \end{picture}
        \end{subfigure} &
        \begin{subfigure}[t]{\colorbarwidth}
            \centering
            \includegraphics[width=\linewidth]{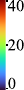}
        \end{subfigure} \\
        %%%%%%%%%%%%%%%%%%%%%%%%%%%%%%%%%%%%%%%%%%%%%%%%%%

        %%%%%%%%%%%%%%%%%%%%%%%%%%%%%%%%%%%%%%%%%%%%%%%%%%
        \rotatebox[origin=l]{90}{\glossysphere} &
        \begin{subfigure}[t]{\cellwidth}
            \centering
            \includegraphics[height=\cellheight]{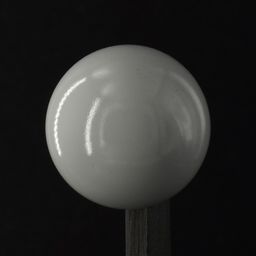}
        \end{subfigure} &
        \begin{subfigure}[t]{\cellwidth}
            \centering
            \includegraphics[height=\cellheight]{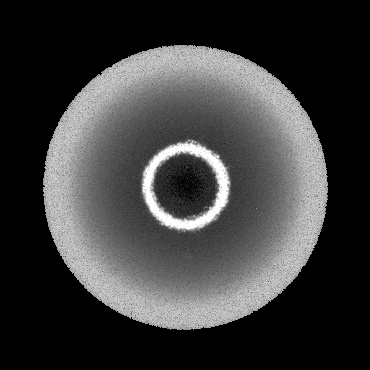}
        \end{subfigure} &
        \begin{subfigure}[t]{\cellwidth}
            \centering
            \includegraphics[height=\cellheight]{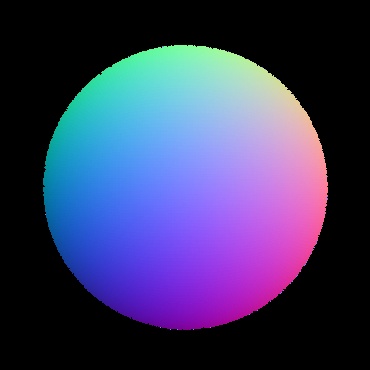}
        \end{subfigure} &
        \begin{subfigure}[t]{\cellwidth}
            \centering
            \includegraphics[height=\cellheight]{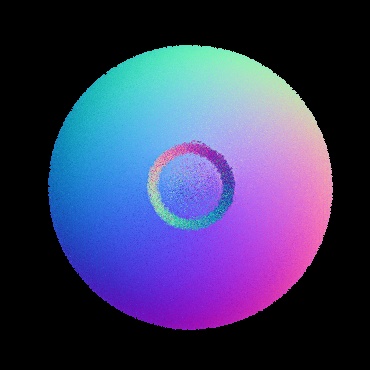}
        \end{subfigure} &
        \begin{subfigure}[t]{\cellwidth}
            \begin{picture}(\cellwidth, \cellheight)
                \put(0,0){\includegraphics[height=\cellheight]{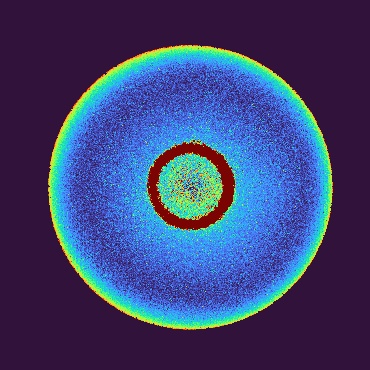}}
                \put(1,2){\mae{10.12}}
            \end{picture}
        \end{subfigure} &
        \begin{subfigure}[t]{\cellwidth}
            \centering
            \includegraphics[height=\cellheight]{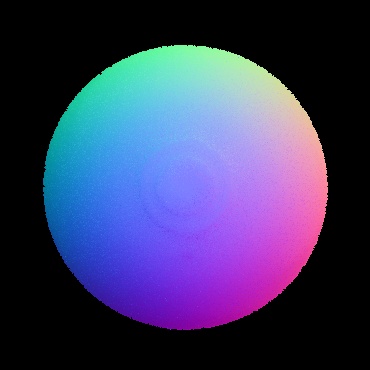}
        \end{subfigure} &
        \begin{subfigure}[t]{\cellwidth}
            \begin{picture}(\cellwidth, \cellheight)
                \put(0,0){\includegraphics[height=\cellheight]{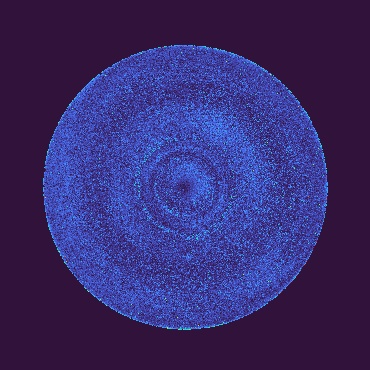}}
                \put(1,2){\mae{3.16}}
            \end{picture}
        \end{subfigure} &
        \begin{subfigure}[t]{\colorbarwidth}
            \centering
            \includegraphics[width=\linewidth]{figs/results/colorbar1.pdf}
        \end{subfigure} \\
        %%%%%%%%%%%%%%%%%%%%%%%%%%%%%%%%%%%%%%%%%%%%%%%%%%
    
        %%%%%%%%%%%%%%%%%%%%%%%%%%%%%%%%%%%%%%%%%%%%%%%%%%
        \rotatebox[origin=l]{90}{\shadowsphere} &
        \begin{subfigure}[t]{\cellwidth}
            \centering
            \includegraphics[height=\cellheight]{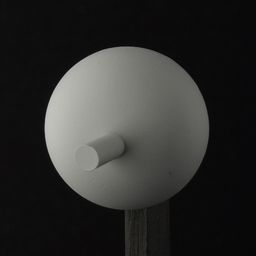}
        \end{subfigure} &
        \begin{subfigure}[t]{\cellwidth}
            \centering
            \includegraphics[height=\cellheight]{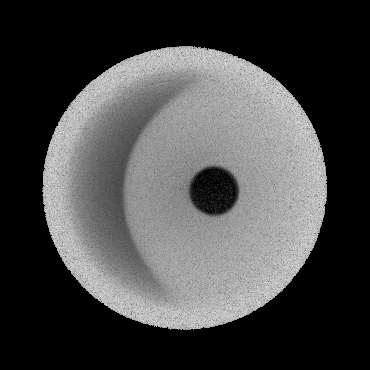}
        \end{subfigure} &
        \begin{subfigure}[t]{\cellwidth}
            \centering
            \includegraphics[height=\cellheight]{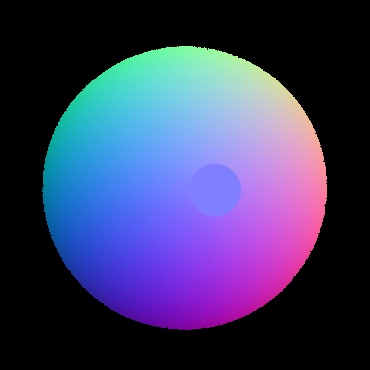}
        \end{subfigure} &
        \begin{subfigure}[t]{\cellwidth}
            \centering
            \includegraphics[height=\cellheight]{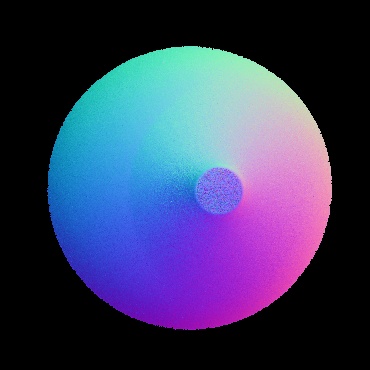}
        \end{subfigure} &
        \begin{subfigure}[t]{\cellwidth}
            \begin{picture}(\cellwidth, \cellheight)
                \put(0,0){\includegraphics[height=\cellheight]{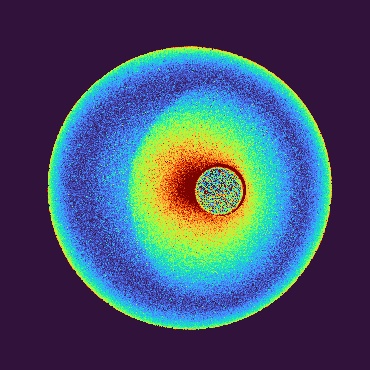}}
                \put(1,2){\mae{12.02}}
            \end{picture}
        \end{subfigure} &
        \begin{subfigure}[t]{\cellwidth}
            \centering
            \includegraphics[height=\cellheight]{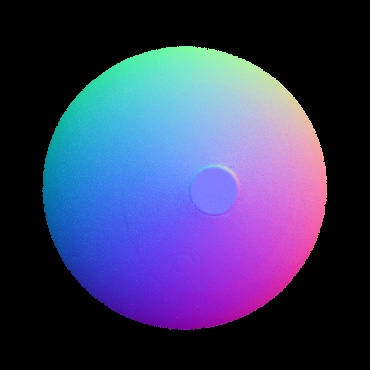}
        \end{subfigure} &
        \begin{subfigure}[t]{\cellwidth}
            \begin{picture}(\cellwidth, \cellheight)
                \put(0,0){\includegraphics[height=\cellheight]{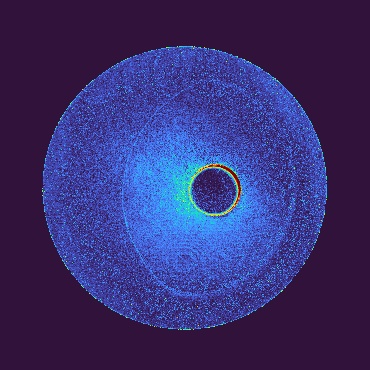}}
                \put(1,2){\mae{3.71}}
            \end{picture}
        \end{subfigure} &
        \begin{subfigure}[t]{\colorbarwidth}
            \centering
            \includegraphics[width=\linewidth]{figs/results/colorbar1.pdf}
        \end{subfigure} \\
        %%%%%%%%%%%%%%%%%%%%%%%%%%%%%%%%%%%%%%%%%%%%%%%%%%
    
        %%%%%%%%%%%%%%%%%%%%%%%%%%%%%%%%%%%%%%%%%%%%%%%%%%
        \rotatebox[origin=l]{90}{\twocolorsphere} &
        \begin{subfigure}[t]{\cellwidth}
            \centering
            \includegraphics[height=\cellheight]{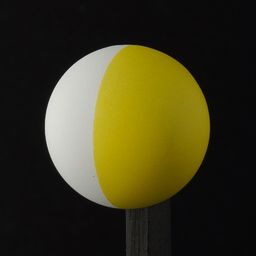}
        \end{subfigure} &
        \begin{subfigure}[t]{\cellwidth}
            \centering
            \includegraphics[height=\cellheight]{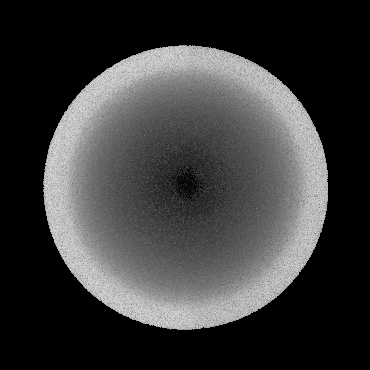}
        \end{subfigure} &
        \begin{subfigure}[t]{\cellwidth}
            \centering
            \includegraphics[height=\cellheight]{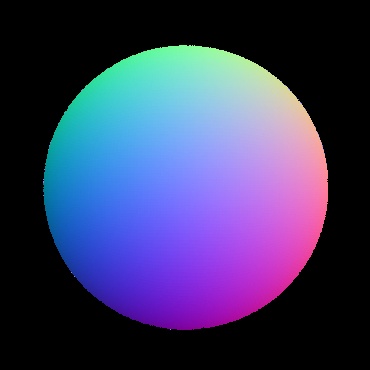}
        \end{subfigure} &
        \begin{subfigure}[t]{\cellwidth}
            \centering
            \includegraphics[height=\cellheight]{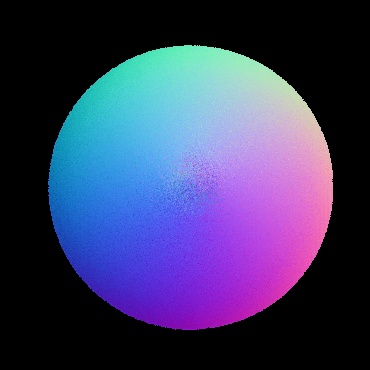}
        \end{subfigure} &
        \begin{subfigure}[t]{\cellwidth}
            \begin{picture}(\cellwidth, \cellheight)
                \put(0,0){\includegraphics[height=\cellheight]{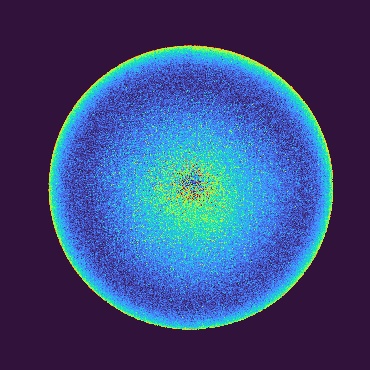}}
                \put(1,2){\mae{7.78}}
            \end{picture}
        \end{subfigure} &
        \begin{subfigure}[t]{\cellwidth}
            \centering
            \includegraphics[height=\cellheight]{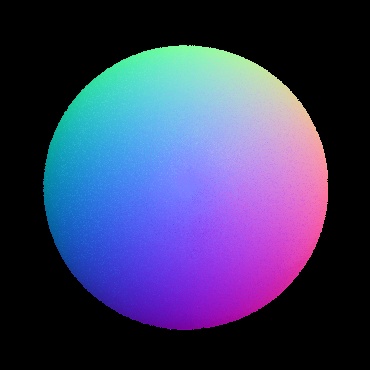}
        \end{subfigure} &
        \begin{subfigure}[t]{\cellwidth}
            \begin{picture}(\cellwidth, \cellheight)
                \put(0,0){\includegraphics[height=\cellheight]{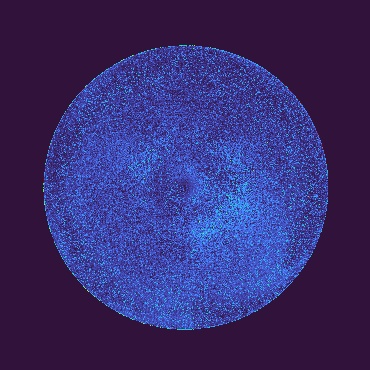}}
                \put(1,2){\mae{3.38}}
            \end{picture}
        \end{subfigure} &
        \begin{subfigure}[t]{\colorbarwidth}
            \centering
            \includegraphics[width=\linewidth]{figs/results/colorbar1.pdf}
        \end{subfigure} \\
        %%%%%%%%%%%%%%%%%%%%%%%%%%%%%%%%%%%%%%%%%%%%%%%%%%
    
        %%%%%%%%%%%%%%%%%%%%%%%%%%%%%%%%%%%%%%%%%%%%%%%%%%
        \rotatebox[origin=l]{90}{\multirefsphere} &
        \begin{subfigure}[t]{\cellwidth}
            \centering
            \includegraphics[height=\cellheight]{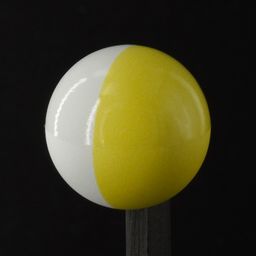}
        \end{subfigure} &
        \begin{subfigure}[t]{\cellwidth}
            \centering
            \includegraphics[height=\cellheight]{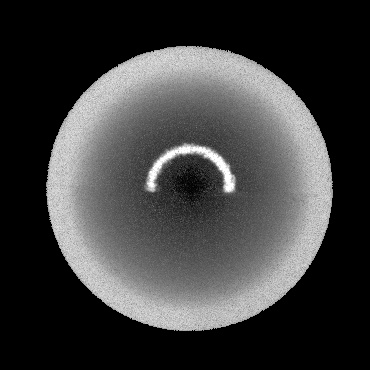}
        \end{subfigure} &
        \begin{subfigure}[t]{\cellwidth}
            \centering
            \includegraphics[height=\cellheight]{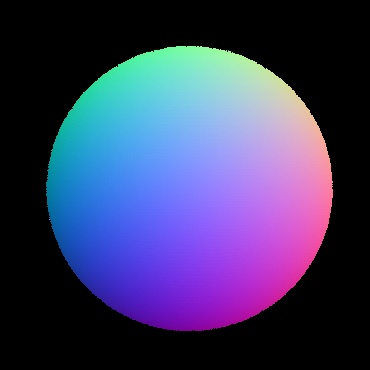}
        \end{subfigure} &
        \begin{subfigure}[t]{\cellwidth}
            \centering
            \includegraphics[height=\cellheight]{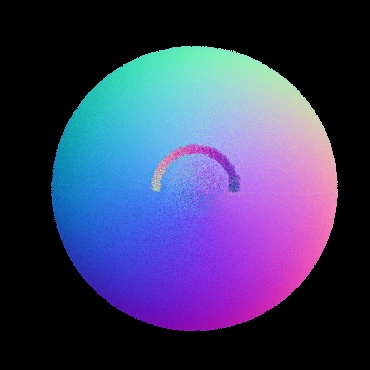}
        \end{subfigure} &
        \begin{subfigure}[t]{\cellwidth}
            \begin{picture}(\cellwidth, \cellheight)
                \put(0,0){\includegraphics[height=\cellheight]{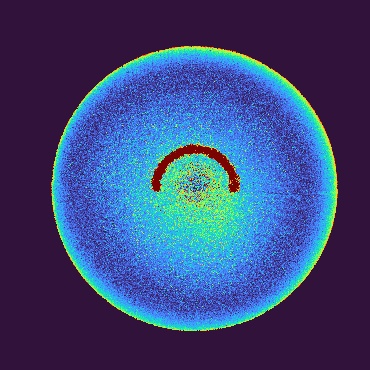}}
                \put(1,2){\mae{8.68}}
            \end{picture}
        \end{subfigure} &
        \begin{subfigure}[t]{\cellwidth}
            \centering
            \includegraphics[height=\cellheight]{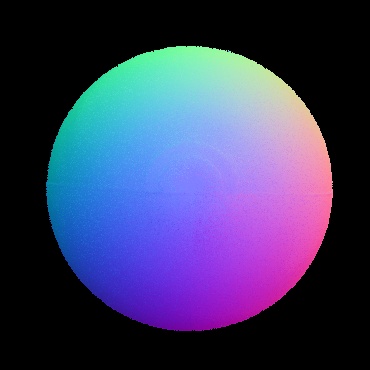}
        \end{subfigure} &
        \begin{subfigure}[t]{\cellwidth}
            \begin{picture}(\cellwidth, \cellheight)
                \put(0,0){\includegraphics[height=\cellheight]{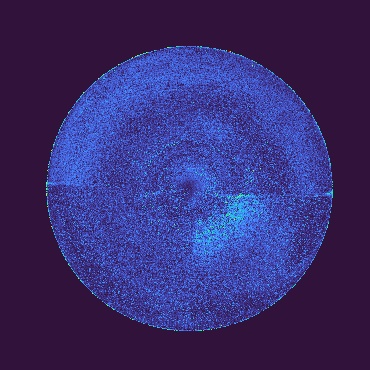}}
                \put(1,2){\mae{3.30}}
            \end{picture}
        \end{subfigure} &
        \begin{subfigure}[t]{\colorbarwidth}
            \centering
            \includegraphics[width=\linewidth]{figs/results/colorbar1.pdf}
        \end{subfigure} \\
        %%%%%%%%%%%%%%%%%%%%%%%%%%%%%%%%%%%%%%%%%%%%%%%%%%
    
    \end{tabular*}
    
    \caption{Quantitative evaluation with the sphere-based shapes. From left to right: photograph, event accumulation image from the events, ground truth, normal map recovered by EventPS-FCN~\cite{EventPS} with its angular error map, and those by ours.}
    \label{fig:eval spheres}
\end{figure}

In addition to the three sphere-based shapes (\diffusesphere, \glossysphere, and \shadowsphere), two additional spheres are used: \twocolorsphere, which has two colors, and \multirefsphere, with a glossy top half and a diffuse bottom half. The experimental results are shown in~\cref{fig:eval spheres}. These results demonstrate that the proposed masks effectively reduce angular errors for \glossysphere, \shadowsphere, and \multirefsphere, even under strong non-Lambertian effects. For results without masks, refer to the supplementary material. Compared to EventPS-FCN~\cite{EventPS}, our method shows greater robustness and accuracy. Including six additional sphere-based shapes shown in the supplementary material, the average MAEs for EventPS and our method are \SI{11.87}{\degree} and \SI{6.72}{\degree}, respectively.

\paragraph{Complex shapes.}

\begin{figure}[t]
    % Settings
    \newcommand{\cellwidth}{0.134\linewidth}
    \newcommand{\colorbarwidth}{0.03\linewidth}
    \newcommand{\cellheight}{\linewidth}
    \setlength\tabcolsep{0pt}
    % \footnotesize
    \scriptsize
    \centering

    \begin{tabular*}{\linewidth}{@{\extracolsep{\fill}}ccccccccc}
        %%%%%%%%%%%%%%%%%%%%%%%%%%%%%%%%%%%%%%%%%%%%%%%%%%
        & Photo & \heatmap & GT & \multicolumn{2}{c}{EventPS-FCN~\cite{EventPS}} & \multicolumn{2}{c}{Ours} \\
        %%%%%%%%%%%%%%%%%%%%%%%%%%%%%%%%%%%%%%%%%%%%%%%%%%
    
        %%%%%%%%%%%%%%%%%%%%%%%%%%%%%%%%%%%%%%%%%%%%%%%%%%
        \rotatebox[origin=l]{90}{\bunny} &
        \begin{subfigure}[t]{\cellwidth}
            \centering
            \includegraphics[height=\cellheight]{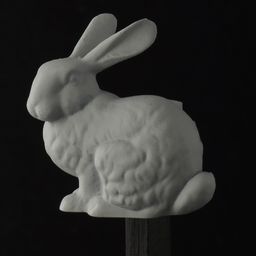}
        \end{subfigure} &
        \begin{subfigure}[t]{\cellwidth}
            \centering
            \includegraphics[height=\cellheight]{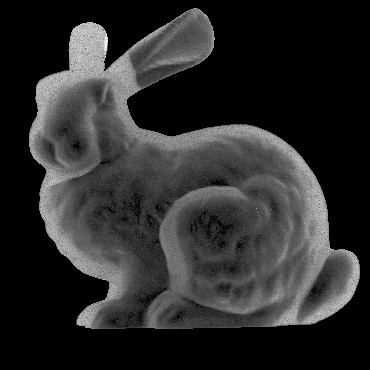}
        \end{subfigure} &
        \begin{subfigure}[t]{\cellwidth}
            \centering
            \includegraphics[height=\cellheight]{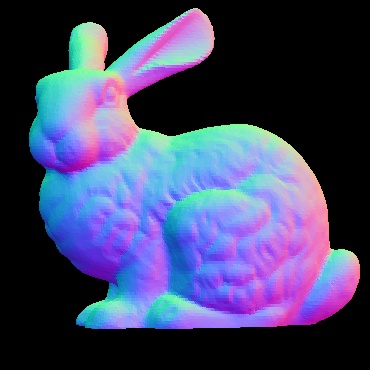}
        \end{subfigure} &
        \begin{subfigure}[t]{\cellwidth}
            \centering
            \includegraphics[height=\cellheight]{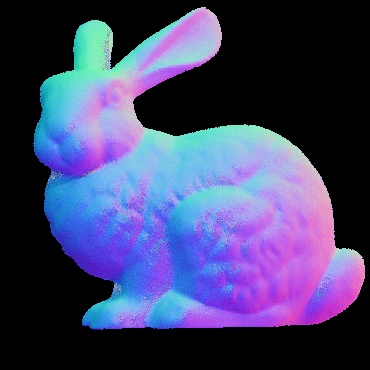}
        \end{subfigure} &
        \begin{subfigure}[t]{\cellwidth}
            \begin{picture}(\cellwidth, \cellheight)
                \put(0,0){\includegraphics[height=\cellheight]{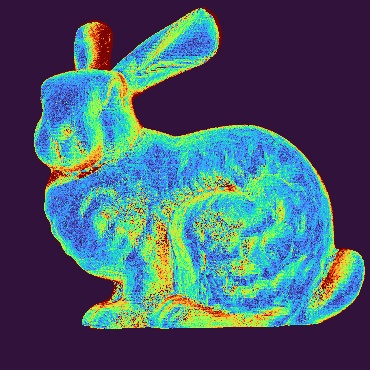}}
                \put(1,2){\mae{14.82}}
            \end{picture}
        \end{subfigure} &
        \begin{subfigure}[t]{\cellwidth}
            \centering
            \includegraphics[height=\cellheight]{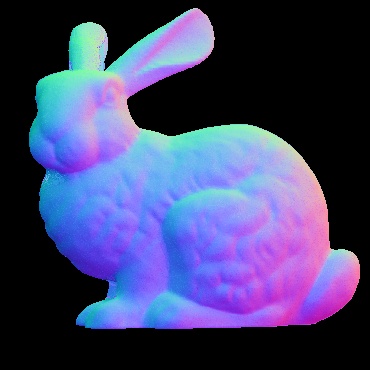}
        \end{subfigure} &
        \begin{subfigure}[t]{\cellwidth}
            \begin{picture}(\cellwidth, \cellheight)
                \put(0,0){\includegraphics[height=\cellheight]{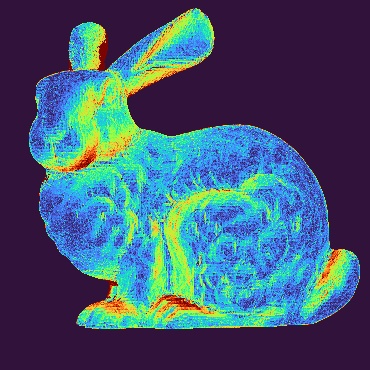}}
                \put(1,2){\mae{12.0}}
            \end{picture}
        \end{subfigure} &
        \begin{subfigure}[t]{\colorbarwidth}             
            \centering
            \includegraphics[width=\linewidth]{figs/results/colorbar1.pdf}
        \end{subfigure} \\
        %%%%%%%%%%%%%%%%%%%%%%%%%%%%%%%%%%%%%%%%%%%%%%%%%%
    
        %%%%%%%%%%%%%%%%%%%%%%%%%%%%%%%%%%%%%%%%%%%%%%%%%%
        \rotatebox[origin=l]{90}{\caesar} &
        \begin{subfigure}[t]{\cellwidth}
            \centering
            \includegraphics[height=\cellheight]{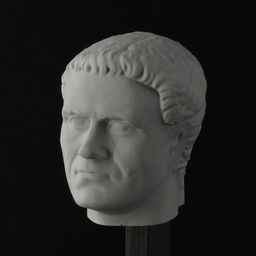}
        \end{subfigure} &
        \begin{subfigure}[t]{\cellwidth}
            \centering
            \includegraphics[height=\cellheight]{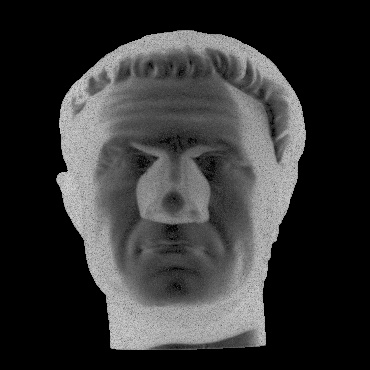}
        \end{subfigure} &
        \begin{subfigure}[t]{\cellwidth}
            \centering
            \includegraphics[height=\cellheight]{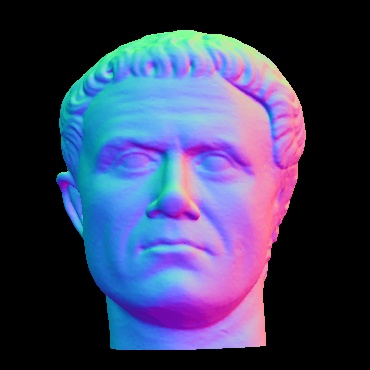}
        \end{subfigure} &
        \begin{subfigure}[t]{\cellwidth}
            \centering
            \includegraphics[height=\cellheight]{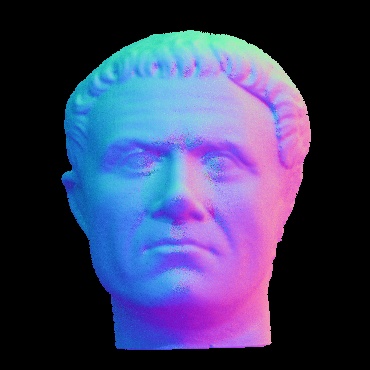}
        \end{subfigure} &
        \begin{subfigure}[t]{\cellwidth}
            \begin{picture}(\cellwidth, \cellheight)
                \put(0,0){\includegraphics[height=\cellheight]{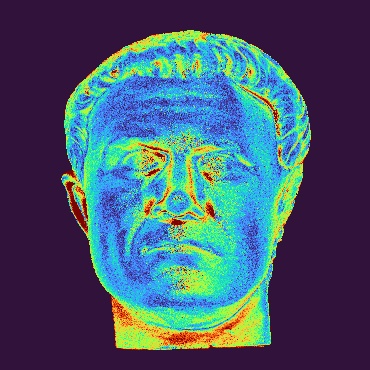}}
                \put(1,2){\mae{13.36}}
            \end{picture}
        \end{subfigure} &
        \begin{subfigure}[t]{\cellwidth}
            \centering
            \includegraphics[height=\cellheight]{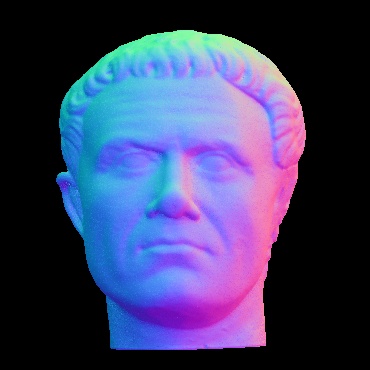}
        \end{subfigure} &
        \begin{subfigure}[t]{\cellwidth}
            \begin{picture}(\cellwidth, \cellheight)
                \put(0,0){\includegraphics[height=\cellheight]{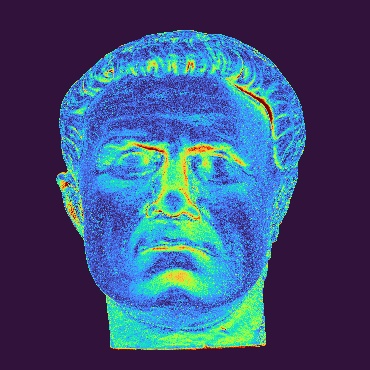}}
                \put(1,2){\mae{9.35}}
            \end{picture}
        \end{subfigure} &
        \begin{subfigure}[t]{\colorbarwidth}             
            \centering
            \includegraphics[width=\linewidth]{figs/results/colorbar1.pdf}
        \end{subfigure} \\
        %%%%%%%%%%%%%%%%%%%%%%%%%%%%%%%%%%%%%%%%%%%%%%%%%%
    
        %%%%%%%%%%%%%%%%%%%%%%%%%%%%%%%%%%%%%%%%%%%%%%%%%%
        \rotatebox[origin=l]{90}{\baboon} &
        \begin{subfigure}[t]{\cellwidth}
            \centering
            \includegraphics[height=\cellheight]{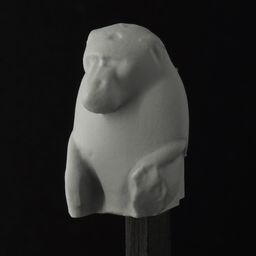}
        \end{subfigure} &
        \begin{subfigure}[t]{\cellwidth}
            \centering
            \includegraphics[height=\cellheight]{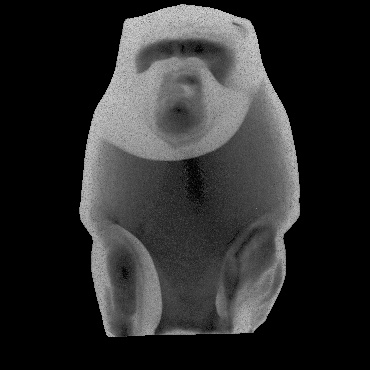}
        \end{subfigure} &
        \begin{subfigure}[t]{\cellwidth}
            \centering
            \includegraphics[height=\cellheight]{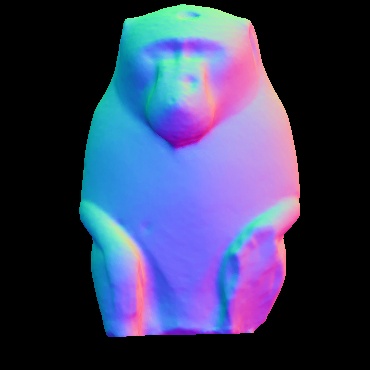}
        \end{subfigure} &
        \begin{subfigure}[t]{\cellwidth}
            \centering
            \includegraphics[height=\cellheight]{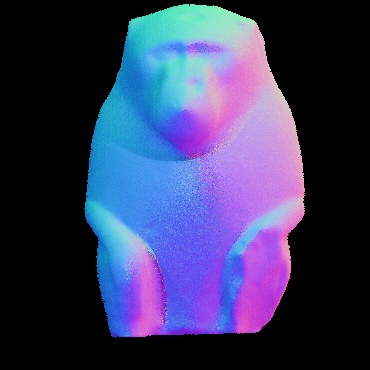}
        \end{subfigure} &
        \begin{subfigure}[t]{\cellwidth}
            \begin{picture}(\cellwidth, \cellheight)
                \put(0,0){\includegraphics[height=\cellheight]{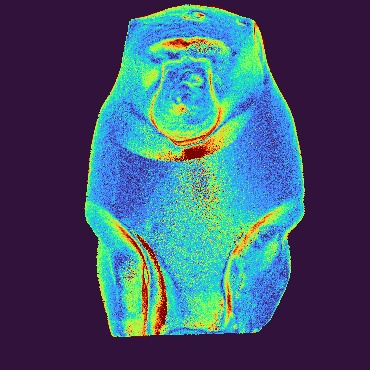}}
                \put(1,2){\mae{13.06}}
            \end{picture}
        \end{subfigure} &
        \begin{subfigure}[t]{\cellwidth}
            \centering
            \includegraphics[height=\cellheight]{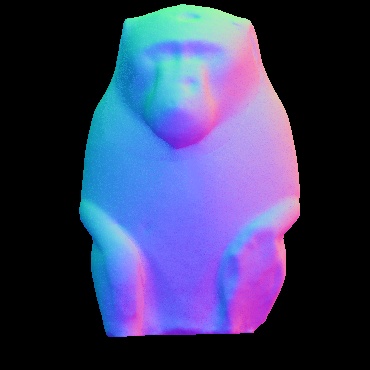}
        \end{subfigure} &
        \begin{subfigure}[t]{\cellwidth}
            \begin{picture}(\cellwidth, \cellheight)
                \put(0,0){\includegraphics[height=\cellheight]{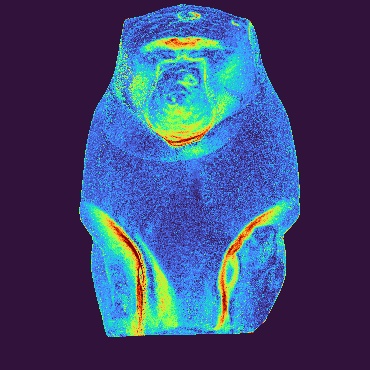}}
                \put(1,2){\mae{8.80}}
            \end{picture}
        \end{subfigure} &
        \begin{subfigure}[t]{\colorbarwidth}             
            \centering
            \includegraphics[width=\linewidth]{figs/results/colorbar1.pdf}
        \end{subfigure} \\
        %%%%%%%%%%%%%%%%%%%%%%%%%%%%%%%%%%%%%%%%%%%%%%%%%%
        
        %%%%%%%%%%%%%%%%%%%%%%%%%%%%%%%%%%%%%%%%%%%%%%%%%%
        \rotatebox[origin=l]{90}{\cv} &
        \begin{subfigure}[t]{\cellwidth}
            \centering
            \includegraphics[height=\cellheight]{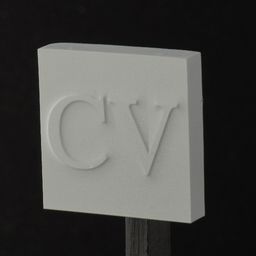}
        \end{subfigure} &
        \begin{subfigure}[t]{\cellwidth}
            \centering
            \includegraphics[height=\cellheight]{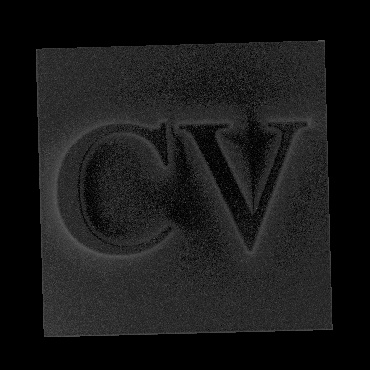}
        \end{subfigure} &
        \begin{subfigure}[t]{\cellwidth}
            \centering
            \includegraphics[height=\cellheight]{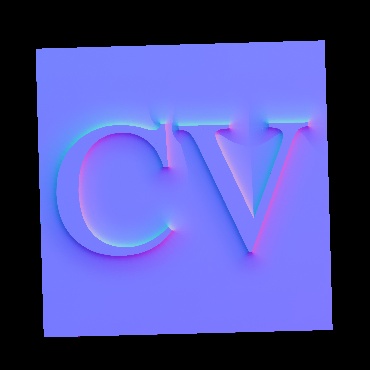}
        \end{subfigure} &
        \begin{subfigure}[t]{\cellwidth}
            \centering
            \includegraphics[height=\cellheight]{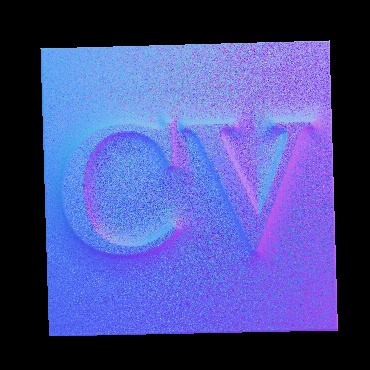}
        \end{subfigure} &
        \begin{subfigure}[t]{\cellwidth}
            \begin{picture}(\cellwidth, \cellheight)
                \put(0,0){\includegraphics[height=\cellheight]{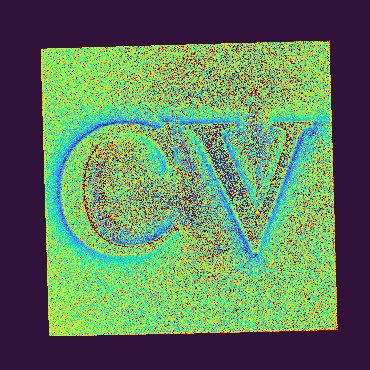}}
                \put(1,2){\mae{19.54}}
            \end{picture}
        \end{subfigure} &
        \begin{subfigure}[t]{\cellwidth}
            \centering
            \includegraphics[height=\cellheight]{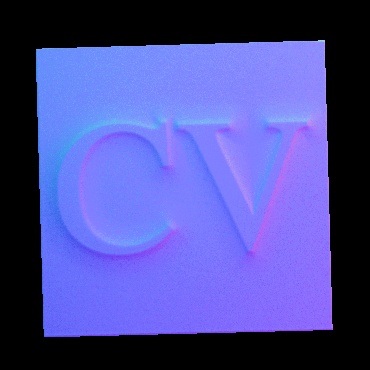}
        \end{subfigure} &
        \begin{subfigure}[t]{\cellwidth}
            \begin{picture}(\cellwidth, \cellheight)
                \put(0,0){\includegraphics[height=\cellheight]{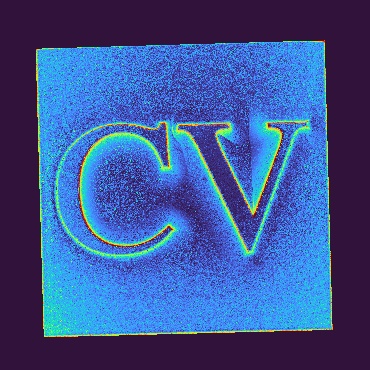}}
                \put(1,2){\mae{6.99}}
            \end{picture}
        \end{subfigure} &
        \begin{subfigure}[t]{\colorbarwidth}             
            \centering
            \includegraphics[width=\linewidth]{figs/results/colorbar1.pdf}
        \end{subfigure} \\
        %%%%%%%%%%%%%%%%%%%%%%%%%%%%%%%%%%%%%%%%%%%%%%%%%%
        
        %%%%%%%%%%%%%%%%%%%%%%%%%%%%%%%%%%%%%%%%%%%%%%%%%%
        \rotatebox[origin=l]{90}{\floral} &
        \begin{subfigure}[t]{\cellwidth}
            \centering
            \includegraphics[height=\cellheight]{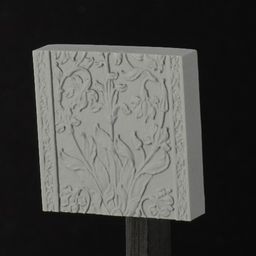}
        \end{subfigure} &
        \begin{subfigure}[t]{\cellwidth}
            \centering
            \includegraphics[height=\cellheight]{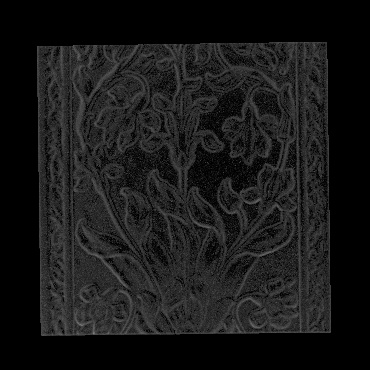}
        \end{subfigure} &
        \begin{subfigure}[t]{\cellwidth}
            \centering
            \includegraphics[height=\cellheight]{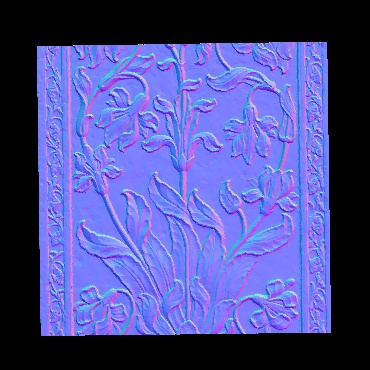}
        \end{subfigure} &
        \begin{subfigure}[t]{\cellwidth}
            \centering
            \includegraphics[height=\cellheight]{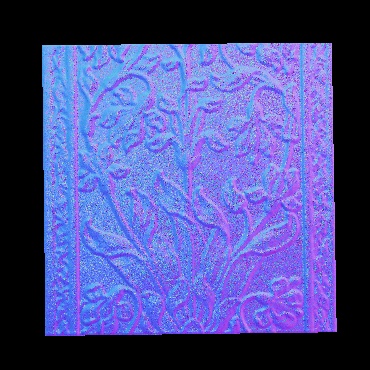}
        \end{subfigure} &
        \begin{subfigure}[t]{\cellwidth}
            \begin{picture}(\cellwidth, \cellheight)
                \put(0,0){\includegraphics[height=\cellheight]{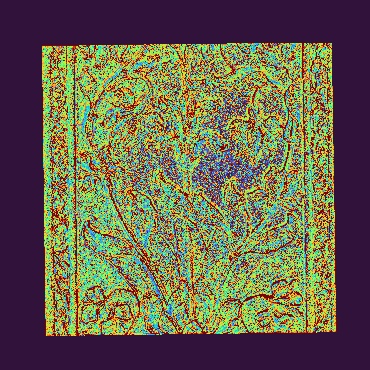}}
                \put(1,2){\mae{25.41}}
            \end{picture}
        \end{subfigure} &
        \begin{subfigure}[t]{\cellwidth}
            \centering
            \includegraphics[height=\cellheight]{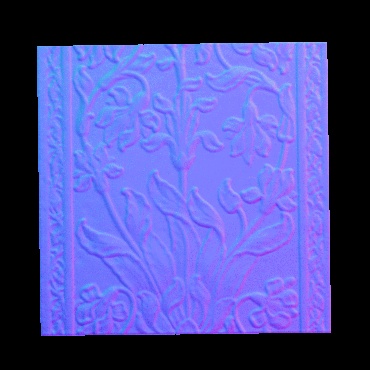}
        \end{subfigure} &
        \begin{subfigure}[t]{\cellwidth}
            \begin{picture}(\cellwidth, \cellheight)
                \put(0,0){\includegraphics[height=\cellheight]{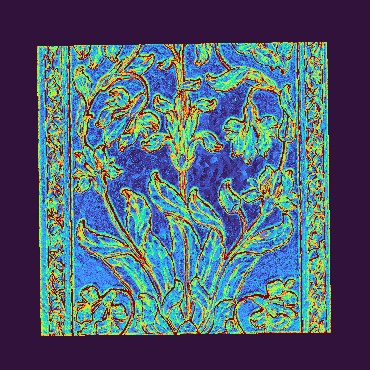}}
                \put(1,2){\mae{15.86}}
            \end{picture}
        \end{subfigure} &
        \begin{subfigure}[t]{\colorbarwidth}             
            \centering
            \includegraphics[width=\linewidth]{figs/results/colorbar1.pdf}
        \end{subfigure} \\
        %%%%%%%%%%%%%%%%%%%%%%%%%%%%%%%%%%%%%%%%%%%%%%%%%%

    \end{tabular*}
    
    \caption{Quantitative evaluation with the complex shapes. From left to right: photograph, event accumulation image from the events, ground truth, normal map recovered by EventPS-FCN~\cite{EventPS} with its angular error map, and those by ours.}
    \label{fig:eval complicated}
\end{figure}

\begin{figure}[t]
    % Settings
    \newcommand{\cellwidth}{0.162\linewidth}
    \newcommand{\cellheight}{\linewidth}
    \setlength\tabcolsep{0pt}
    % \footnotesize
    \scriptsize
    \centering

    \begin{tabular*}{\linewidth}{@{\extracolsep{\fill}}ccccccc}
        %%%%%%%%%%%%%%%%%%%%%%%%%%%%%%%%%%%%%%%%%%%%%%%%%%
        & Photo & \heatmap & \multicolumn{2}{c}{EventPS-FCN~\cite{EventPS}} & \multicolumn{2}{c}{Ours} \\
        %%%%%%%%%%%%%%%%%%%%%%%%%%%%%%%%%%%%%%%%%%%%%%%%%%
        
        %%%%%%%%%%%%%%%%%%%%%%%%%%%%%%%%%%%%%%%%%%%%%%%%%%
        \rotatebox[origin=l]{90}{\miffy} &
        \begin{subfigure}[t]{\cellwidth}
            \centering
            \includegraphics[height=\cellheight]{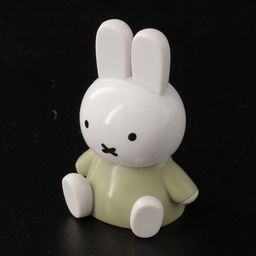}
        \end{subfigure} &
        \begin{subfigure}[t]{\cellwidth}
            \centering
            \includegraphics[height=\cellheight]{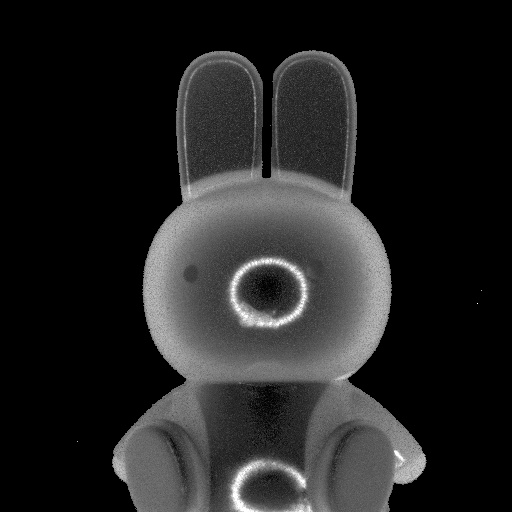}
        \end{subfigure} &
        \begin{subfigure}[t]{\cellwidth}
            \centering
            \includegraphics[height=\cellheight]{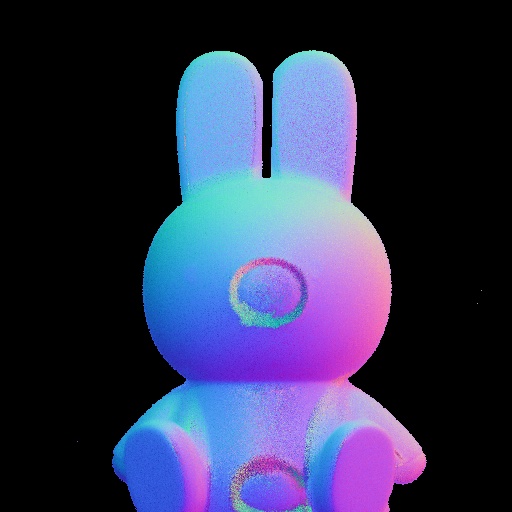}
        \end{subfigure} &
        \begin{subfigure}[t]{\cellwidth}
            \centering
            \includegraphics[height=\cellheight]{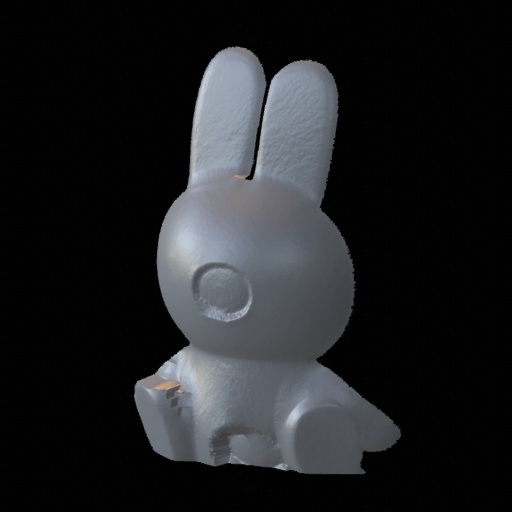}
        \end{subfigure} &
        \begin{subfigure}[t]{\cellwidth}
            \centering
            \includegraphics[height=\cellheight]{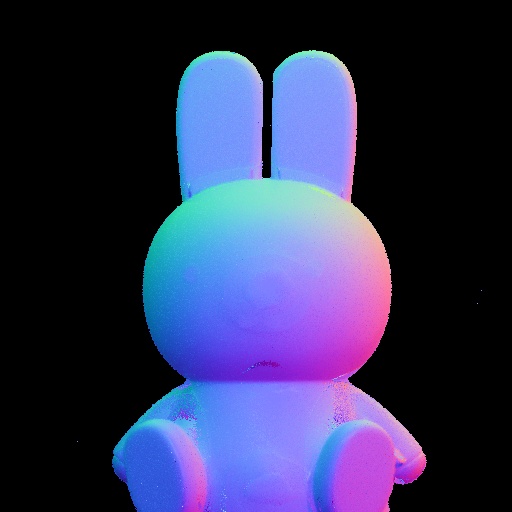}
        \end{subfigure} &
        \begin{subfigure}[t]{\cellwidth}
            \centering
            \includegraphics[height=\cellheight]{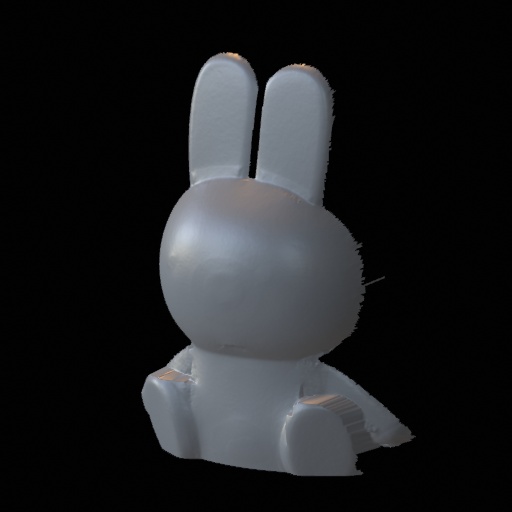}
        \end{subfigure} \\
        %%%%%%%%%%%%%%%%%%%%%%%%%%%%%%%%%%%%%%%%%%%%%%%%%%
    
        %%%%%%%%%%%%%%%%%%%%%%%%%%%%%%%%%%%%%%%%%%%%%%%%%%
        \rotatebox[origin=l]{90}{\retriever} &
        \begin{subfigure}[t]{\cellwidth}
            \centering
            \includegraphics[height=\cellheight]{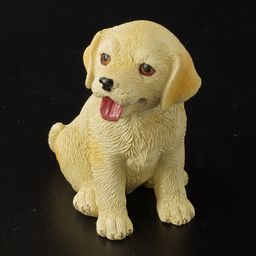}
        \end{subfigure} &
        \begin{subfigure}[t]{\cellwidth}
            \centering
            \includegraphics[height=\cellheight]{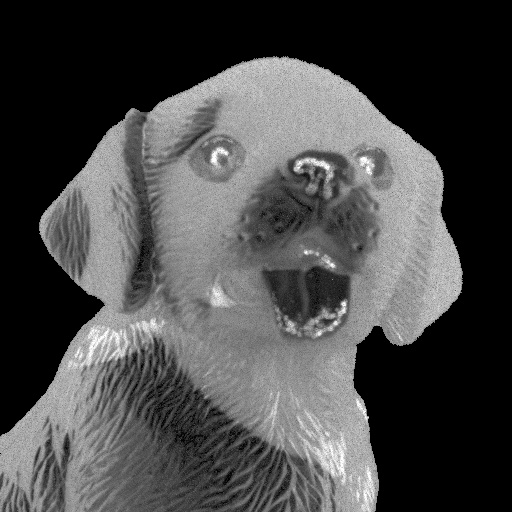}
        \end{subfigure} &
        \begin{subfigure}[t]{\cellwidth}
            \centering
            \includegraphics[height=\cellheight]{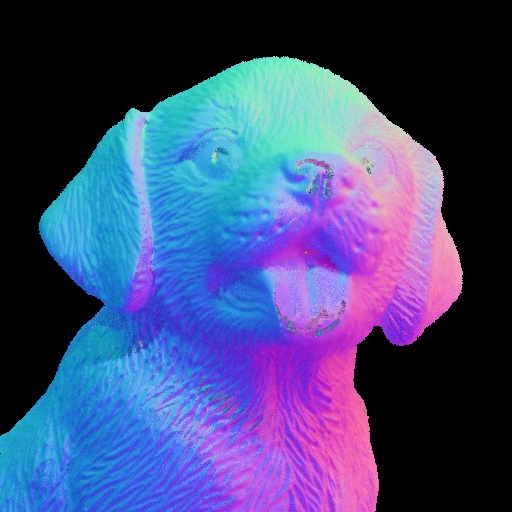}
        \end{subfigure} &
        \begin{subfigure}[t]{\cellwidth}
            \centering
            \includegraphics[height=\cellheight]{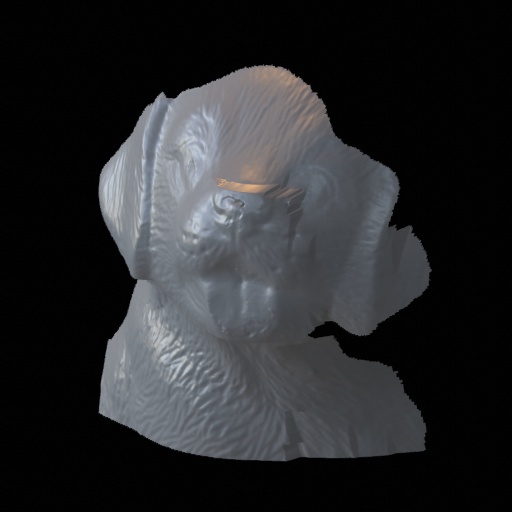}
        \end{subfigure} &
        \begin{subfigure}[t]{\cellwidth}
            \centering
            \includegraphics[height=\cellheight]{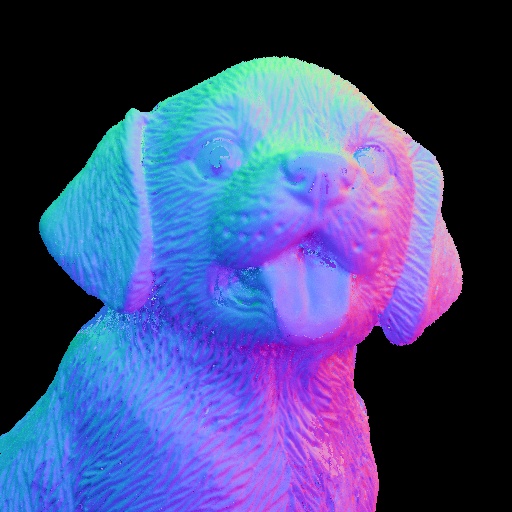}
        \end{subfigure} &
        \begin{subfigure}[t]{\cellwidth}
            \centering
            \includegraphics[height=\cellheight]{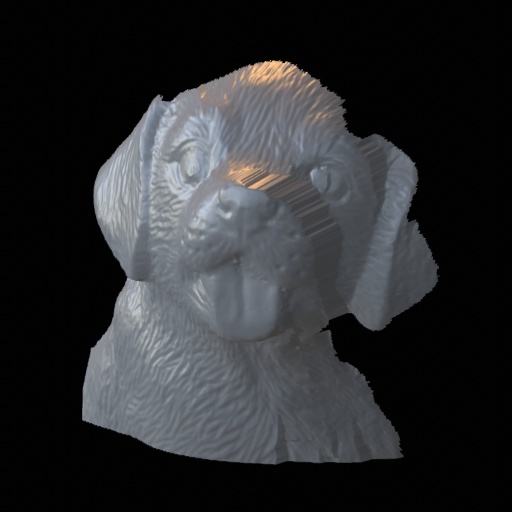}
        \end{subfigure} \\
        %%%%%%%%%%%%%%%%%%%%%%%%%%%%%%%%%%%%%%%%%%%%%%%%%%
    
        %%%%%%%%%%%%%%%%%%%%%%%%%%%%%%%%%%%%%%%%%%%%%%%%%%
        \rotatebox[origin=l]{90}{\cactusA} &
        \begin{subfigure}[t]{\cellwidth}
            \centering
            \includegraphics[height=\cellheight]{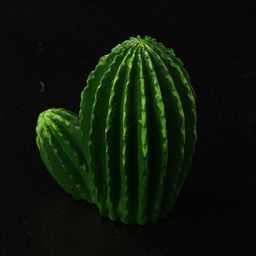}
        \end{subfigure} &
        \begin{subfigure}[t]{\cellwidth}
            \centering
            \includegraphics[height=\cellheight]{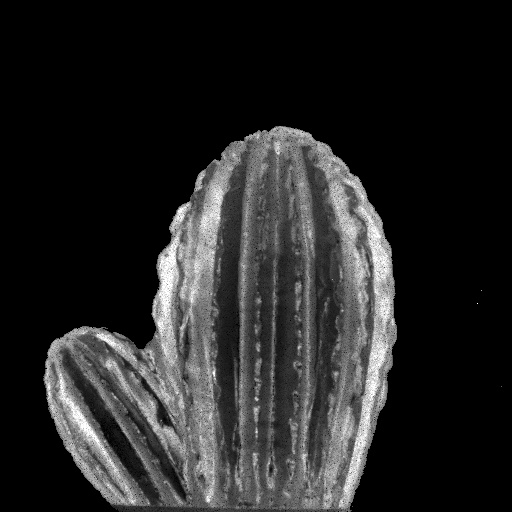}
        \end{subfigure} &
        \begin{subfigure}[t]{\cellwidth}
            \centering
            \includegraphics[height=\cellheight]{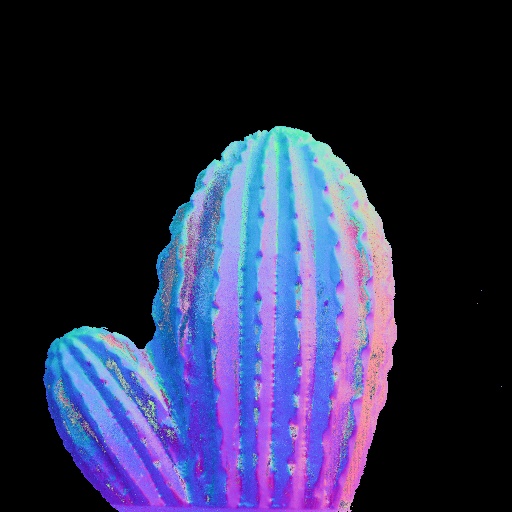}
        \end{subfigure} &
        \begin{subfigure}[t]{\cellwidth}
            \centering
            \includegraphics[height=\cellheight]{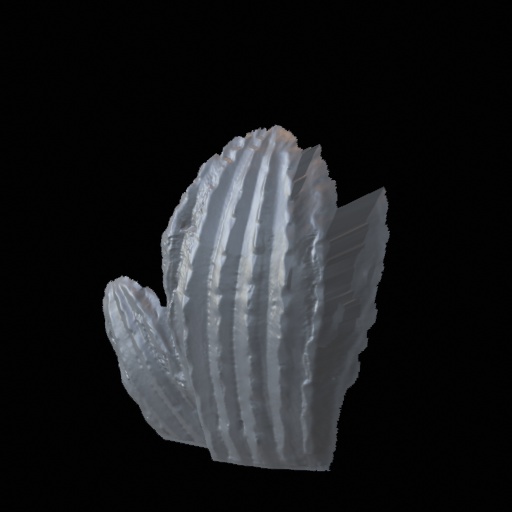}
        \end{subfigure} &
        \begin{subfigure}[t]{\cellwidth}
            \centering
            \includegraphics[height=\cellheight]{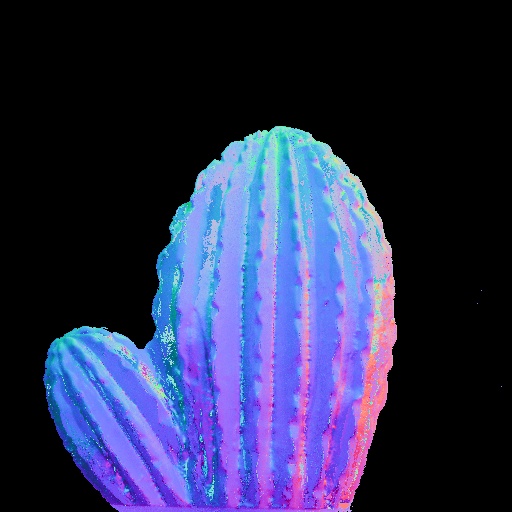}
        \end{subfigure} &
        \begin{subfigure}[t]{\cellwidth}
            \centering
            \includegraphics[height=\cellheight]{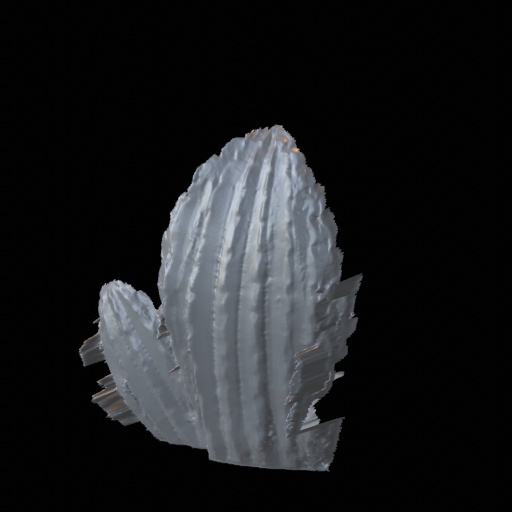}
        \end{subfigure} \\
        %%%%%%%%%%%%%%%%%%%%%%%%%%%%%%%%%%%%%%%%%%%%%%%%%%
    
        %%%%%%%%%%%%%%%%%%%%%%%%%%%%%%%%%%%%%%%%%%%%%%%%%%
        \rotatebox[origin=l]{90}{\medalcat} &
        \begin{subfigure}[t]{\cellwidth}
            \centering
            \includegraphics[height=\cellheight]{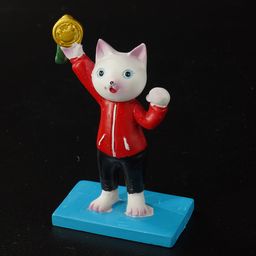}
        \end{subfigure} &
        \begin{subfigure}[t]{\cellwidth}
            \centering
            \includegraphics[height=\cellheight]{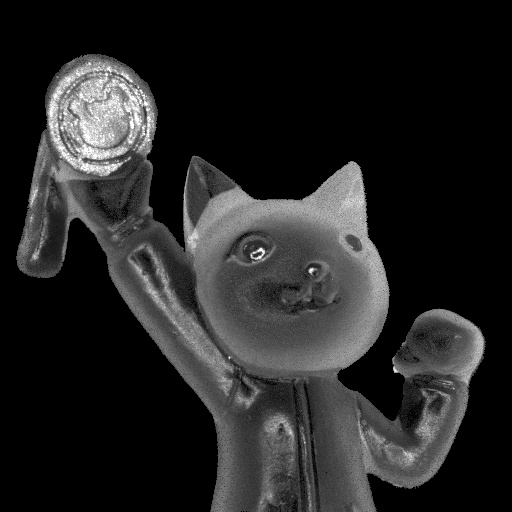}
        \end{subfigure} &
        \begin{subfigure}[t]{\cellwidth}
            \centering
            \includegraphics[height=\cellheight]{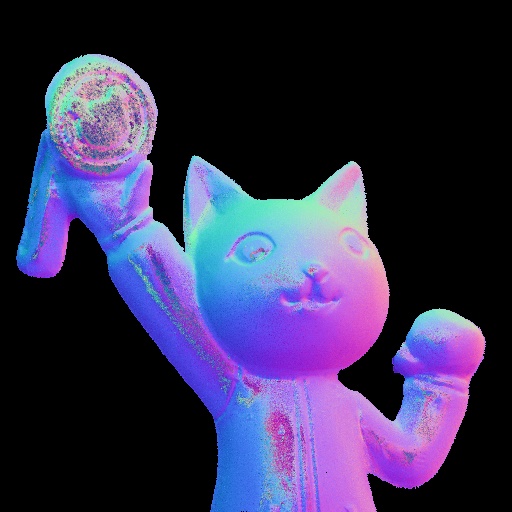}
        \end{subfigure} &
        \begin{subfigure}[t]{\cellwidth}
            \centering
            \includegraphics[height=\cellheight]{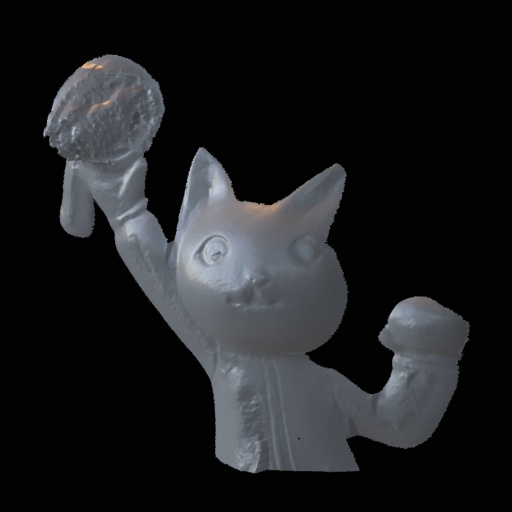}
        \end{subfigure} &
        \begin{subfigure}[t]{\cellwidth}
            \centering
            \includegraphics[height=\cellheight]{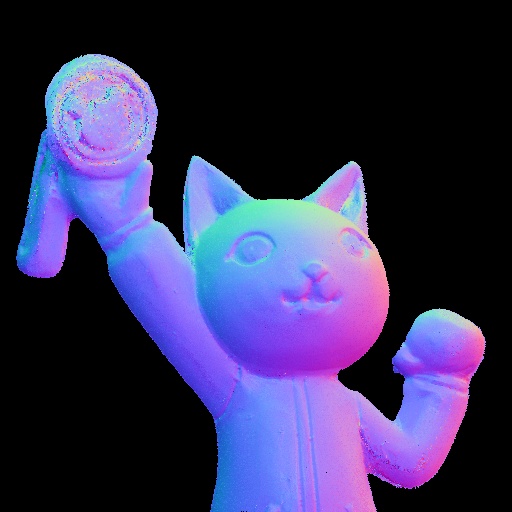}
        \end{subfigure} &
        \begin{subfigure}[t]{\cellwidth}
            \centering
            \includegraphics[height=\cellheight]{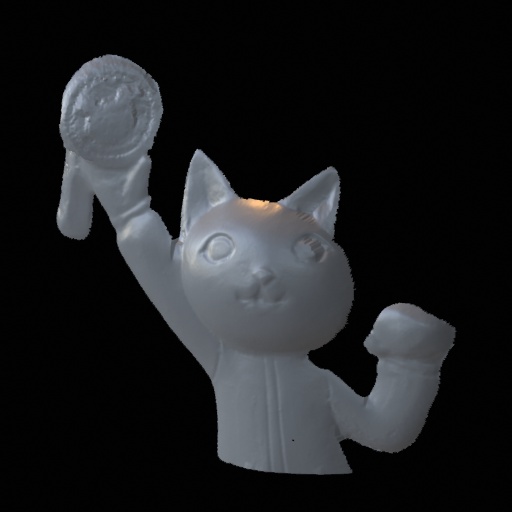}
        \end{subfigure} \\
        %%%%%%%%%%%%%%%%%%%%%%%%%%%%%%%%%%%%%%%%%%%%%%%%%%
         
    \end{tabular*}
    
    \caption{
    Qualitative evaluation with the practical objects. From left to right: photograph, event accumulation image from events, normal map recovered by EventPS-FCN~\cite{EventPS} and its depth~\cite{bini2022cao}, and those by ours.}
    \label{fig:eval practical}
\end{figure}

Next, five complex shapes were measured for deeper analysis, as shown in~\cref{fig:eval complicated}. Including five additional shapes in the supplementary material, the average MAEs for EventPS and our method were \SI{15.64}{\degree} and \SI{9.66}{\degree}, respectively. Although areas under the jaws of \caesar and \baboon can be occluded and affected by cast shadows, the proposed method reduces these effects effectively. A notable advantage of the proposed method over EventPS is observed for flat-like shapes, such as \cv and \floral, likely due to the insufficient data of flat objects in the training dataset. However, flat objects remain challenging for the proposed method as well, due to the limited number of recorded events, as seen in accumulation images like \heatmap. Additionally, inter-reflections are not managed by the proposed method, leading to higher errors in concave regions, such as the legs of \bunny.

%The mask for the cast shadows works well, for example, in the areas around nose and under jaw of \caesar. On the other hand, some areas like foot of \bunny and under supraorbital ridges of \caesar and \baboon still have relatively large errors. This is because of inter-reflection which is not treated in the principle. For flat-like shapes, \eg, \floral, even detailed edges are recovered but not so sharp as the ground truth. This may be because of the collapsed events and also the quality of 3D printed objects, whose surface is smoothed in a curing step. Finally, the totally MAE is resulted in \SI{10.00}{\degree}, and thus the proposed method is comparable with the state-of-the-art photometric stereo, such as~\cite{Li2022a,Logothetis2021,chen2020}.

\subsection{Qualitative Evaluation}

For more practical cases, four commercially available objects were measured. Since the shapes of these objects are unknown, depth was reconstructed from the normal map using Bilateral Normal Integration~\cite{bini2022cao}. The experimental results are shown in \cref{fig:eval practical}. The findings indicate that our method effectively manages specularity and glossiness compared to EventPS, as demonstrated with \miffy and \medalcat. Due to the pixel-by-pixel approach, our results preserve detailed features, such as the head of the \retriever. Even \cactusA, which has complex shapes, appears well-recovered with highly detailed depth. 
% Additional examples can be found in the supplementary material.

% It can be seen that the normal maps are not affected by their albedos.
% The surface of \miffy is glossy but the effect is almost reduced by masking.
% \retriever has a lot of cast shadows but seems to be recovered well.
% The structure of \cactusA is reconstructed in the depth.
% The areas with specularity can be also recovered, \eg, the gold medal of \medalcat.

\section{Discussions and Limitations}
\label{sec:discussion}

\paragraph{Measurement time.}
The maximum rotational frequency of the current prototype system is limited to \SI{2}{Hz} due to the serial communication among the $60$ LEDs. However, if $60$ separate control signals are prepared, it would be possible to implement a ring light with a rotational frequency exceeding \SI{10000}{Hz}. While EventPS~\cite{EventPS} has demonstrated a system with a DC motor operating at up to \SI{30}{Hz}, electrical control is much easier than mechanical control.

\paragraph{Event thresholds.}
In this paper, we assume constant event thresholds. However, it is well known that the threshold can be modeled as a statistical threshold~\cite{Rebecq2018}. The variation in the threshold significantly affects the reconstruction of event interval profiles. Currently, averaging multiple measurements helps suppress its effect, although it increases the duration of the measurements.

\paragraph{Offset light and arbitrary trajectory.}
Adding offset light is crucial to obtaining reliable event intervals. This could push the boundaries of photometric stereo, making it feasible under ambient illumination. To apply event-based photometric stereo to real-world scenes, the trajectory of the light must be arbitrary. However, photometric stereo methods using conventional cameras under arbitrary light conditions, such as in \cite{Ikehata2022}, present significant challenges.

\paragraph{Limitation.}
\redstart
While this work achieves robust reconstruction of Lambertian surfaces, there are two major limitations.
\color{black}
First, light is assumed to be distant for simplicity. However, this assumption may be expanded to near light, referring to near-light photometric stereo~\cite{liu2018near}. Second, inter-reflections and subsurface scattering are not considered. To address these effects, it may be necessary to apply a photometric decomposition approach, such as~\cite{Herbort2013}.

\section{Conclusion}
\label{sec:conclusion}

This paper presents EIP-PS, a robust event-based photometric stereo method. Unlike EventPS, which treats each event interval independently, EIP-PS uses a time-series profile to capture relationships between intervals. By considering the time derivative of logarithmic intensity, our approach identifies and removes regions affected by non-Lambertian effects, improving surface normal recovery. Experiments show that EIP-PS effectively recovers normal maps from various shapes and materials, even in the presence of noise and non-Lambertian reflections. The average MAEs for all 3D-printed objects was \SI{8.12}{\degree} for EIP-PS, in contract to EventPS, which resulted in \SI{13.66}{\degree}.

Our future work will address several limitations of the current method. First, we plan to relax the assumption of a distant light source to accommodate near-light conditions, similar to near-light photometric stereo methods~\cite{liu2018near}. Second, we aim to account for inter-reflections and subsurface scattering, which are not considered in the current approach. To address these effects, we intend to explore photometric decomposition techniques, such as those proposed by~\cite{Herbort2013}. Additionally, we are interested in integrating our method with deep-learning-based approaches to replace the empirically designed masks with adaptive ones.

% This paper proposed a robust event-based photometric stereo method called EIP-PS (Event-Interval-Profile-based Photometric Stereo). Unlike previous methods like EventPS, which treat each event interval independently, EIP-PS uses a time-series profile formed by event intervals to capture the relationship between them. By considering the time derivative of logarithmic intensity, our approach identifies and removes regions affected by non-Lambertian effects, leading to more accurate surface normal recovery. Experiments with a prototype system demonstrate the method's effectiveness in recovering normal maps of various shapes and materials, even in the presence of noise and non-Lambertian reflections. The average of MAEs for all the 3D-printed objects resulted in \SI{13.66}{\degree} and \SI{8.12}{\degree} for EventPS and ours, respectively.

% \clearpage
% \input{outline/o3_preliminaries}
% \input{outline/o4_method}
% \input{outline/o5_experiments}
% \input{outline/o9_implemantation}
% \input{outline/ox_figs}

{
    \small
    \bibliographystyle{ieeenat_fullname}
    \bibliography{main}
}

% \clearpage
\clearpage
\setcounter{page}{1}
\maketitlesupplementary

In this supplementary material, we provide implementation details and important results that couldn't be included in the main manuscript due to space constrains.

\section{Details of Data Acquisition}

In \cref{sec:implementation}, we discuss the implementation and calibration of the prototype system. In this section, we provide additional details about the implementation process.

\subsection{Bias Setting for Event Camera}

The current event cameras generally offer a range of adjustable settings that enable tuning the sensor performance to meet various application requirements and conditions, such as higher speed, lower background activity, and a higher contrast sensitivity threshold. These customizable sensor settings are referred to as {\it bias setting}.

\begin{table}[b]
    \newcommand{\colwidth}{0.13\linewidth}
    \newcolumntype{P}[1]{>{\centering\arraybackslash}p{#1}}
    \centering
    \footnotesize
    
    \caption{Bias setting. }
    \label{tab:biases}
    \vspace{-2ex}
    \begin{tabular*}{\linewidth}{@{\extracolsep{\fill}}c|P{\colwidth}P{\colwidth}P{\colwidth}P{\colwidth}P{\colwidth}}
         \hline
         Bias & Positive threshold & Negative threshold & \mbox{Low-pass} filter & \mbox{High-pass} filter & Dead time \\
         \hline
         Default & $112$ & $52$ & $23$ & $48$ & $45$ \\
         Ours & $99$ & $57$ & $20$ & $0$ & $50$ \\
         \hline
    \end{tabular*}
\end{table}

The event camera used in this paper (Prophesee EVK Gen4) has five biases: positive and negative thresholds, low-pass and high-pass filters, and dead time. 
The biases for the positive and negative thresholds literally determine the thresholds \posthresh and \negthresh, respectively. In the proposed method, it is preferable for these biases to be as small as possible. However, decreasing the bias for the threshold tends to increase the deviation of the statistical threshold. Refer to Fig.~6 in~\cite{Lichtsteiner2008}. 
The bias for low-pass filter determines the maximum rate of change in intensity that can be detected. In the prototype system, we adjust this bias to ensure that no events derived from the PWM, which is used to control LEDs, are detected.
The bias for high-pass filter determines the minimum rate of change in intensity. In the prototype system, we set this bias as small as possible to cover all frequencies.
The bias for dead time determines the duration during which a pixel is unable to detect a new event after each event. In the proposed method, we prefer this bias to be small; however, a minimal dead time bias may result in an increased number of events exceeding the event recording rate. In the proposed problem, the spatio-temporal density of events is relatively high, compared to other applications, such as object tracking. Thus, we adjust this bias as small as possible to ensure that the profile reconstruction does not collapse.
Finally, we show the biases used in the experiments in \cref{tab:biases}, along with the default biases.

\subsection{Calibration Process}

In the calibration process, intrinsic parameters of the event camera is also estimated to remove lens distortion, similar to conventional cameras. After calibrating the event thresholds for all pixels, the intrinsic parameters are estimated from $10$ poses of an ArUco board under all the LEDs modulated by a triangular wave. Similar to the light trajectory calibration, event accumulation images are reconstructed to facilitate the camera calibration process.

\begin{figure}[b]
    \centering
    \begin{subfigure}{0.32\linewidth}
        \centering
        \includegraphics[width=\linewidth]{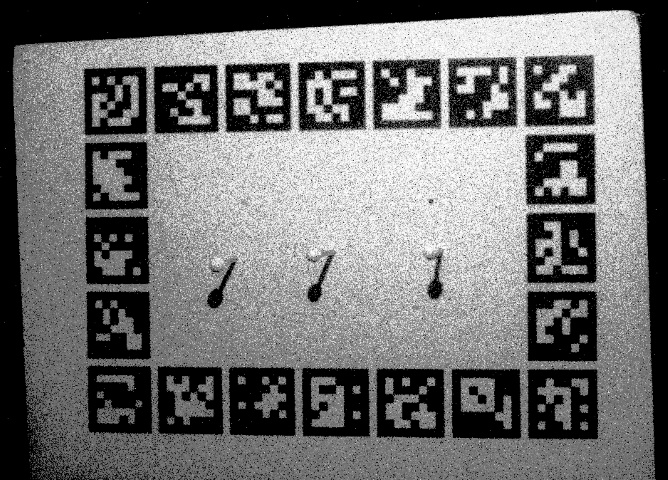}
        \caption{}    
    \end{subfigure}
    \begin{subfigure}{0.32\linewidth}
        \centering
        \includegraphics[width=\linewidth]{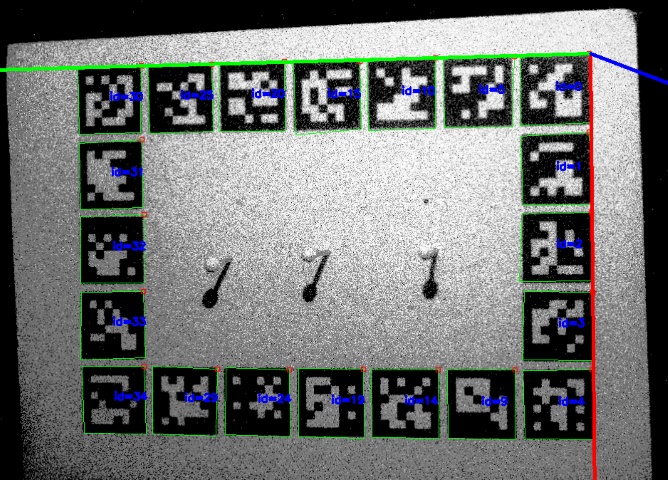}
        \caption{}    
    \end{subfigure}
    \begin{subfigure}{0.32\linewidth}
        \centering
        \includegraphics[width=\linewidth]{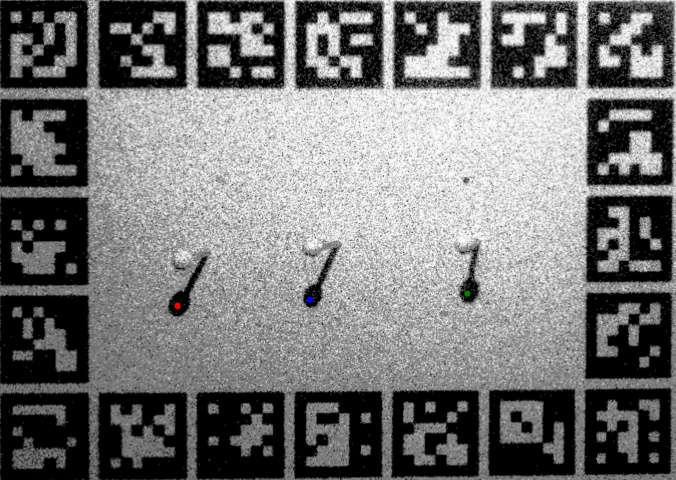}
        \caption{}    
    \end{subfigure}
    \caption{An example of the pattern detection in the calibration. (a)~Event accumulation image. (b)~ArUco detection. (c)~Pins' shadows detection.}
    \label{fig:calib_pattern}
    
    \vspace{2ex}
\end{figure}

As mentioned in the main manuscript, we apply Santo's method~\cite{Santo2020light} to estimate the positions of LEDs. Here, we provide an example of the pin shadow detection in \cref{fig:calib_pattern}. It is important to note that the offset light should be zero to detect any pattern based on albedos in an event accumulation image. 

\subsection{Effect of the Offset Light}

\begin{figure}[b]
    \centering
    \includegraphics[width=\linewidth]{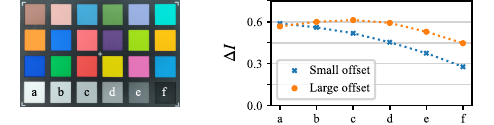}
    \caption{Effect of the offset light. The number of events with a larger offset is relatively equal for different color chips.}
    \label{fig:effect offset}
\end{figure}

In the main manuscript, we demonstrate how the offset light can reduce the albedo dependency in the event-generating mechanism of the prototype system. Here, we present pixel values in the event accumulation images for different color chips, as shown in \cref{fig:effect offset}. The pixel value indicates the number of recorded events. A small offset results in the number of events being strongly influenced by the albedos of the color chips, as depicted in the blue plot. Conversely, with a sufficiently large offset, the number of events becomes relatively equal for all the color chips, as shown in the orange plot, albeit with a decrease in the number of events for darker chips.

\redstart
In our experiments, the offset light intensity was empirically set to approximately 24~\si{\%} of the main light. While there is no theoretical optimal value due to device-specific properties, a rigorous calibration could be performed using a color chart, ensuring stable event generation irrespective of surface albedo. 
\color{black}

\subsection{Linear Interpolation of Reconstructed EIP}
Events occur when changes in intensity exceed a threshold, meaning the samples used for EIP reconstruction are not evenly spaced along the time axis. This results in a temporal distribution bias during fitting with the theoretical EIP, making the outcome overly dependent on specific samples rather than the overall shape of curve. Additionally, when capturing events over multiple lighting cycles, the sampling timing is not always consistent, leading to complications when integrating results across cycles. To address these issues, we adopted a simple interpolation approach. 

Specifically, we reconstructed a continuous EIP from sparse samples using the interpolation, then resampled it at predefined, evenly spaced time intervals to compare these points with the theoretical EIP. In our setup, the intervals between two consecutive events are generally small enough that the segment of the profile between them can be reasonably approximated as a linear function. This assumption simplifies the interpolation process, enabling to model each segment using a linear function. From this continuous representation, we uniformly sampled EIP values. 

\subsection{Examples of Reconstructed EIP}
\begin{figure}[b]
    \centering
    \begin{subfigure}[t]{\linewidth}
        \centering
        \includegraphics[width=\linewidth]{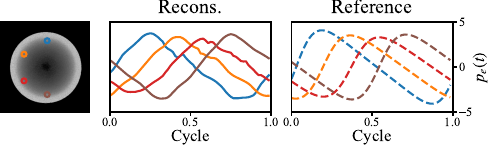}
        \caption{}
        \label{fig:profile character azimuth}
    \end{subfigure}
    \begin{subfigure}[t]{\linewidth}
        \centering
        \includegraphics[width=\linewidth]{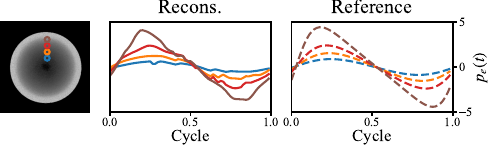}
        \caption{}
        \label{fig:profile character zenith}
    \end{subfigure}
    \caption{Shapes of the EIPs at various surface points. From left to right: surface points on \diffusesphere's \heatmap, reconstructed EIPs for each point, and theoretical EIPs (reference). (a)~Different azimuth angles under the identical zenith angle. (b)~Different zenith angles under the identical azimuth angle.}
    \label{fig:profile character}
\end{figure}

Here, we show some examples of the actual reconstructed EIPs obtained by our prototype system with \diffusesphere to visually demonstrate the actual shapes of the profiles. In \cref{fig:profile character azimuth}, we show both the reconstructed and theoretical EIPs at four surface points with varying azimuth angles, while keeping their zenith angles constant. The EIPs have similar overall shapes, but their phases are shifted according to the azimuth angles. In contrast, \cref{fig:profile character zenith} shows the reconstructed and theoretical EIPs at four different points with varying zenith angles, while keeping azimuth angles constant.
In this case, the amplitudes of the profiles vary depending on the zenith angles.

\subsection{Details of the Optimization Process}

\begin{figure}[t]
    \centering
    \includegraphics[width=\linewidth]{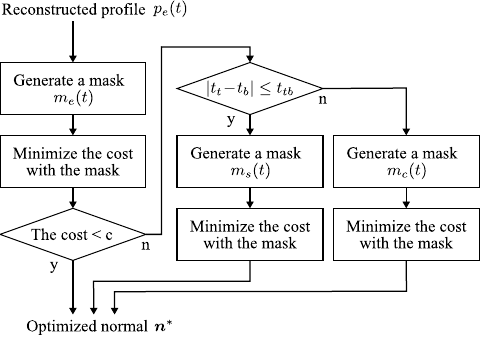}
    \caption{Process of the cost minimization.}
    \label{fig:cost minimization}
    \vspace{2ex}
\end{figure}
\begin{figure}[t]
    % Settings
    \newcommand{\cellwidth}{0.175\linewidth}
    \newcommand{\colorbarwidth}{0.038\linewidth}
    \newcommand{\costcolorbarwidth}{0.0425\linewidth}
    \newcommand{\cellheight}{\linewidth}
    \setlength\tabcolsep{0pt}
    \footnotesize
    \centering
        
    \begin{tabular*}{\linewidth}{@{\extracolsep{\fill}}cccccccc}
        %%%%%%%%%%%%%%%%%%%%%%%%%%%%%%%%%%%%%%%%%%%%%%%%%%
        & \multicolumn{4}{c}{\cellcolor{lightgray}1st minimization} & \multicolumn{3}{c}{\cellcolor{pink}2nd minimization} \\
        & Cost & & Normal & AE & Normal & AE & \\
        %%%%%%%%%%%%%%%%%%%%%%%%%%%%%%%%%%%%%%%%%%%%%%%%%%
    
        %%%%%%%%%%%%%%%%%%%%%%%%%%%%%%%%%%%%%%%%%%%%%%%%%%
        \rotatebox[origin=l]{90}{\glossysphere} &
        \begin{subfigure}[t]{\cellwidth}
            \centering
            \includegraphics[height=\cellheight]{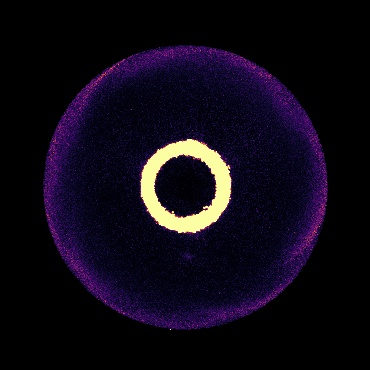}
        \end{subfigure} &
        \begin{subfigure}[t]{\costcolorbarwidth}
            \centering
            \includegraphics[width=\linewidth]{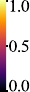}
        \end{subfigure} &
        \begin{subfigure}[t]{\cellwidth}
            \centering
            \includegraphics[height=\cellheight]{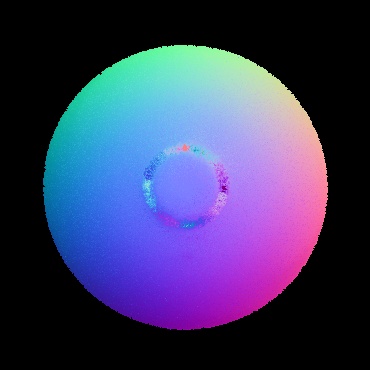}
        \end{subfigure} &
        \begin{subfigure}[t]{\cellwidth}
            \begin{picture}(\cellwidth, \cellheight)
                \put(0,0){\includegraphics[height=\cellheight]{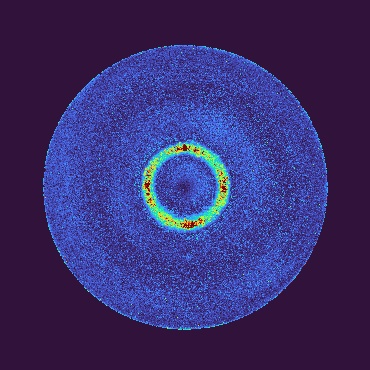}}
                \put(3,3){\mae{3.93}}
            \end{picture}
        \end{subfigure} &
        \begin{subfigure}[t]{\cellwidth}
            \centering
            \includegraphics[height=\cellheight]{figs/results/sphere_gloss/normals_obs.jpg}
        \end{subfigure} &
        \begin{subfigure}[t]{\cellwidth}
            \begin{picture}(\cellwidth, \cellheight)
                \put(0,0){\includegraphics[height=\cellheight]{figs/results/sphere_gloss/ang_error.jpg}}
                \put(3,3){\mae{3.16}}
            \end{picture}
        \end{subfigure} &
        \begin{subfigure}[t]{\colorbarwidth}
            \centering
            \includegraphics[width=\linewidth]{figs/results/colorbar1.pdf}
        \end{subfigure} \\
        %%%%%%%%%%%%%%%%%%%%%%%%%%%%%%%%%%%%%%%%%%%%%%%%%%
    
        %%%%%%%%%%%%%%%%%%%%%%%%%%%%%%%%%%%%%%%%%%%%%%%%%%
        \rotatebox[origin=l]{90}{\shadowsphere} &
        \begin{subfigure}[t]{\cellwidth}
            \centering
            \includegraphics[height=\cellheight]{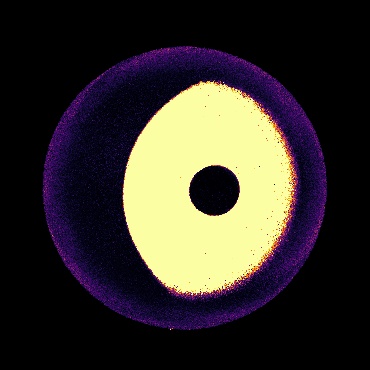}
        \end{subfigure} &
        \begin{subfigure}[t]{\costcolorbarwidth}
            \centering
            \includegraphics[width=\linewidth]{figs_supp/cost/inferno.pdf}
        \end{subfigure} &
        \begin{subfigure}[t]{\cellwidth}
            \centering
            \includegraphics[height=\cellheight]{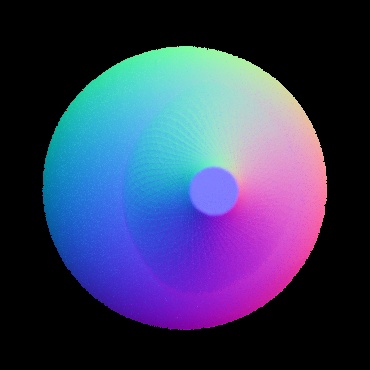}
        \end{subfigure} &
        \begin{subfigure}[t]{\cellwidth}
            \begin{picture}(\cellwidth, \cellheight)
                \put(0,0){\includegraphics[height=\cellheight]{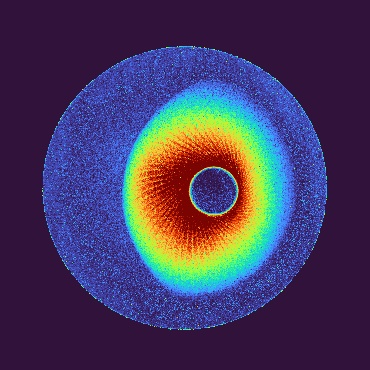}}
                \put(3,3){\mae{11.06}}
            \end{picture}
        \end{subfigure} &
        \begin{subfigure}[t]{\cellwidth}
            \centering
            \includegraphics[height=\cellheight]{figs/results/sphere_cast/normals_obs.jpg}
        \end{subfigure} &
        \begin{subfigure}[t]{\cellwidth}
            \begin{picture}(\cellwidth, \cellheight)
                \put(0,0){\includegraphics[height=\cellheight]{figs/results/sphere_cast/ang_error.jpg}}
                \put(3,3){\mae{3.71}}
            \end{picture}
        \end{subfigure} &
        \begin{subfigure}[t]{\colorbarwidth}
            \centering
            \includegraphics[width=\linewidth]{figs/results/colorbar1.pdf}
        \end{subfigure} \\
        %%%%%%%%%%%%%%%%%%%%%%%%%%%%%%%%%%%%%%%%%%%%%%%%%%

    \end{tabular*}
    
    \caption{Effect of the second mask process. From left to right: cost map, recovered normal map, and angular error map at the first minimization, and normal map and angular error map at the second minimization.}
    \label{fig:cost maps}
\end{figure}

The proposed method employs a multi-stage approach. Initially, optimization is performed without considering outliers such as shadows and specular reflections. Then, the presence of outliers is assessed, and a mask is generated accordingly for re-optimization. The process of the surface normal recovery is illustrated in \cref{fig:cost minimization}. Initially, the mask for collapsed events is applied to all surface points. If the minimized cost exceeds a threshold, \costthresh, the profile on that point may contain non-Lambertian effects. For instance, maps of costs after the first minimization are shown in the first column of \cref{fig:cost maps} when targeting \glossysphere and \shadowsphere. The recovered normal maps without the second mask process exhibit significant errors in surface points affected by the non-Lambertian effects.

In the second mask process, the mask selection is determined based on the temporal distance between the top and bottom peaks, denoted as $|\timetoppeak - \timebottompeak|$. For specular highlights, the temporal distance is typically very short, while cast shadows tend to have a relatively longer temporal effect on the surface point. Therefore, if the temporal distance exceeds a threshold, \durationtopbottom, the mask for cast shadows is applied; otherwise, the mask for specularity is used.

\redstart
We generate a single mask per outlier segment, regardless of the number of outlier types (\eg, multiple cast shadows, as \twopoles in Fig.~16). Empirically, multiple masks often reduce inlier observations, causing optimization instability. In future work, we will explore more robust methods, such as L1 loss or low-rank priors, to handle multiple outliers simultaneously.
\color{black}

\redstart
\subsection{Time Consumption}
Though not our top priority, computational speed remains important.
Our current implementation has larger time consumption compared to EventPS in terms of measurements and inferences.

The total measurement time is determined by the time for the light source rotation and the number of measurements, which can be reduced as suggested in \cref{fig:freq with correction} and \cref{fig:avg angular error}, respectively.
If future event cameras demonstrate improved fidelity to the event model, it would become possible to achieve a higher rotation frequency and require fewer measurements.

The long inference time is due to the complexity of our pipeline.
Algorithmically, the full pipeline (\ie, EIP reconstruction, mask generation, and surface normal optimization) performs at 0.07~\si{fps}. This is slower than EventPS-FCN (2~\si{fps}) but comparable to the pixel-based EventPS-CNN (0.1~\si{fps}). Notably, our method runs on a CPU, while the others use GPU-based implementations, suggesting that GPU parallelization could significantly boost performance.
\color{black}

%%%%%%%%%%%%%%%%%%%%%%%%%%%%%%%%%%%%%%%%%%%%%%%%%%
\section{Additional Experiments}

\subsection{Mask Margin for Non-Lambertian Effects}

In the second mask process, the margins of the masks related to specularity and cast shadows are considerable to enhance performance. Here, we analyze the performance of the proposed method in relation to these margins.

\begin{figure}[t]
    \centering
    \begin{subfigure}[t]{0.49\linewidth}
        \centering
        \includegraphics[width=\linewidth]{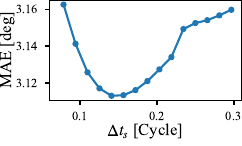}
        \caption{}
        \label{fig:margin gloss}
    \end{subfigure}
    \begin{subfigure}[t]{0.49\linewidth}
        \centering
        \includegraphics[width=\linewidth]{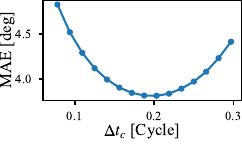}
        \caption{}
        \label{fig:margin cast}
    \end{subfigure}
    \caption{Analysis on the margins of masks. (a)~Mean angular error with respect to different \marginglossiness. (b)~Mean angular error with respect to different \margincast.}
    \label{fig:margins}
\end{figure}

For \glossysphere, we analyze the MAE for varying margins, \marginglossiness, as shown in \cref{fig:margin gloss}. The margin for specularity is changed in a range of $8$ to \SI{30}{\%} of the entire cycle. A small margin may allow specular lobes to still influence the profile, while a larger margin reduces the regions used for the cost minimization. In our prototype system, a margin of \SI{14}{\%} is identified as optimal for \marginglossiness.
Similarly, we analyze the margin for cast shadows, \margincast, for \shadowsphere, as shown in \cref{fig:margin cast}. In this case, a \SI{20}{\%} margin is found to be optimal for \margincast. 
However, since these findings are only based on \glossysphere and \shadowsphere, it is crucial to conduct further analysis of the margins using various objects with different types of specular characteristics and more complex shapes to ensure robustness. This aspect will be addressed in future work.

\redstart
\subsection{EventPS with Our Mask Mechanism}
\begin{figure}[!h]
    \centering    
    \vspace{-20pt}
    \includegraphics[trim=0 0 0 10,clip,width=0.99\linewidth]{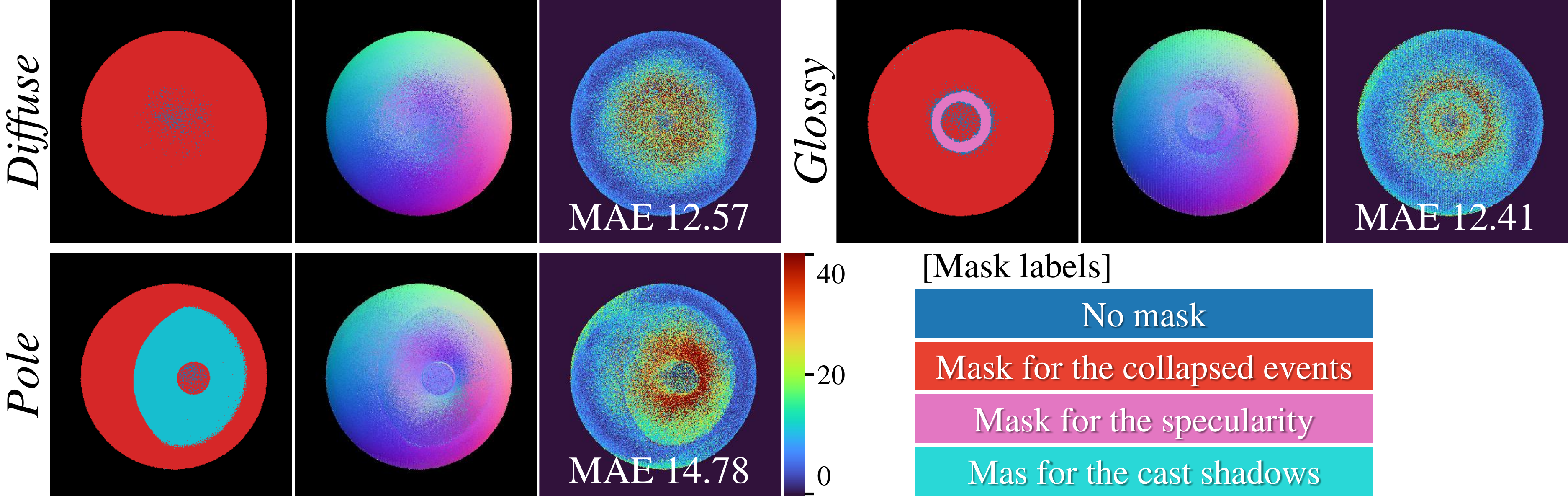}
    \caption{Quantitative evaluation of EventPS-FCN~\cite{EventPS} with events filtered by our masks. From left to right: applied mask labels, recovered normal map, and its angular error map.}
    \label{fig:eval with our mask}
\end{figure}

We evaluated the impact only on the removal of outlier regions by applying EventPS-FCN to event data (\diffusesphere, \glossysphere, \shadowsphere) masked by our method. Fig.~\ref{fig:eval with our mask} shows improvements over ~\cref{fig:eval spheres}, particularly in shadowed regions near edges and areas with specularities and cast shadows. However, the mean angular error (MAE) increased: \diffusesphere: $\SI{7.96}{\degree} \rightarrow \SI{12.57}{\degree}$, \glossysphere: $\SI{10.12}{\degree} \rightarrow \SI{12.41}{\degree}$, \shadowsphere: $\SI{12.02}{\degree} \rightarrow \SI{14.78}{\degree}$. This suggests that while filtering enhances robustness in non-Lambertian regions, it may reduce informative inliers, highlighting the need to refine the mask mechanism. The results also affirm PS-EIP’s superiority over EventPS-FCN for the same filtered events.
\color{black}

\subsection{Additional Quantitative Comparisons}
\begin{figure}[t]
    % Settings
    \newcommand{\cellwidth}{0.134\linewidth} % 11.390
    \newcommand{\colorbarwidth}{0.03\linewidth} % 0.185 / 16
    \newcommand{\cellheight}{\linewidth}
    \setlength\tabcolsep{0pt}
    % \footnotesize
    \scriptsize
    \centering
        
    \begin{tabular*}{\linewidth}{@{\extracolsep{\fill}}ccccccccc}
        %%%%%%%%%%%%%%%%%%%%%%%%%%%%%%%%%%%%%%%%%%%%%%%%%%
        & Photo & \heatmap & GT & \multicolumn{2}{c}{EventPS-FCN~\cite{EventPS}} & \multicolumn{2}{c}{Ours} \\
        %%%%%%%%%%%%%%%%%%%%%%%%%%%%%%%%%%%%%%%%%%%%%%%%%%

        %%%%%%%%%%%%%%%%%%%%%%%%%%%%%%%%%%%%%%%%%%%%%%%%%%
        \rotatebox[origin=l]{90}{\twocolorglossysphere} &
        \begin{subfigure}[t]{\cellwidth}
            \centering
            \includegraphics[height=\cellheight]{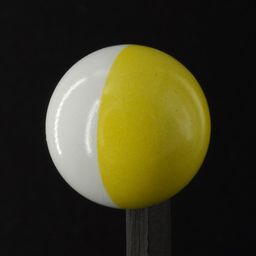}
        \end{subfigure} &
        \begin{subfigure}[t]{\cellwidth}
            \centering
            \includegraphics[height=\cellheight]{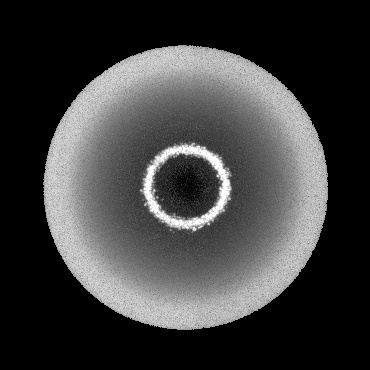}
        \end{subfigure} &
        \begin{subfigure}[t]{\cellwidth}
            \centering
            \includegraphics[height=\cellheight]{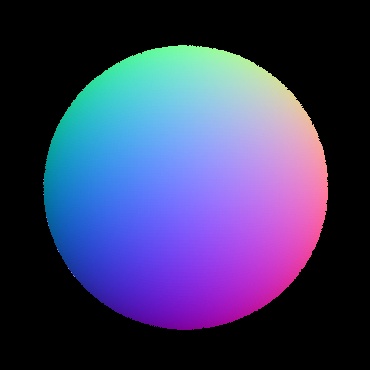}
        \end{subfigure} &
        \begin{subfigure}[t]{\cellwidth}
            \centering
            \includegraphics[height=\cellheight]{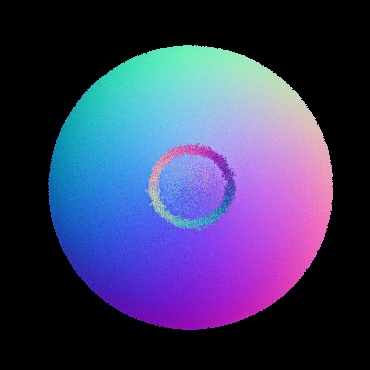}
        \end{subfigure} &
        \begin{subfigure}[t]{\cellwidth}
            \begin{picture}(\cellwidth, \cellheight)
                \put(0,0){\includegraphics[height=\cellheight]{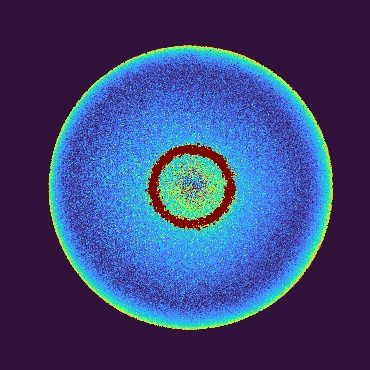}}
                \put(1,2){\mae{9.31}}
            \end{picture}
        \end{subfigure} &
        \begin{subfigure}[t]{\cellwidth}
            \centering
            \includegraphics[height=\cellheight]{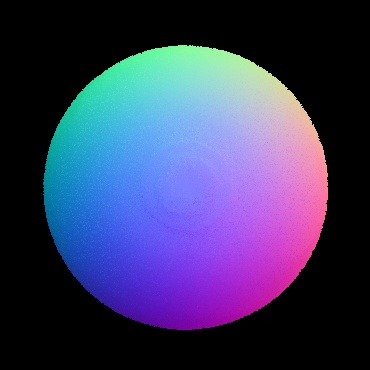}
        \end{subfigure} &
        \begin{subfigure}[t]{\cellwidth}
            \begin{picture}(\cellwidth, \cellheight)
                \put(0,0){\includegraphics[height=\cellheight]{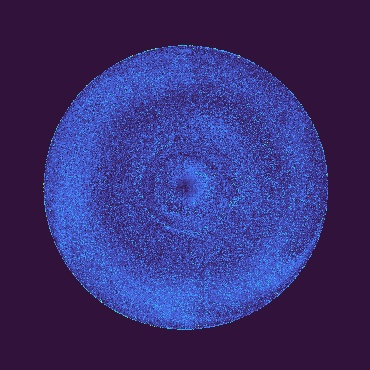}}
                \put(1,2){\mae{3.48}}
            \end{picture}
        \end{subfigure} &
        \begin{subfigure}[t]{\colorbarwidth}
            \centering
            \includegraphics[width=\linewidth]{figs/results/colorbar1.pdf}
        \end{subfigure} \\
        %%%%%%%%%%%%%%%%%%%%%%%%%%%%%%%%%%%%%%%%%%%%%%%%%%

        %%%%%%%%%%%%%%%%%%%%%%%%%%%%%%%%%%%%%%%%%%%%%%%%%%
        \rotatebox[origin=l]{90}{\diffuseyr} &
        \begin{subfigure}[t]{\cellwidth}
            \centering
            \includegraphics[height=\cellheight]{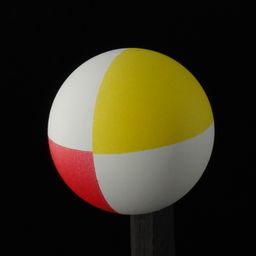}
        \end{subfigure} &
        \begin{subfigure}[t]{\cellwidth}
            \centering
            \includegraphics[height=\cellheight]{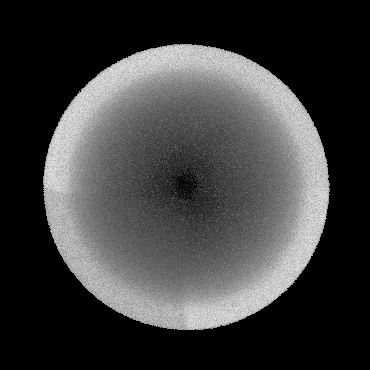}
        \end{subfigure} &
        \begin{subfigure}[t]{\cellwidth}
            \centering
            \includegraphics[height=\cellheight]{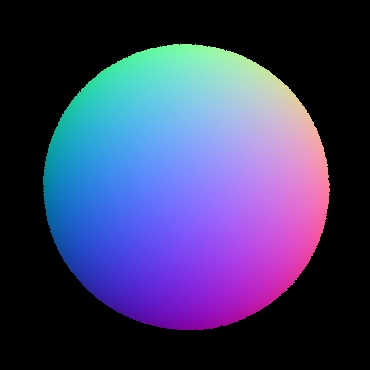}
        \end{subfigure} &
        \begin{subfigure}[t]{\cellwidth}
            \centering
            \includegraphics[height=\cellheight]{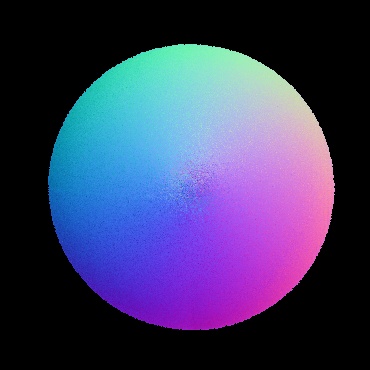}
        \end{subfigure} &
        \begin{subfigure}[t]{\cellwidth}
            \begin{picture}(\cellwidth, \cellheight)
                \put(0,0){\includegraphics[height=\cellheight]{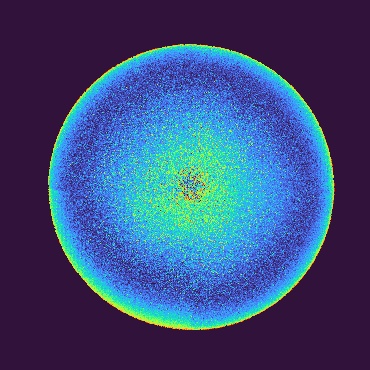}}
                \put(1,2){\mae{8.29}}
            \end{picture}
        \end{subfigure} &
        \begin{subfigure}[t]{\cellwidth}
            \centering
            \includegraphics[height=\cellheight]{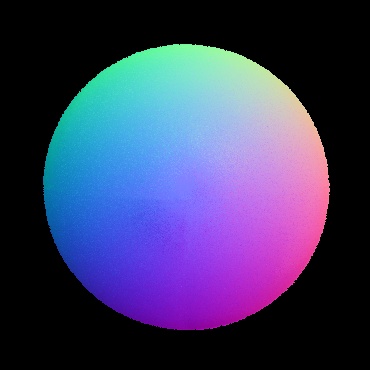}
        \end{subfigure} &
        \begin{subfigure}[t]{\cellwidth}
            \begin{picture}(\cellwidth, \cellheight)
                \put(0,0){\includegraphics[height=\cellheight]{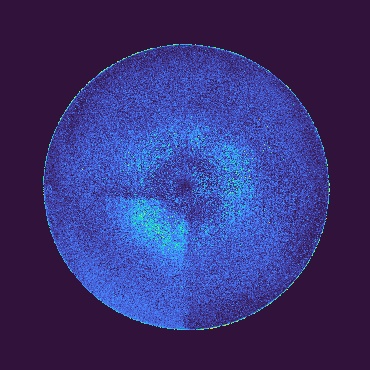}}
                \put(1,2){\mae{3.79}}
            \end{picture}
        \end{subfigure} &
        \begin{subfigure}[t]{\colorbarwidth}
            \centering
            \includegraphics[width=\linewidth]{figs/results/colorbar1.pdf}
        \end{subfigure} \\
        %%%%%%%%%%%%%%%%%%%%%%%%%%%%%%%%%%%%%%%%%%%%%%%%%%

        %%%%%%%%%%%%%%%%%%%%%%%%%%%%%%%%%%%%%%%%%%%%%%%%%%
        \rotatebox[origin=l]{90}{\diffusebg} &
        \begin{subfigure}[t]{\cellwidth}
            \centering
            \includegraphics[height=\cellheight]{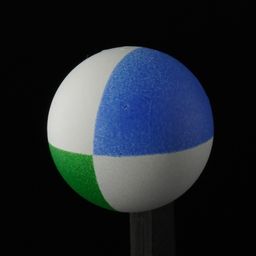}
        \end{subfigure} &
        \begin{subfigure}[t]{\cellwidth}
            \centering
            \includegraphics[height=\cellheight]{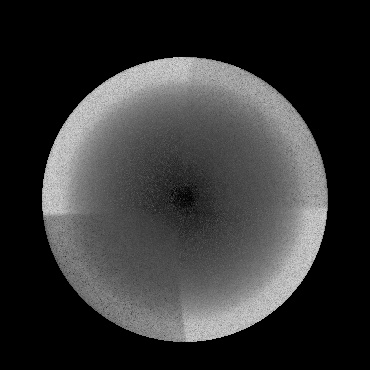}
        \end{subfigure} &
        \begin{subfigure}[t]{\cellwidth}
            \centering
            \includegraphics[height=\cellheight]{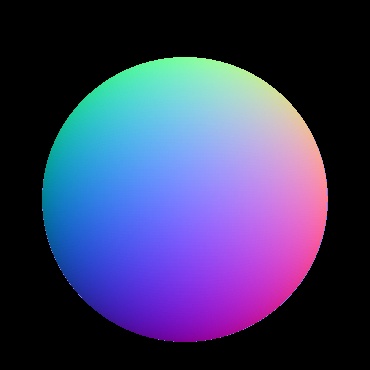}
        \end{subfigure} &
        \begin{subfigure}[t]{\cellwidth}
            \centering
            \includegraphics[height=\cellheight]{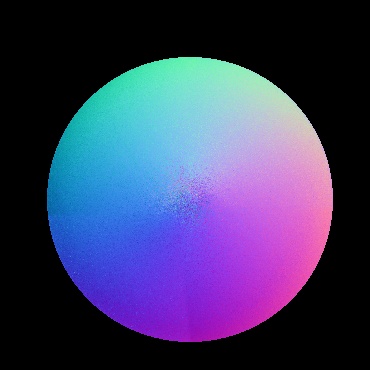}
        \end{subfigure} &
        \begin{subfigure}[t]{\cellwidth}
            \begin{picture}(\cellwidth, \cellheight)
                \put(0,0){\includegraphics[height=\cellheight]{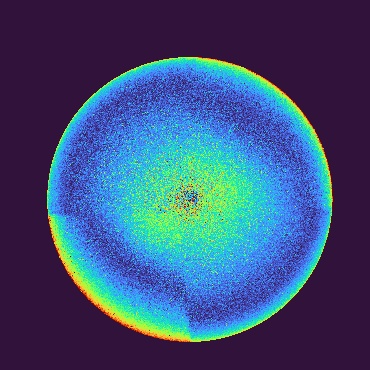}}
                \put(1,2){\mae{9.40}}
            \end{picture}
        \end{subfigure} &
        \begin{subfigure}[t]{\cellwidth}
            \centering
            \includegraphics[height=\cellheight]{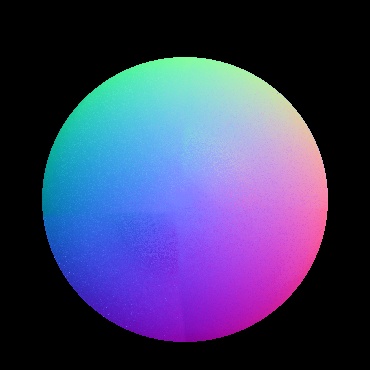}
        \end{subfigure} &
        \begin{subfigure}[t]{\cellwidth}
            \begin{picture}(\cellwidth, \cellheight)
                \put(0,0){\includegraphics[height=\cellheight]{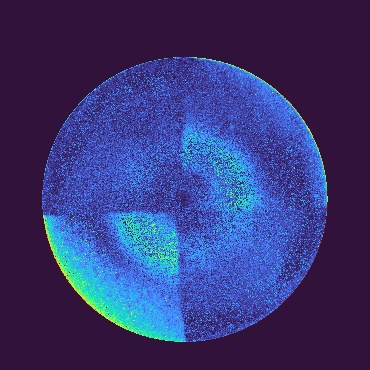}}
                \put(1,2){\mae{5.02}}
            \end{picture}
        \end{subfigure} &
        \begin{subfigure}[t]{\colorbarwidth}
            \centering
            \includegraphics[width=\linewidth]{figs/results/colorbar1.pdf}
        \end{subfigure} \\
        %%%%%%%%%%%%%%%%%%%%%%%%%%%%%%%%%%%%%%%%%%%%%%%%%%

        %%%%%%%%%%%%%%%%%%%%%%%%%%%%%%%%%%%%%%%%%%%%%%%%%%
        \rotatebox[origin=l]{90}{\semiglossy} &
        \begin{subfigure}[t]{\cellwidth}
            \centering
            \includegraphics[height=\cellheight]{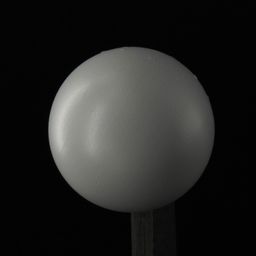}
        \end{subfigure} &
        \begin{subfigure}[t]{\cellwidth}
            \centering
            \includegraphics[height=\cellheight]{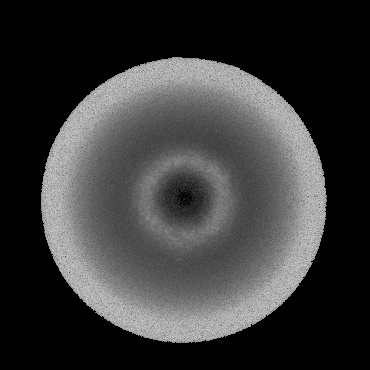}
        \end{subfigure} &
        \begin{subfigure}[t]{\cellwidth}
            \centering
            \includegraphics[height=\cellheight]{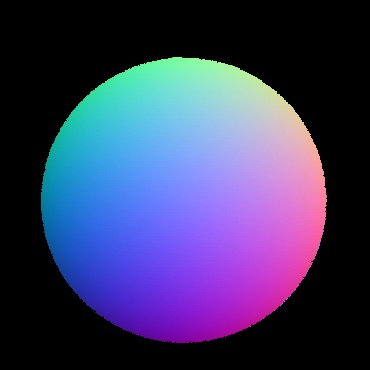}
        \end{subfigure} &
        \begin{subfigure}[t]{\cellwidth}
            \centering
            \includegraphics[height=\cellheight]{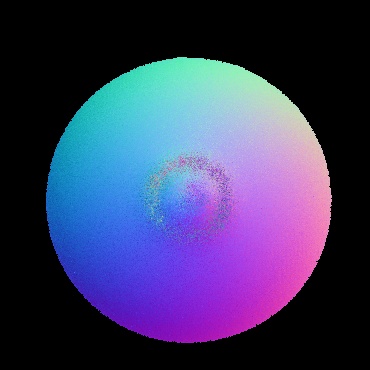}
        \end{subfigure} &
        \begin{subfigure}[t]{\cellwidth}
            \begin{picture}(\cellwidth, \cellheight)
                \put(0,0){\includegraphics[height=\cellheight]{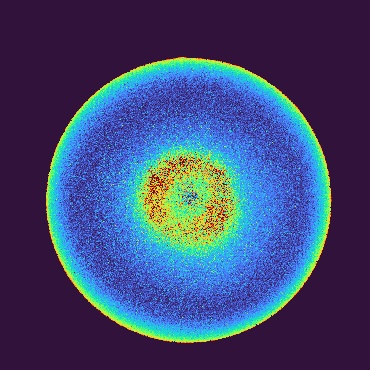}}
                \put(1,2){\mae{9.35}}
            \end{picture}
        \end{subfigure} &
        \begin{subfigure}[t]{\cellwidth}
            \centering
            \includegraphics[height=\cellheight]{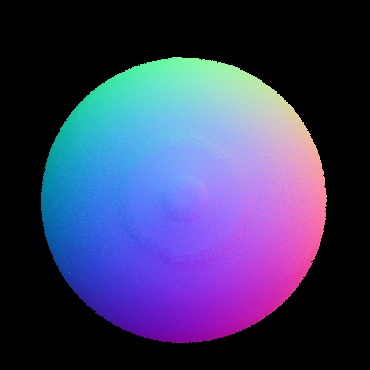}
        \end{subfigure} &
        \begin{subfigure}[t]{\cellwidth}
            \begin{picture}(\cellwidth, \cellheight)
                \put(0,0){\includegraphics[height=\cellheight]{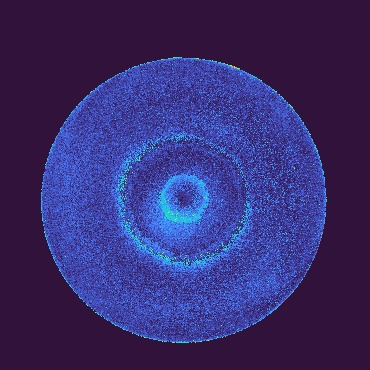}}
                \put(1,2){\mae{3.43}}
            \end{picture}
        \end{subfigure} &
        \begin{subfigure}[t]{\colorbarwidth}
            \centering
            \includegraphics[width=\linewidth]{figs/results/colorbar1.pdf}
        \end{subfigure} \\
        %%%%%%%%%%%%%%%%%%%%%%%%%%%%%%%%%%%%%%%%%%%%%%%%%%
    
        %%%%%%%%%%%%%%%%%%%%%%%%%%%%%%%%%%%%%%%%%%%%%%%%%%
        \rotatebox[origin=l]{90}{\cube} &
        \begin{subfigure}[t]{\cellwidth}
            \centering
            \includegraphics[height=\cellheight]{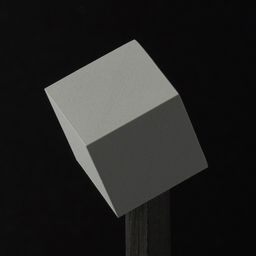}
        \end{subfigure} &
        \begin{subfigure}[t]{\cellwidth}
            \centering
            \includegraphics[height=\cellheight]{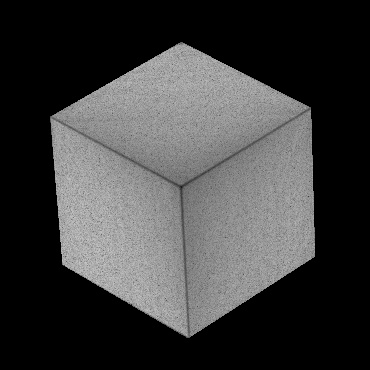}
        \end{subfigure} &
        \begin{subfigure}[t]{\cellwidth}
            \centering
            \includegraphics[height=\cellheight]{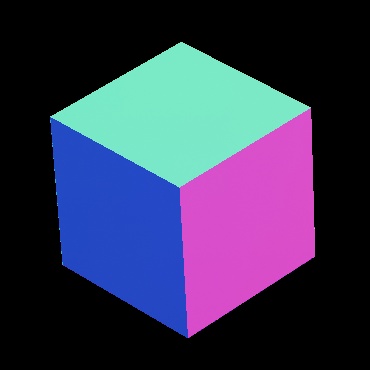}
        \end{subfigure} &
        \begin{subfigure}[t]{\cellwidth}
            \centering
            \includegraphics[height=\cellheight]{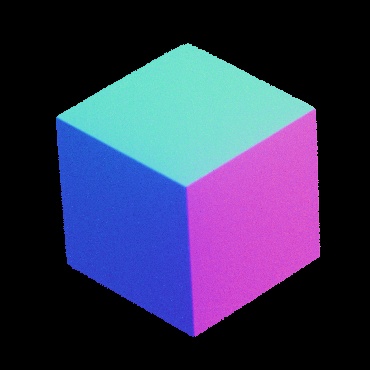}
        \end{subfigure} &
        \begin{subfigure}[t]{\cellwidth}
            \begin{picture}(\cellwidth, \cellheight)
                \put(0,0){\includegraphics[height=\cellheight]{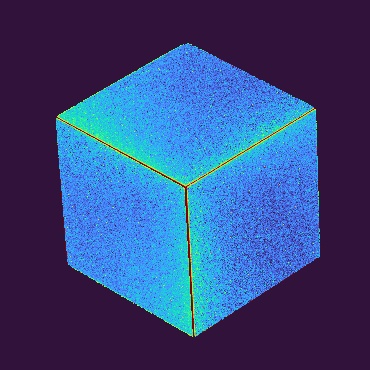}}
                \put(1,2){\mae{8.71}}
            \end{picture}
        \end{subfigure} &
        \begin{subfigure}[t]{\cellwidth}
            \centering
            \includegraphics[height=\cellheight]{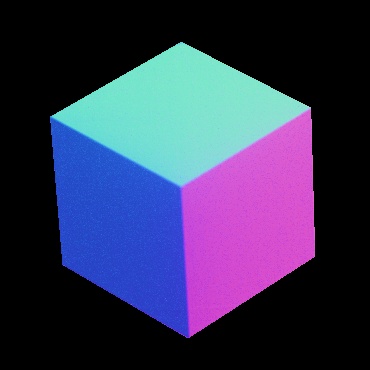}
        \end{subfigure} &
        \begin{subfigure}[t]{\cellwidth}
            \begin{picture}(\cellwidth, \cellheight)
                \put(0,0){\includegraphics[height=\cellheight]{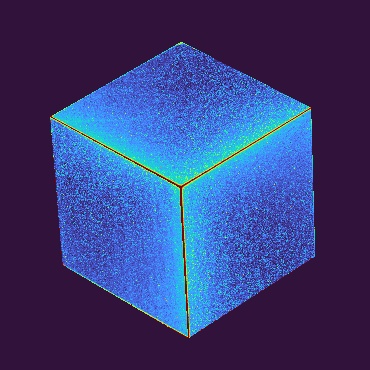}}
                \put(1,2){\mae{6.96}}
            \end{picture}
        \end{subfigure} &
        \begin{subfigure}[t]{\colorbarwidth}
            \centering
            \includegraphics[width=\linewidth]{figs/results/colorbar1.pdf}
        \end{subfigure} \\
        %%%%%%%%%%%%%%%%%%%%%%%%%%%%%%%%%%%%%%%%%%%%%%%%%%
    
        %%%%%%%%%%%%%%%%%%%%%%%%%%%%%%%%%%%%%%%%%%%%%%%%%%
        \rotatebox[origin=l]{90}{\brick} &
        \begin{subfigure}[t]{\cellwidth}
            \centering
            \includegraphics[height=\cellheight]{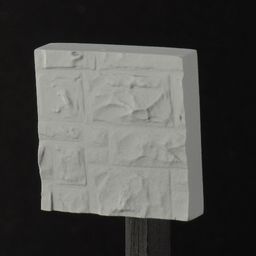}
        \end{subfigure} &
        \begin{subfigure}[t]{\cellwidth}
            \centering
            \includegraphics[height=\cellheight]{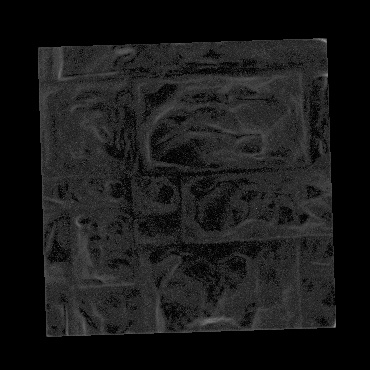}
        \end{subfigure} &
        \begin{subfigure}[t]{\cellwidth}
            \centering
            \includegraphics[height=\cellheight]{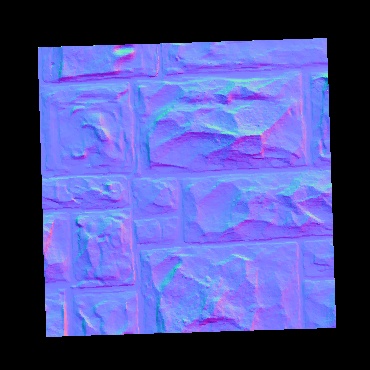}
        \end{subfigure} &
        \begin{subfigure}[t]{\cellwidth}
            \centering
            \includegraphics[height=\cellheight]{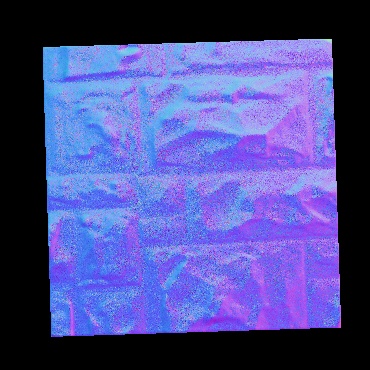}
        \end{subfigure} &
        \begin{subfigure}[t]{\cellwidth}
            \begin{picture}(\cellwidth, \cellheight)
                \put(0,0){\includegraphics[height=\cellheight]{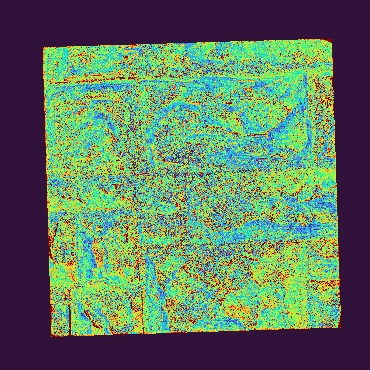}}
                \put(1,2){\mae{19.38}}
            \end{picture}
        \end{subfigure} &
        \begin{subfigure}[t]{\cellwidth}
            \centering
            \includegraphics[height=\cellheight]{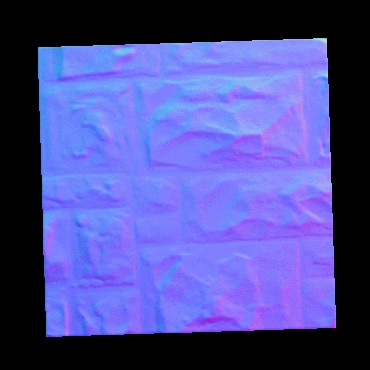}
        \end{subfigure} &
        \begin{subfigure}[t]{\cellwidth}
            \begin{picture}(\cellwidth, \cellheight)
                \put(0,0){\includegraphics[height=\cellheight]{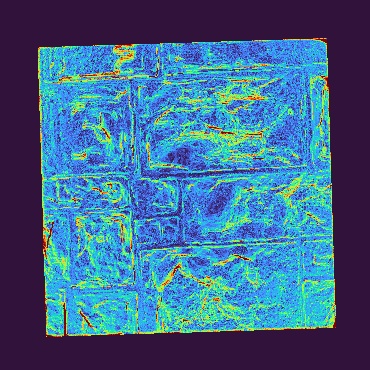}}
                \put(1,2){\mae{10.05}}
            \end{picture}
        \end{subfigure} &
        \begin{subfigure}[t]{\colorbarwidth}
            \centering
            \includegraphics[width=\linewidth]{figs/results/colorbar1.pdf}
        \end{subfigure} \\
        %%%%%%%%%%%%%%%%%%%%%%%%%%%%%%%%%%%%%%%%%%%%%%%%%%
    
        %%%%%%%%%%%%%%%%%%%%%%%%%%%%%%%%%%%%%%%%%%%%%%%%%%
        \rotatebox[origin=l]{90}{\glossycv} &
        \begin{subfigure}[t]{\cellwidth}
            \centering
            \includegraphics[height=\cellheight]{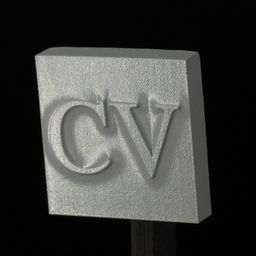}
        \end{subfigure} &
        \begin{subfigure}[t]{\cellwidth}
            \centering
            \includegraphics[height=\cellheight]{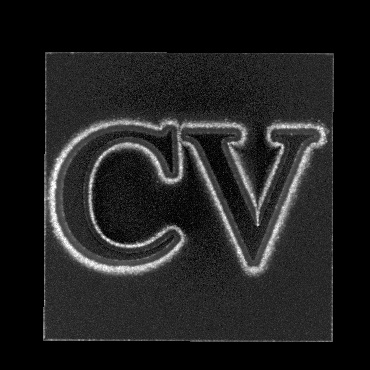}
        \end{subfigure} &
        \begin{subfigure}[t]{\cellwidth}
            \centering
            \includegraphics[height=\cellheight]{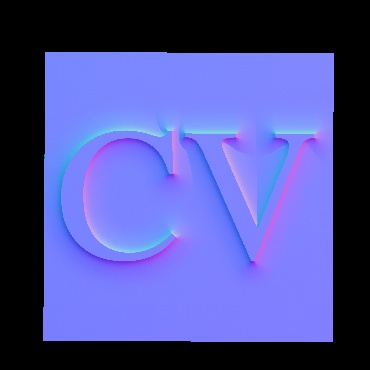}
        \end{subfigure} &
        \begin{subfigure}[t]{\cellwidth}
            \centering
            \includegraphics[height=\cellheight]{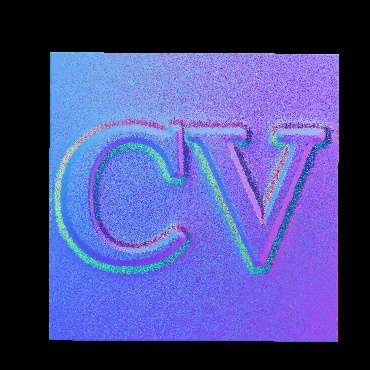}
        \end{subfigure} &
        \begin{subfigure}[t]{\cellwidth}
            \begin{picture}(\cellwidth, \cellheight)
                \put(0,0){\includegraphics[height=\cellheight]{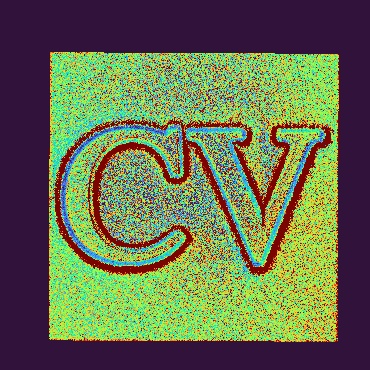}}
                \put(1,2){\mae{24.90}}
            \end{picture}
        \end{subfigure} &
        \begin{subfigure}[t]{\cellwidth}
            \centering
            \includegraphics[height=\cellheight]{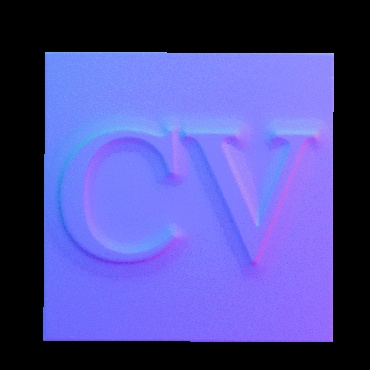}
        \end{subfigure} &
        \begin{subfigure}[t]{\cellwidth}
            \begin{picture}(\cellwidth, \cellheight)
                \put(0,0){\includegraphics[height=\cellheight]{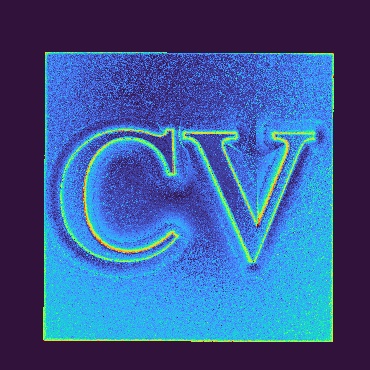}}
                \put(1,2){\mae{7.24}}
            \end{picture}
        \end{subfigure} &
        \begin{subfigure}[t]{\colorbarwidth}
            \centering
            \includegraphics[width=\linewidth]{figs/results/colorbar1.pdf}
        \end{subfigure} \\
        %%%%%%%%%%%%%%%%%%%%%%%%%%%%%%%%%%%%%%%%%%%%%%%%%%
    
        %%%%%%%%%%%%%%%%%%%%%%%%%%%%%%%%%%%%%%%%%%%%%%%%%%
        \rotatebox[origin=l]{90}{\multirefsbunny} &
        \begin{subfigure}[t]{\cellwidth}
            \centering
            \includegraphics[height=\cellheight]{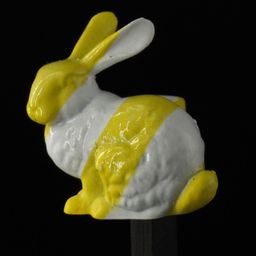}
        \end{subfigure} &
        \begin{subfigure}[t]{\cellwidth}
            \centering
            \includegraphics[height=\cellheight]{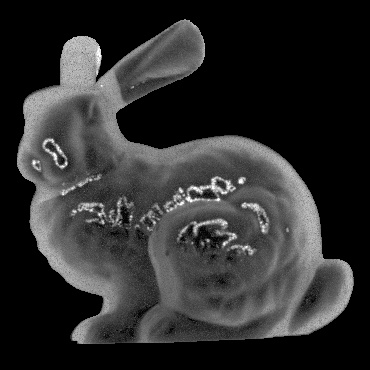}
        \end{subfigure} &
        \begin{subfigure}[t]{\cellwidth}
            \centering
            \includegraphics[height=\cellheight]{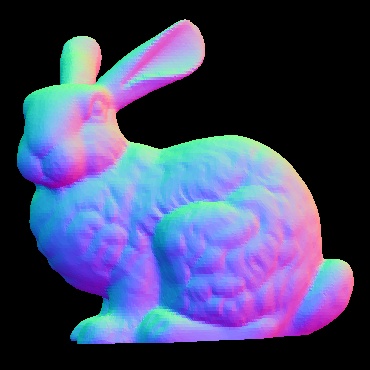}
        \end{subfigure} &
        \begin{subfigure}[t]{\cellwidth}
            \centering
            \includegraphics[height=\cellheight]{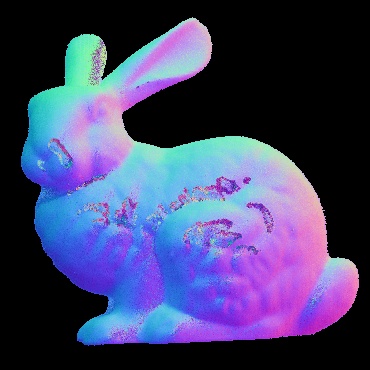}
        \end{subfigure} &
        \begin{subfigure}[t]{\cellwidth}
            \begin{picture}(\cellwidth, \cellheight)
                \put(0,0){\includegraphics[height=\cellheight]{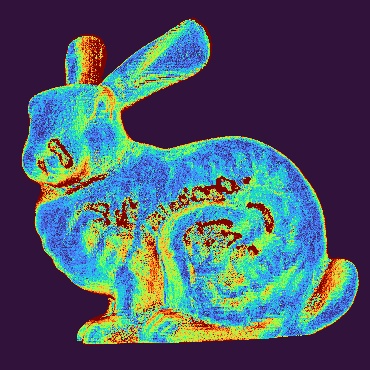}}
                \put(1,2){\mae{17.17}}
            \end{picture}
        \end{subfigure} &
        \begin{subfigure}[t]{\cellwidth}
            \centering
            \includegraphics[height=\cellheight]{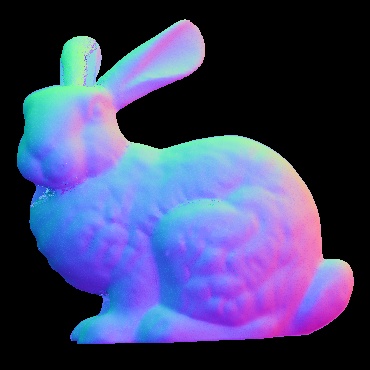}
        \end{subfigure} &
        \begin{subfigure}[t]{\cellwidth}
            \begin{picture}(\cellwidth, \cellheight)
                \put(0,0){\includegraphics[height=\cellheight]{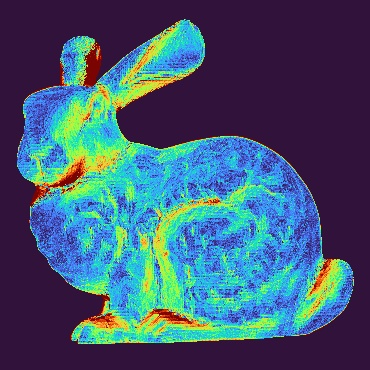}}
                \put(1,2){\mae{12.43}}
            \end{picture}
        \end{subfigure} &
        \begin{subfigure}[t]{\colorbarwidth}
            \centering
            \includegraphics[width=\linewidth]{figs/results/colorbar1.pdf}
        \end{subfigure} \\
        %%%%%%%%%%%%%%%%%%%%%%%%%%%%%%%%%%%%%%%%%%%%%%%%%%

        %%%%%%%%%%%%%%%%%%%%%%%%%%%%%%%%%%%%%%%%%%%%%%%%%%
        \rotatebox[origin=l]{90}{\twopoles} &
        \begin{subfigure}[t]{\cellwidth}
            \centering
            \includegraphics[height=\cellheight]{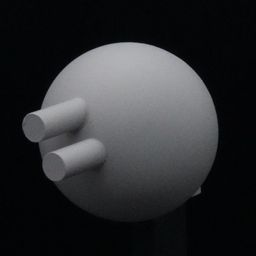}
        \end{subfigure} &
        \begin{subfigure}[t]{\cellwidth}
            \centering
            \includegraphics[height=\cellheight]{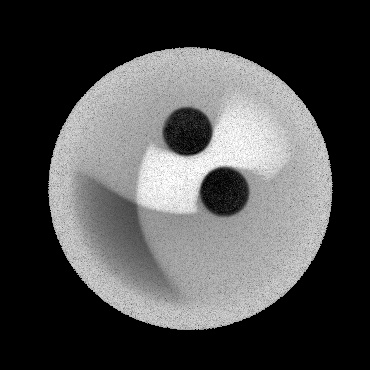}
        \end{subfigure} &
        \begin{subfigure}[t]{\cellwidth}
            \centering
            \includegraphics[height=\cellheight]{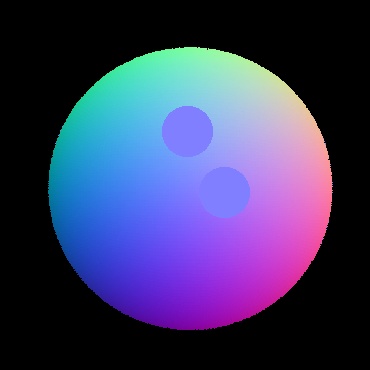}
        \end{subfigure} &
        \begin{subfigure}[t]{\cellwidth}
            \centering
            \includegraphics[height=\cellheight]{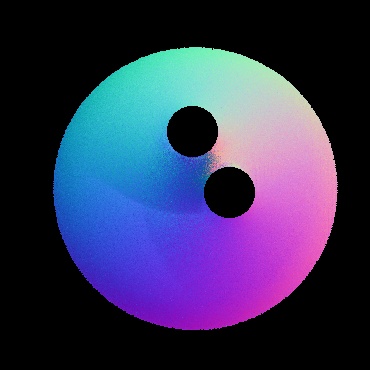}
        \end{subfigure} &
        \begin{subfigure}[t]{\cellwidth}
            \begin{picture}(\cellwidth, \cellheight)
                \put(0,0){\includegraphics[height=\cellheight]{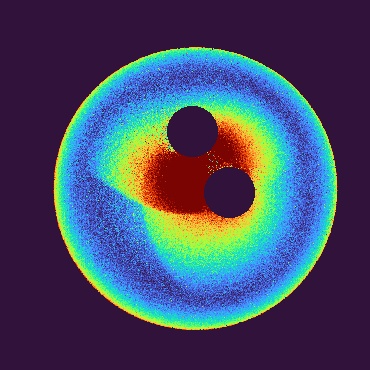}}
                \put(1,2){\mae{14.43}}
            \end{picture}
        \end{subfigure} &
        \begin{subfigure}[t]{\cellwidth}
            \centering
            \includegraphics[height=\cellheight]{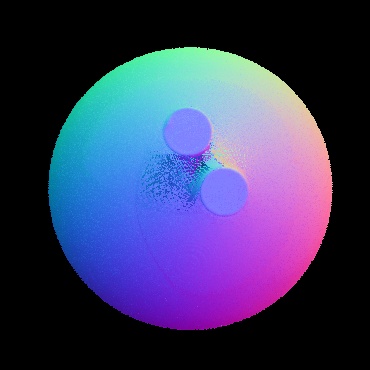}
        \end{subfigure} &
        \begin{subfigure}[t]{\cellwidth}
            \begin{picture}(\cellwidth, \cellheight)
                \put(0,0){\includegraphics[height=\cellheight]{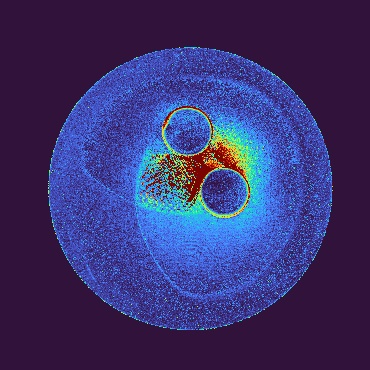}}
                \put(1,2){\mae{5.98}}
            \end{picture}
        \end{subfigure} &
        \begin{subfigure}[t]{\colorbarwidth}
            \centering
            \includegraphics[width=\linewidth]{figs/results/colorbar1.pdf}
        \end{subfigure} \\
        %%%%%%%%%%%%%%%%%%%%%%%%%%%%%%%%%%%%%%%%%%%%%%%%%%

        %%%%%%%%%%%%%%%%%%%%%%%%%%%%%%%%%%%%%%%%%%%%%%%%%%
        \rotatebox[origin=l]{90}{\blacksphere} &
        \begin{subfigure}[t]{\cellwidth}
            \centering
            \includegraphics[height=\cellheight]{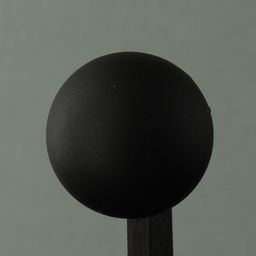}
        \end{subfigure} &
        \begin{subfigure}[t]{\cellwidth}
            \centering
            \includegraphics[height=\cellheight]{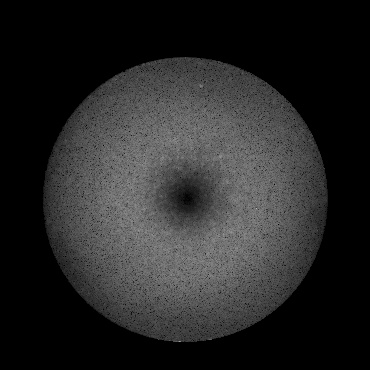}
        \end{subfigure} &
        \begin{subfigure}[t]{\cellwidth}
            \centering
            \includegraphics[height=\cellheight]{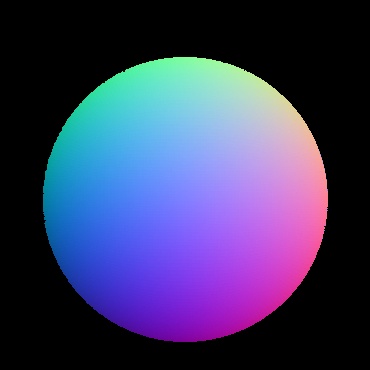}
        \end{subfigure} &
        \begin{subfigure}[t]{\cellwidth}
            \centering
            \includegraphics[height=\cellheight]{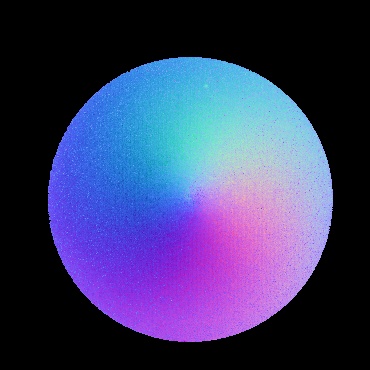}
        \end{subfigure} &
        \begin{subfigure}[t]{\cellwidth}
            \begin{picture}(\cellwidth, \cellheight)
                \put(0,0){\includegraphics[height=\cellheight]{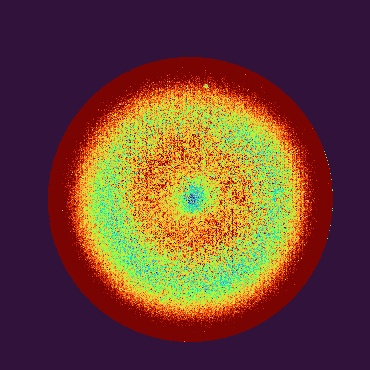}}
                \put(1,2){\mae{33.27}}
            \end{picture}
        \end{subfigure} &
        \begin{subfigure}[t]{\cellwidth}
            \centering
            \includegraphics[height=\cellheight]{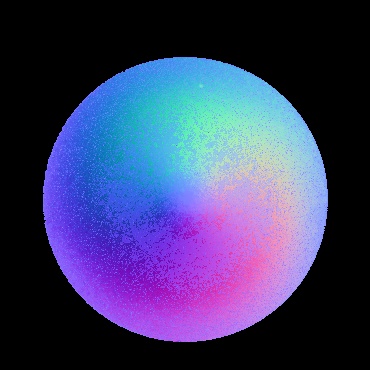}
        \end{subfigure} &
        \begin{subfigure}[t]{\cellwidth}
            \begin{picture}(\cellwidth, \cellheight)
                \put(0,0){\includegraphics[height=\cellheight]{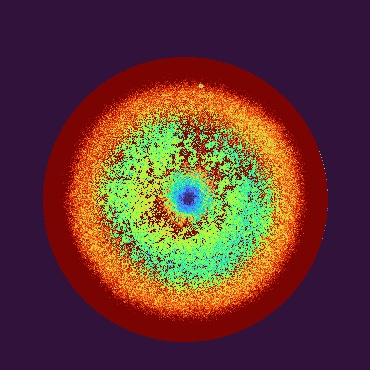}}
                \put(1,2){\mae{35.77}}
            \end{picture}
        \end{subfigure} &
        \begin{subfigure}[t]{\colorbarwidth}
            \centering
            \includegraphics[width=\linewidth]{figs/results/colorbar1.pdf}
        \end{subfigure} \\
        %%%%%%%%%%%%%%%%%%%%%%%%%%%%%%%%%%%%%%%%%%%%%%%%%%
    
    \end{tabular*}
    
    \caption{Additional, quantitative evaluation with 3D printed objects. From left to right: photograph, event accumulation image from the events, ground truth, normal map recovered by EventPS-FCN~\cite{EventPS} with its angular error map, and those by ours.}
    \label{fig:additional eval printed}
\end{figure}

We additionally targeted $10$ more 3D printed objects with different shapes and reflectances. The experimental results are shown in \cref{fig:additional eval printed}. \twocolorglossysphere is pained with two different colors and then given a glossy top coat. The MAE remains low, comparable to the case without the top coat, \twocolorsphere.
\diffuseyr and \diffusebg are 3-color spheres with different colors. Recovering normals for blue and green surfaces is more difficult compared to yellow and red surfaces. This difficulty may arise due to the lower radiance of these surfaces and the spectral sensitivity characteristics of the event camera.
Despite \semiglossy having wider specular lobes than \glossysphere, the mask for specularity functions effectively.
\multirefsbunny contains two-color surfaces, and its top half is glossy. The MAE is quite comparable to the gray, diffuse case of \bunny.

\twopoles poses a challenge as certain regions contain cast shadows from two distinct parts. The current mask for cast shadows assumes shadows derived from a single part in the scene. In future work, addressing such complex shapes will be a focus.
\blacksphere is also challenging to recover, much like conventional photometric stereo methods. As mentioned earlier, the principle does not work well when the radiance is low due to the log-amp characteristics.

\subsection{Additional Qualitative Comparisons}

\begin{figure}[t]
    % Settings
    \newcommand{\cellwidth}{0.162\linewidth}
    \newcommand{\cellheight}{\linewidth}
    \setlength\tabcolsep{0pt}
    % \footnotesize
    \scriptsize
    \centering

    \begin{tabular*}{\linewidth}{@{\extracolsep{\fill}}ccccccc}
        %%%%%%%%%%%%%%%%%%%%%%%%%%%%%%%%%%%%%%%%%%%%%%%%%%
        & Photo & \heatmap & \multicolumn{2}{c}{EventPS-FCN~\cite{EventPS}} & \multicolumn{2}{c}{Ours} \\
        %%%%%%%%%%%%%%%%%%%%%%%%%%%%%%%%%%%%%%%%%%%%%%%%%%

        %%%%%%%%%%%%%%%%%%%%%%%%%%%%%%%%%%%%%%%%%%%%%%%%%%
        \rotatebox[origin=l]{90}{\owl} &
        \begin{subfigure}[t]{\cellwidth}
            \centering
            \includegraphics[height=\cellheight]{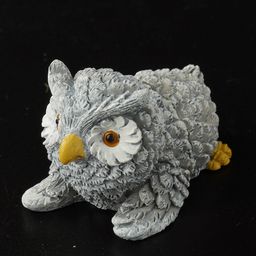}
        \end{subfigure} &
        \begin{subfigure}[t]{\cellwidth}
            \centering
            \includegraphics[height=\cellheight]{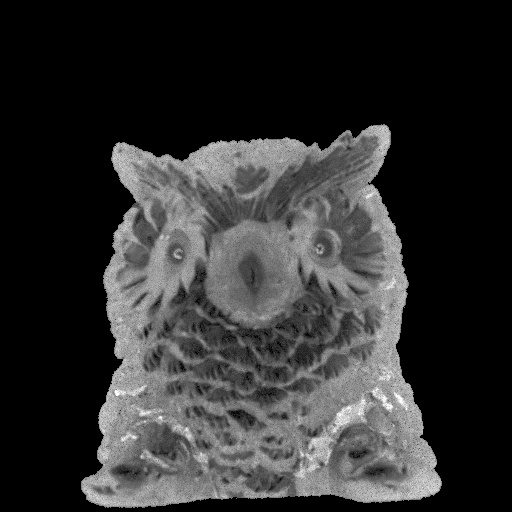}
        \end{subfigure} &
        \begin{subfigure}[t]{\cellwidth}
            \centering
            \includegraphics[height=\cellheight]{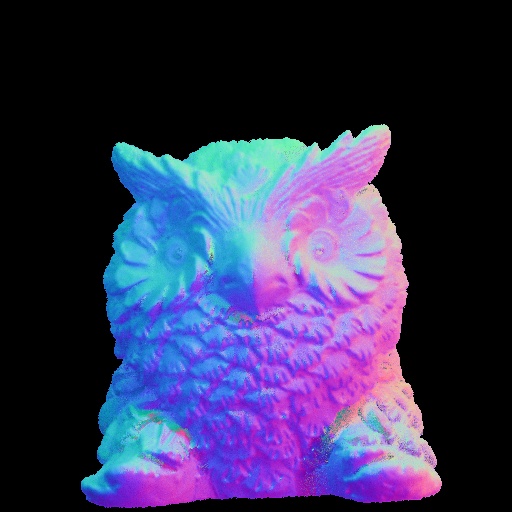}
        \end{subfigure} &
        \begin{subfigure}[t]{\cellwidth}
            \centering
            \includegraphics[height=\cellheight]{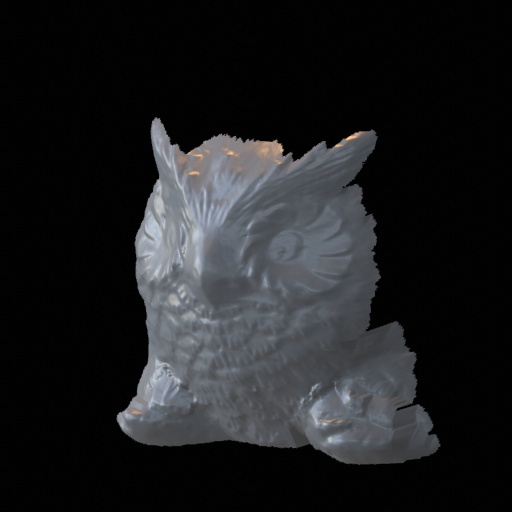}
        \end{subfigure} &
        \begin{subfigure}[t]{\cellwidth}
            \centering
            \includegraphics[height=\cellheight]{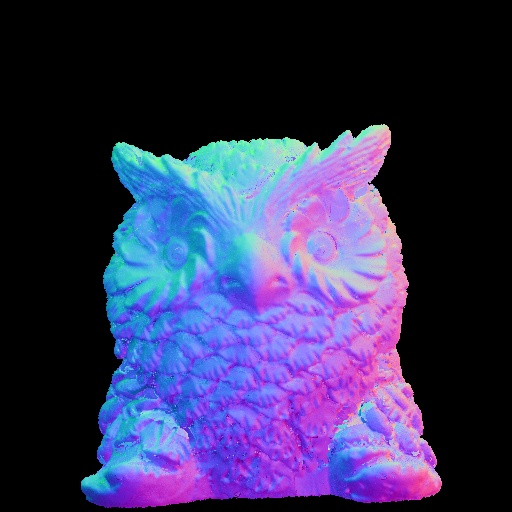}
        \end{subfigure} &
        \begin{subfigure}[t]{\cellwidth}
            \centering
            \includegraphics[height=\cellheight]{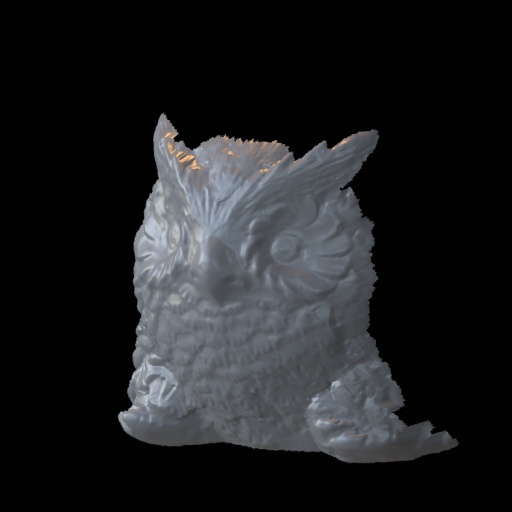}
        \end{subfigure} \\
        %%%%%%%%%%%%%%%%%%%%%%%%%%%%%%%%%%%%%%%%%%%%%%%%%%
    
        %%%%%%%%%%%%%%%%%%%%%%%%%%%%%%%%%%%%%%%%%%%%%%%%%%
        \rotatebox[origin=l]{90}{\whitedog} &
        \begin{subfigure}[t]{\cellwidth}
            \centering
            \includegraphics[height=\cellheight]{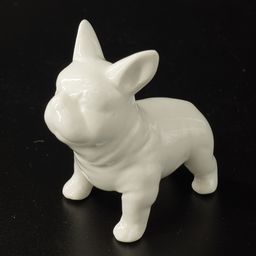}
        \end{subfigure} &
        \begin{subfigure}[t]{\cellwidth}
            \centering
            \includegraphics[height=\cellheight]{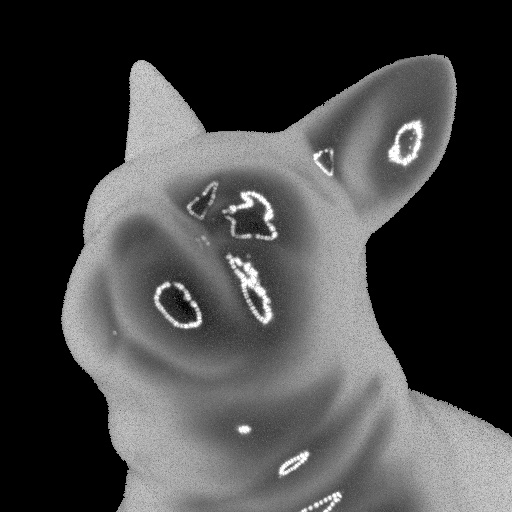}
        \end{subfigure} &
        \begin{subfigure}[t]{\cellwidth}
            \centering
            \includegraphics[height=\cellheight]{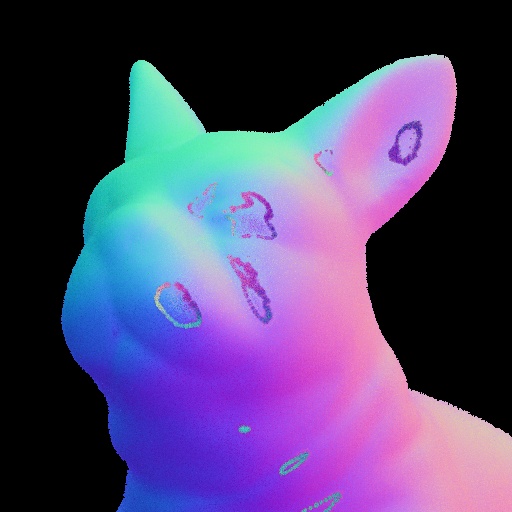}
        \end{subfigure} &
        \begin{subfigure}[t]{\cellwidth}
            \centering
            \includegraphics[height=\cellheight]{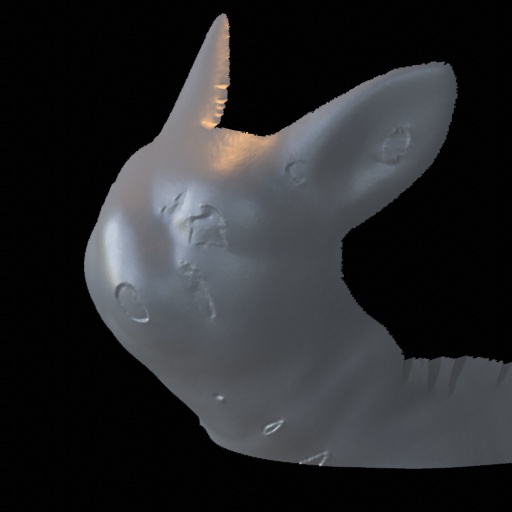}
        \end{subfigure} &
        \begin{subfigure}[t]{\cellwidth}
            \centering
            \includegraphics[height=\cellheight]{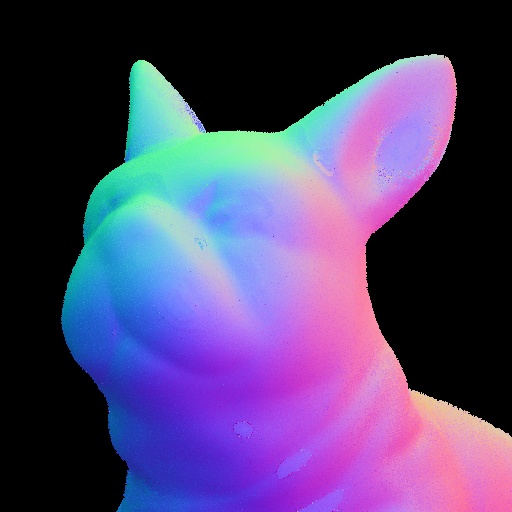}
        \end{subfigure} &
        \begin{subfigure}[t]{\cellwidth}
            \centering
            \includegraphics[height=\cellheight]{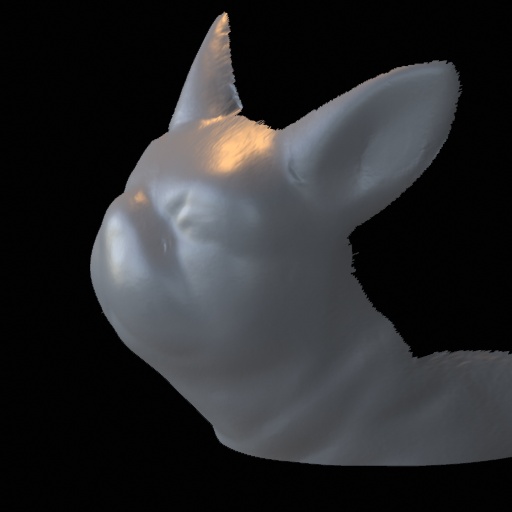}
        \end{subfigure} \\
        %%%%%%%%%%%%%%%%%%%%%%%%%%%%%%%%%%%%%%%%%%%%%%%%%%
    
        %%%%%%%%%%%%%%%%%%%%%%%%%%%%%%%%%%%%%%%%%%%%%%%%%%
        \rotatebox[origin=l]{90}{\satwhitedog} &
        \begin{subfigure}[t]{\cellwidth}
            \centering
            \includegraphics[height=\cellheight]{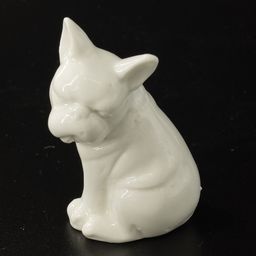}
        \end{subfigure} &
        \begin{subfigure}[t]{\cellwidth}
            \centering
            \includegraphics[height=\cellheight]{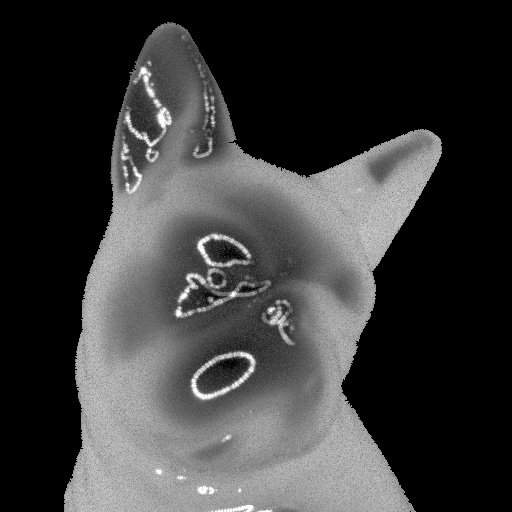}
        \end{subfigure} &
        \begin{subfigure}[t]{\cellwidth}
            \centering
            \includegraphics[height=\cellheight]{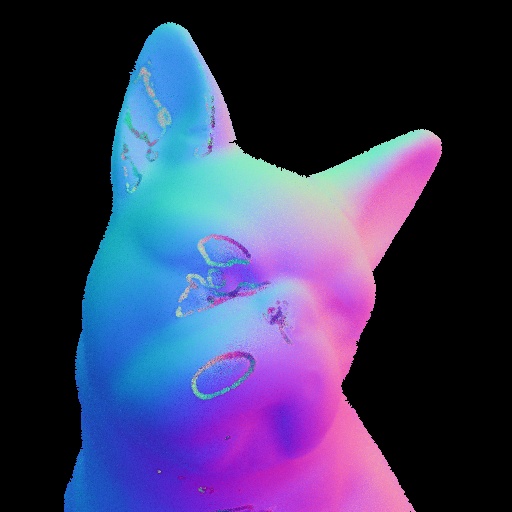}
        \end{subfigure} &
        \begin{subfigure}[t]{\cellwidth}
            \centering
            \includegraphics[height=\cellheight]{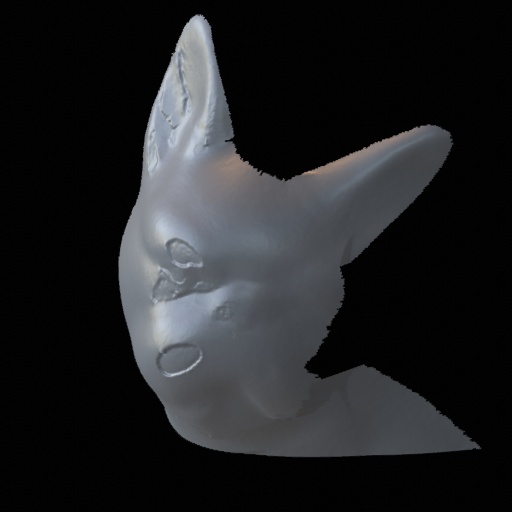}
        \end{subfigure} &
        \begin{subfigure}[t]{\cellwidth}
            \centering
            \includegraphics[height=\cellheight]{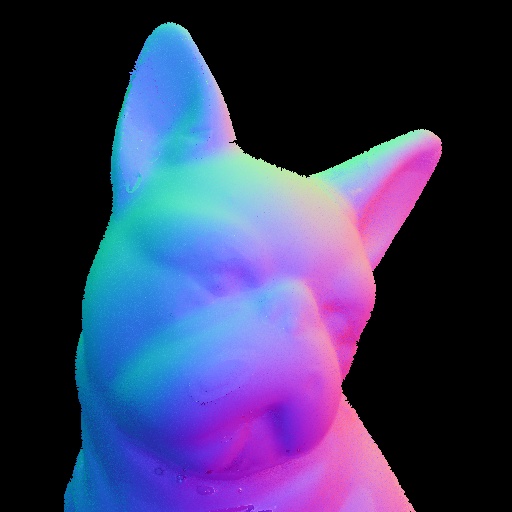}
        \end{subfigure} &
        \begin{subfigure}[t]{\cellwidth}
            \centering
            \includegraphics[height=\cellheight]{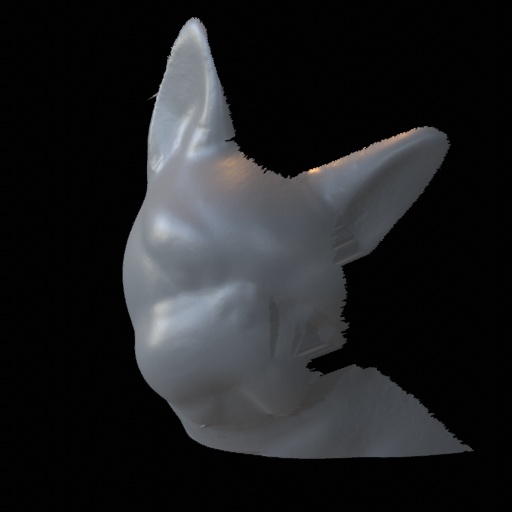}
        \end{subfigure} \\
        %%%%%%%%%%%%%%%%%%%%%%%%%%%%%%%%%%%%%%%%%%%%%%%%%%
    
        %%%%%%%%%%%%%%%%%%%%%%%%%%%%%%%%%%%%%%%%%%%%%%%%%%
        \rotatebox[origin=l]{90}{\cactusB} &
        \begin{subfigure}[t]{\cellwidth}
            \centering
            \includegraphics[height=\cellheight]{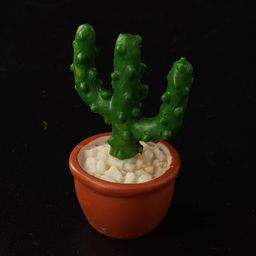}
        \end{subfigure} &
        \begin{subfigure}[t]{\cellwidth}
            \centering
            \includegraphics[height=\cellheight]{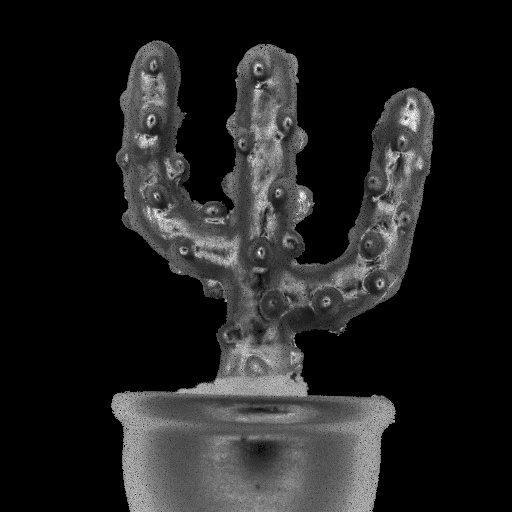}
        \end{subfigure} &
        \begin{subfigure}[t]{\cellwidth}
            \centering
            \includegraphics[height=\cellheight]{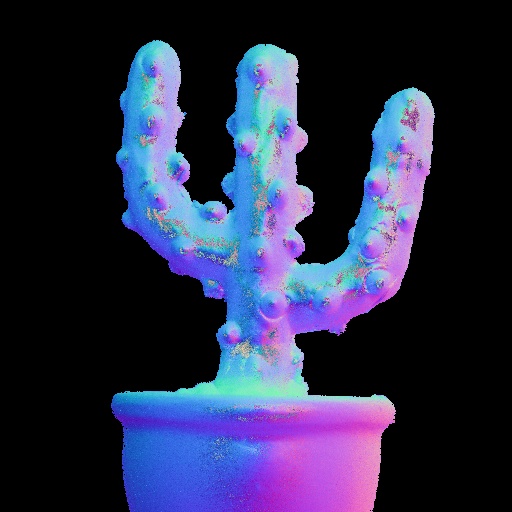}
        \end{subfigure} &
        \begin{subfigure}[t]{\cellwidth}
            \centering
            \includegraphics[height=\cellheight]{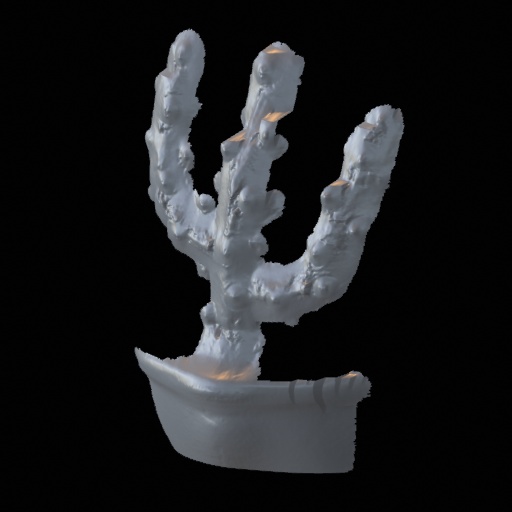}
        \end{subfigure} &
        \begin{subfigure}[t]{\cellwidth}
            \centering
            \includegraphics[height=\cellheight]{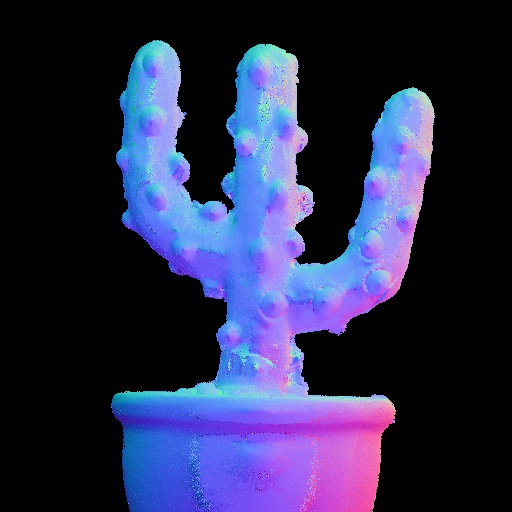}
        \end{subfigure} &
        \begin{subfigure}[t]{\cellwidth}
            \centering
            \includegraphics[height=\cellheight]{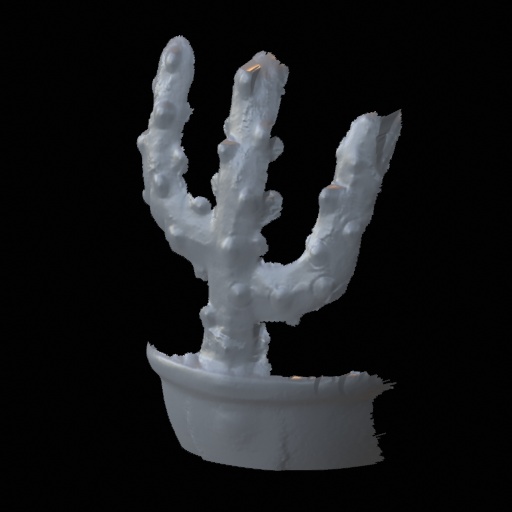}
        \end{subfigure} \\
        %%%%%%%%%%%%%%%%%%%%%%%%%%%%%%%%%%%%%%%%%%%%%%%%%%
    
        %%%%%%%%%%%%%%%%%%%%%%%%%%%%%%%%%%%%%%%%%%%%%%%%%%
        \rotatebox[origin=l]{90}{\cactusC} &
        \begin{subfigure}[t]{\cellwidth}
            \centering
            \includegraphics[height=\cellheight]{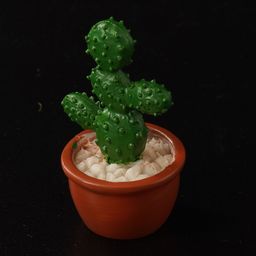}
        \end{subfigure} &
        \begin{subfigure}[t]{\cellwidth}
            \centering
            \includegraphics[height=\cellheight]{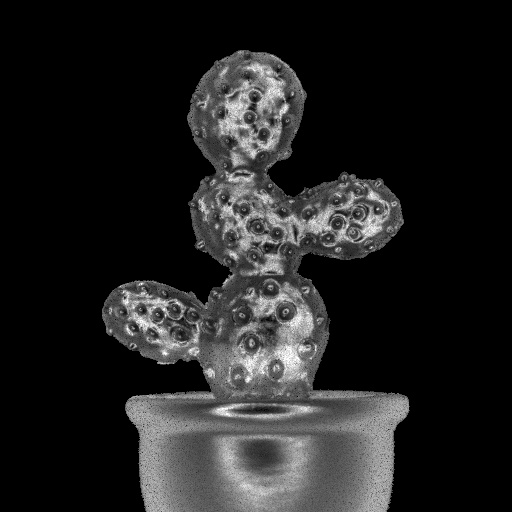}
        \end{subfigure} &
        \begin{subfigure}[t]{\cellwidth}
            \centering
            \includegraphics[height=\cellheight]{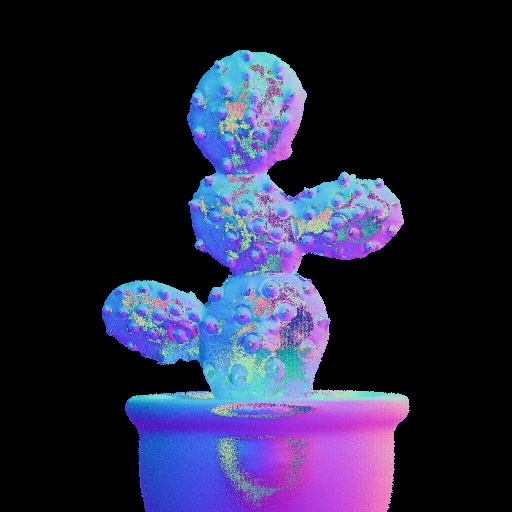}
        \end{subfigure} &
        \begin{subfigure}[t]{\cellwidth}
            \centering
            \includegraphics[height=\cellheight]{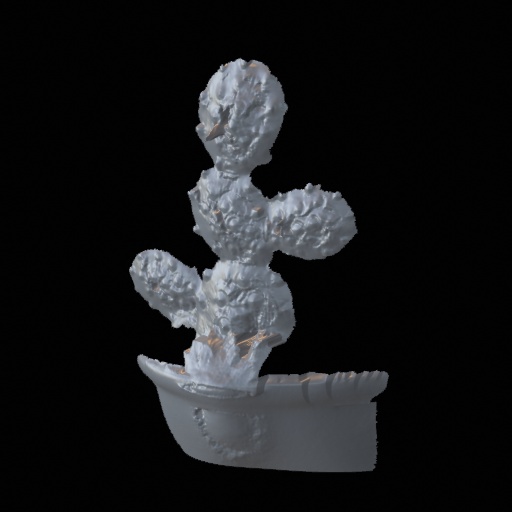}
        \end{subfigure} &
        \begin{subfigure}[t]{\cellwidth}
            \centering
            \includegraphics[height=\cellheight]{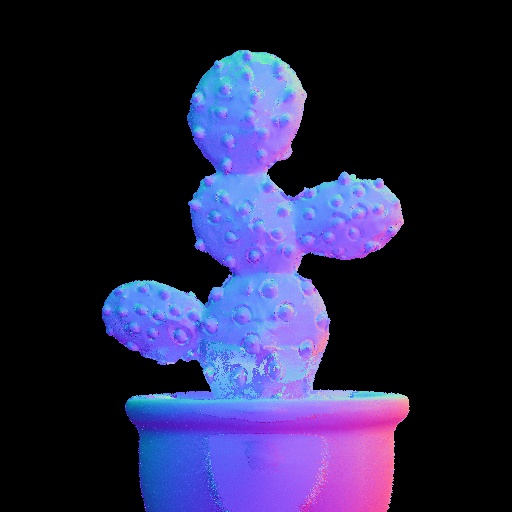}
        \end{subfigure} &
        \begin{subfigure}[t]{\cellwidth}
            \centering
            \includegraphics[height=\cellheight]{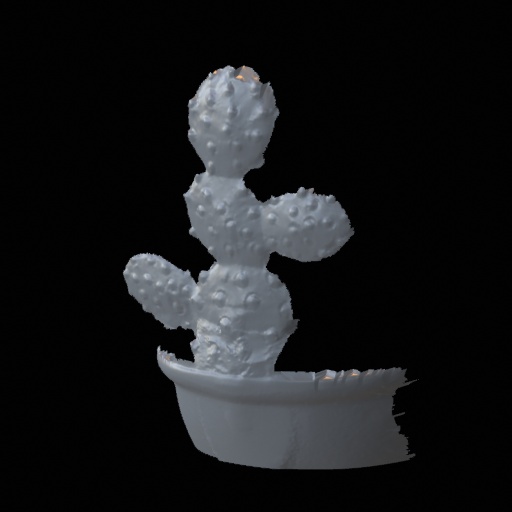}
        \end{subfigure} \\
        %%%%%%%%%%%%%%%%%%%%%%%%%%%%%%%%%%%%%%%%%%%%%%%%%%
    
        %%%%%%%%%%%%%%%%%%%%%%%%%%%%%%%%%%%%%%%%%%%%%%%%%%
        \rotatebox[origin=l]{90}{\elmo} &
        \begin{subfigure}[t]{\cellwidth}
            \centering
            \includegraphics[height=\cellheight]{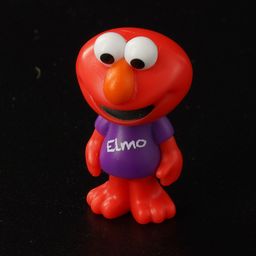}
        \end{subfigure} &
        \begin{subfigure}[t]{\cellwidth}
            \centering
            \includegraphics[height=\cellheight]{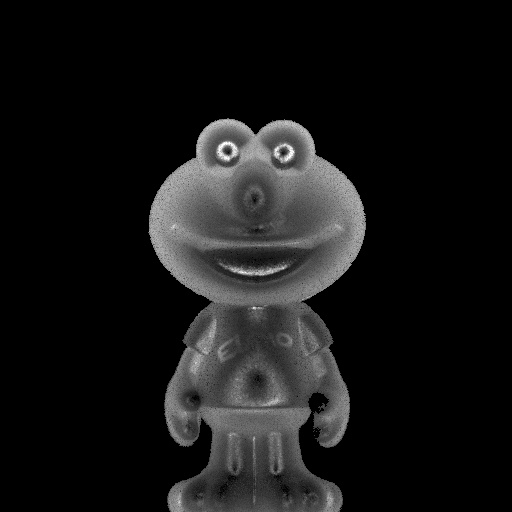}
        \end{subfigure} &
        \begin{subfigure}[t]{\cellwidth}
            \centering
            \includegraphics[height=\cellheight]{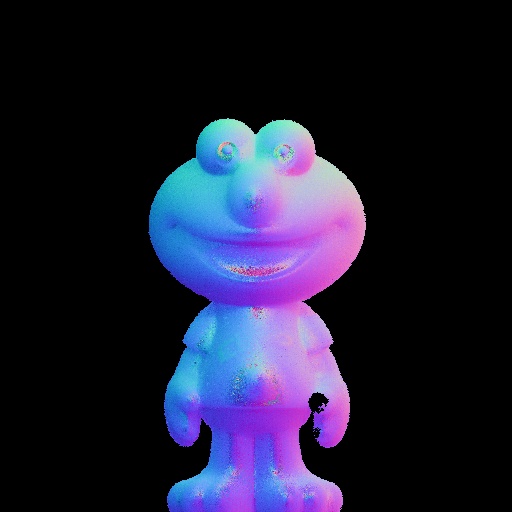}
        \end{subfigure} &
        \begin{subfigure}[t]{\cellwidth}
            \centering
            \includegraphics[height=\cellheight]{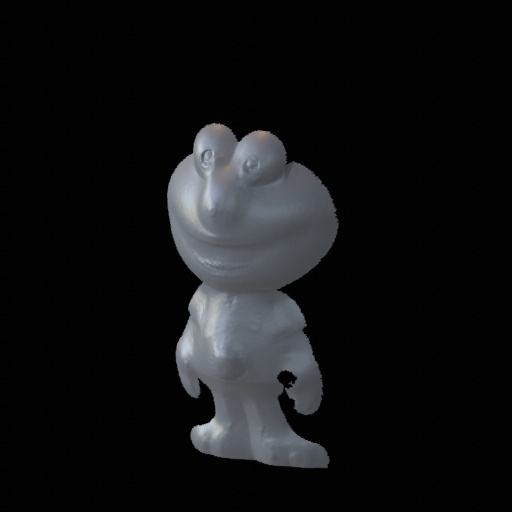}
        \end{subfigure} &
        \begin{subfigure}[t]{\cellwidth}
            \centering
            \includegraphics[height=\cellheight]{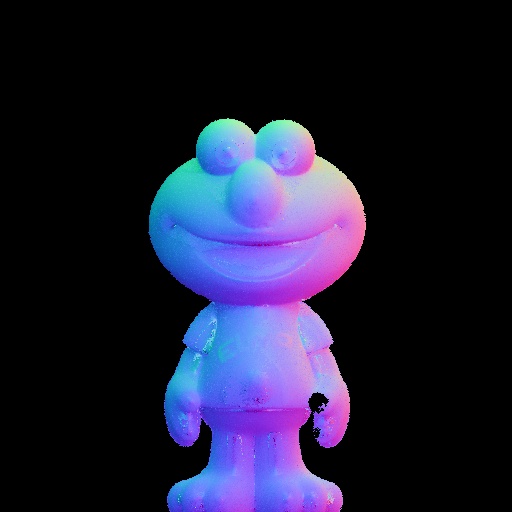}
        \end{subfigure} &
        \begin{subfigure}[t]{\cellwidth}
            \centering
            \includegraphics[height=\cellheight]{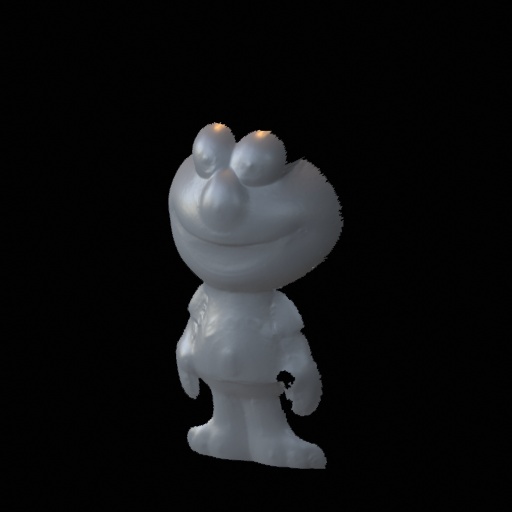}
        \end{subfigure} \\
        %%%%%%%%%%%%%%%%%%%%%%%%%%%%%%%%%%%%%%%%%%%%%%%%%%

        %%%%%%%%%%%%%%%%%%%%%%%%%%%%%%%%%%%%%%%%%%%%%%%%%%
        \rotatebox[origin=l]{90}{\bear} &
        \begin{subfigure}[t]{\cellwidth}
            \centering
            \includegraphics[height=\cellheight]{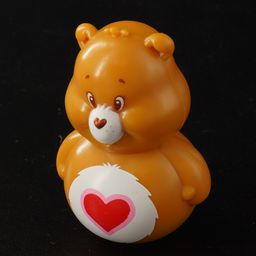}
        \end{subfigure} &
        \begin{subfigure}[t]{\cellwidth}
            \centering
            \includegraphics[height=\cellheight]{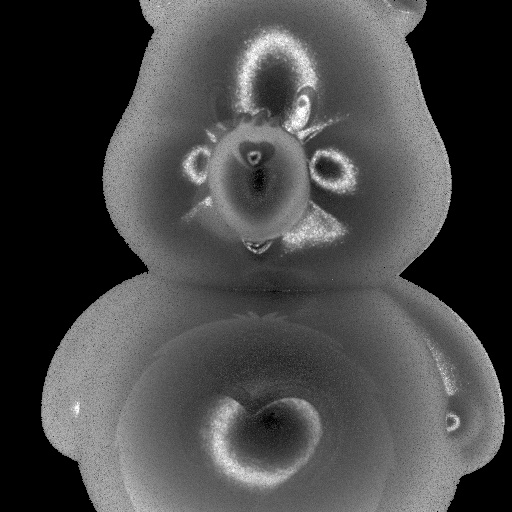}
        \end{subfigure} &
        \begin{subfigure}[t]{\cellwidth}
            \centering
            \includegraphics[height=\cellheight]{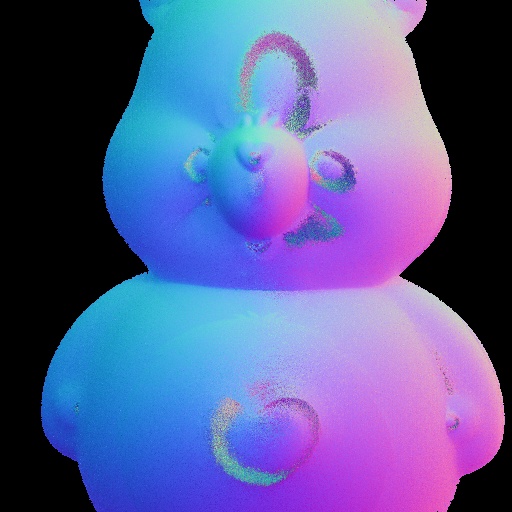}
        \end{subfigure} &
        \begin{subfigure}[t]{\cellwidth}
            \centering
            \includegraphics[height=\cellheight]{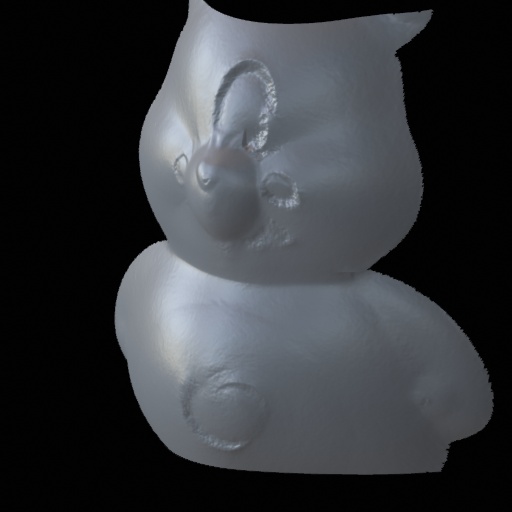}
        \end{subfigure} &
        \begin{subfigure}[t]{\cellwidth}
            \centering
            \includegraphics[height=\cellheight]{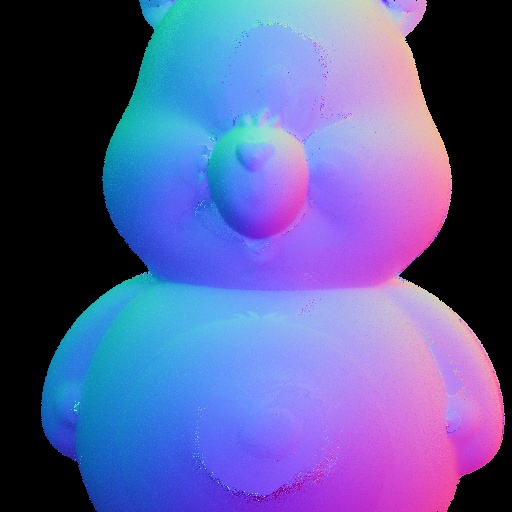}
        \end{subfigure} &
        \begin{subfigure}[t]{\cellwidth}
            \centering
            \includegraphics[height=\cellheight]{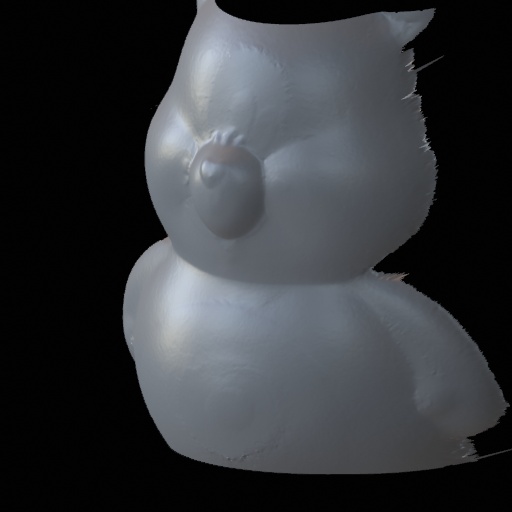}
        \end{subfigure} \\
        %%%%%%%%%%%%%%%%%%%%%%%%%%%%%%%%%%%%%%%%%%%%%%%%%%
    
        %%%%%%%%%%%%%%%%%%%%%%%%%%%%%%%%%%%%%%%%%%%%%%%%%%
        \rotatebox[origin=l]{90}{\panda} &
        \begin{subfigure}[t]{\cellwidth}
            \centering
            \includegraphics[height=\cellheight]{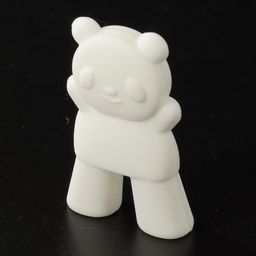}
        \end{subfigure} &
        \begin{subfigure}[t]{\cellwidth}
            \centering
            \includegraphics[height=\cellheight]{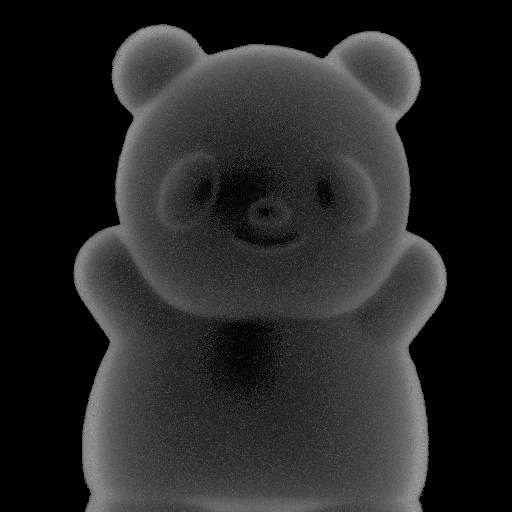}
        \end{subfigure} &
        \begin{subfigure}[t]{\cellwidth}
            \centering
            \includegraphics[height=\cellheight]{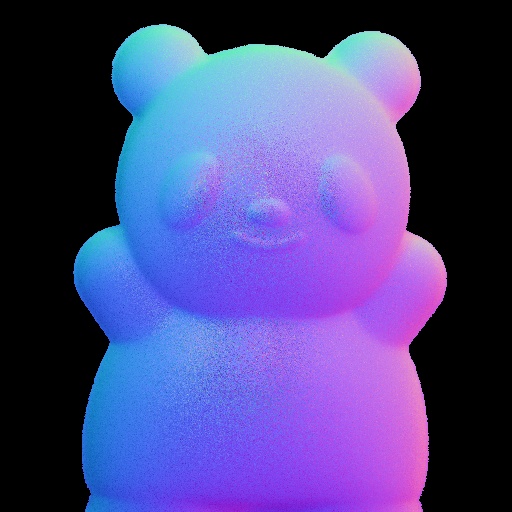}
        \end{subfigure} &
        \begin{subfigure}[t]{\cellwidth}
            \centering
            \includegraphics[height=\cellheight]{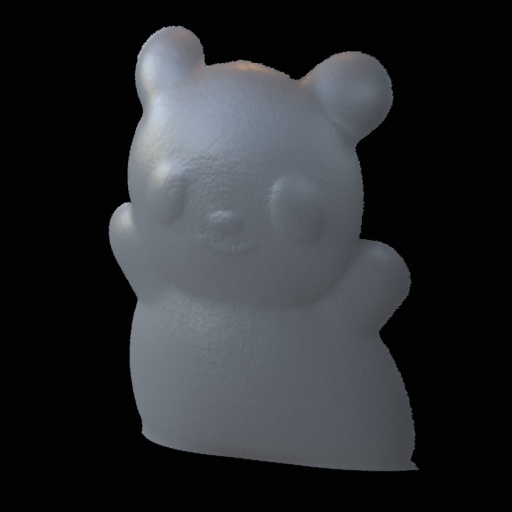}
        \end{subfigure} &
        \begin{subfigure}[t]{\cellwidth}
            \centering
            \includegraphics[height=\cellheight]{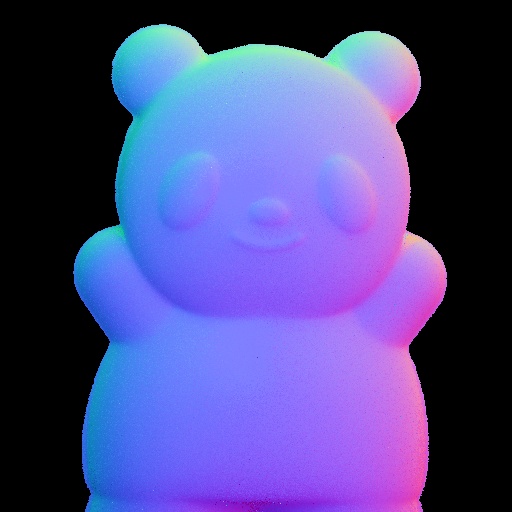}
        \end{subfigure} &
        \begin{subfigure}[t]{\cellwidth}
            \centering
            \includegraphics[height=\cellheight]{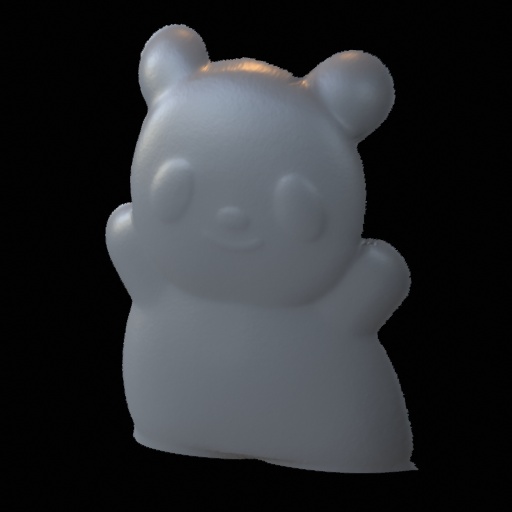}
        \end{subfigure} \\
        %%%%%%%%%%%%%%%%%%%%%%%%%%%%%%%%%%%%%%%%%%%%%%%%%%
    
        %%%%%%%%%%%%%%%%%%%%%%%%%%%%%%%%%%%%%%%%%%%%%%%%%%
        \rotatebox[origin=l]{90}{\rody} &
        \begin{subfigure}[t]{\cellwidth}
            \centering
            \includegraphics[height=\cellheight]{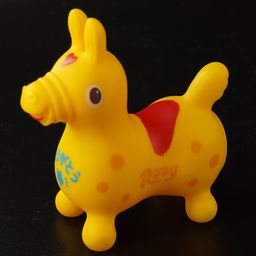}
        \end{subfigure} &
        \begin{subfigure}[t]{\cellwidth}
            \centering
            \includegraphics[height=\cellheight]{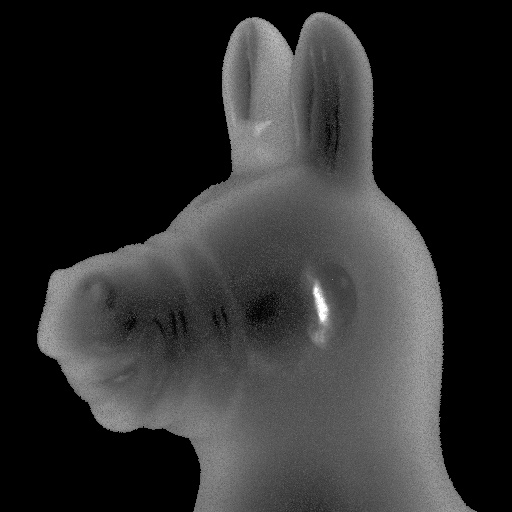}
        \end{subfigure} &
        \begin{subfigure}[t]{\cellwidth}
            \centering
            \includegraphics[height=\cellheight]{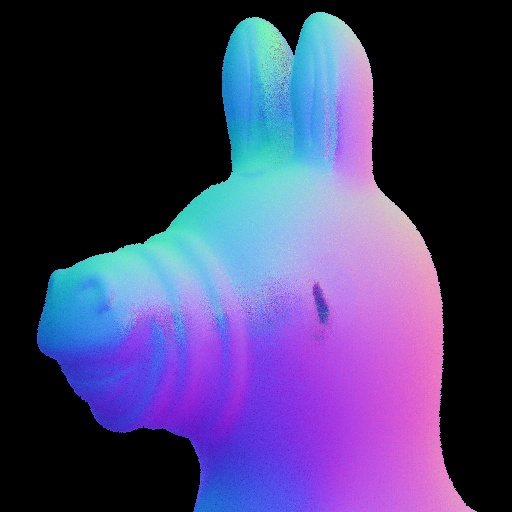}
        \end{subfigure} &
        \begin{subfigure}[t]{\cellwidth}
            \centering
            \includegraphics[height=\cellheight]{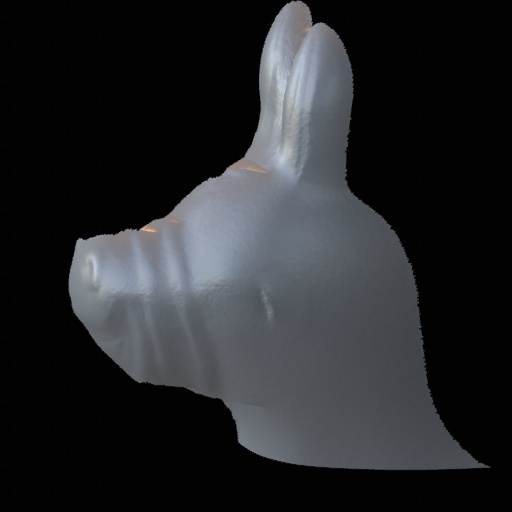}
        \end{subfigure} &
        \begin{subfigure}[t]{\cellwidth}
            \centering
            \includegraphics[height=\cellheight]{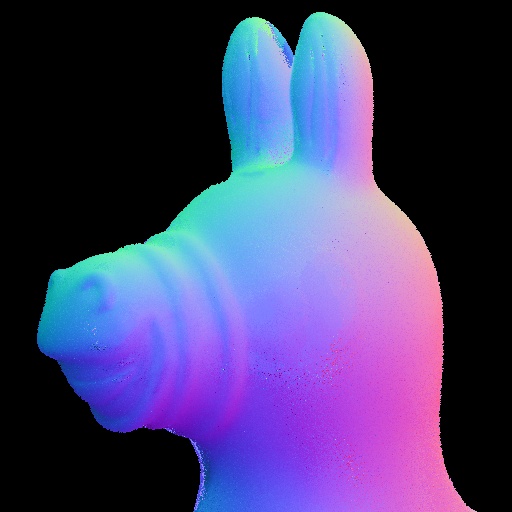}
        \end{subfigure} &
        \begin{subfigure}[t]{\cellwidth}
            \centering
            \includegraphics[height=\cellheight]{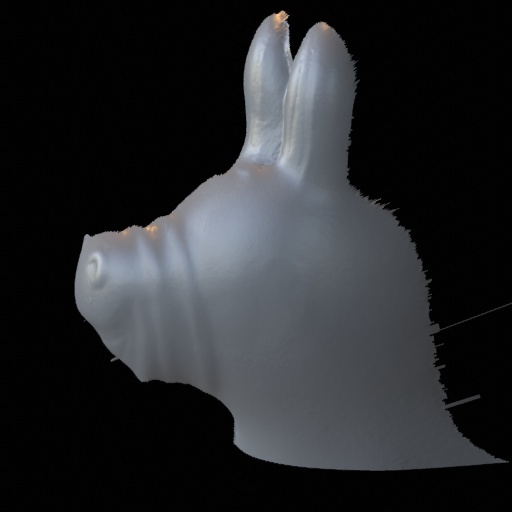}
        \end{subfigure} \\
        %%%%%%%%%%%%%%%%%%%%%%%%%%%%%%%%%%%%%%%%%%%%%%%%%%
    
        %%%%%%%%%%%%%%%%%%%%%%%%%%%%%%%%%%%%%%%%%%%%%%%%%%
        \rotatebox[origin=l]{90}{\soccerfrog} &
        \begin{subfigure}[t]{\cellwidth}
            \centering
            \includegraphics[height=\cellheight]{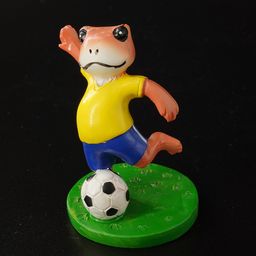}
        \end{subfigure} &
        \begin{subfigure}[t]{\cellwidth}
            \centering
            \includegraphics[height=\cellheight]{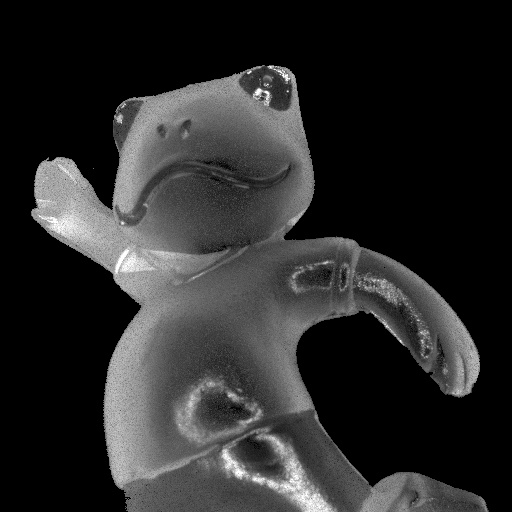}
        \end{subfigure} &
        \begin{subfigure}[t]{\cellwidth}
            \centering
            \includegraphics[height=\cellheight]{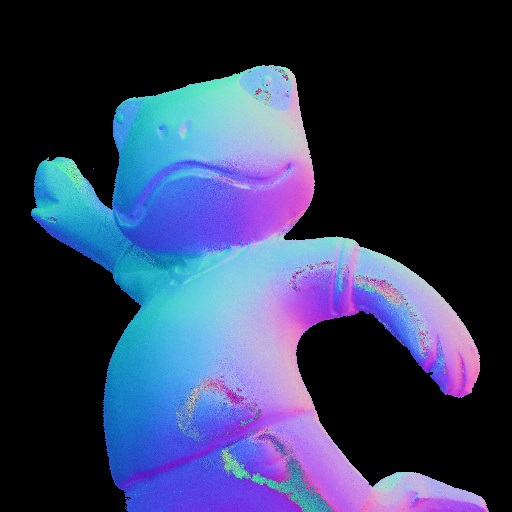}
        \end{subfigure} &
        \begin{subfigure}[t]{\cellwidth}
            \centering
            \includegraphics[height=\cellheight]{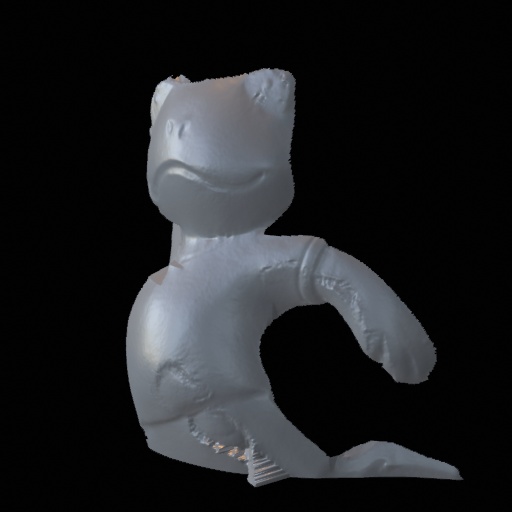}
        \end{subfigure} &
        \begin{subfigure}[t]{\cellwidth}
            \centering
            \includegraphics[height=\cellheight]{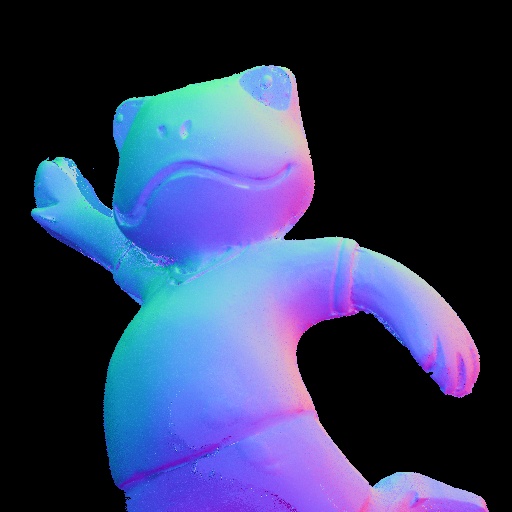}
        \end{subfigure} &
        \begin{subfigure}[t]{\cellwidth}
            \centering
            \includegraphics[height=\cellheight]{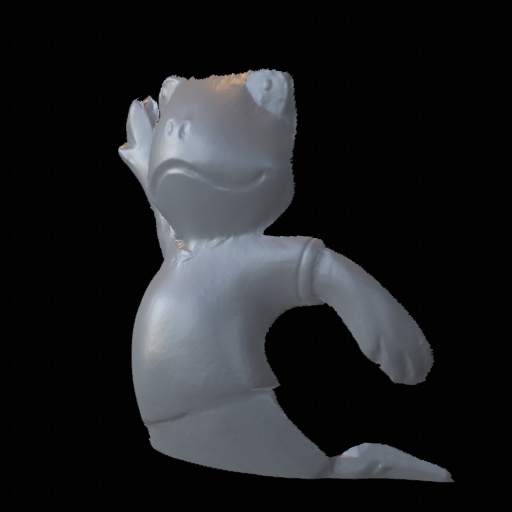}
        \end{subfigure} \\
        %%%%%%%%%%%%%%%%%%%%%%%%%%%%%%%%%%%%%%%%%%%%%%%%%%
    
        %%%%%%%%%%%%%%%%%%%%%%%%%%%%%%%%%%%%%%%%%%%%%%%%%%
        \rotatebox[origin=l]{90}{\tenniscat} &
        \begin{subfigure}[t]{\cellwidth}
            \centering
            \includegraphics[height=\cellheight]{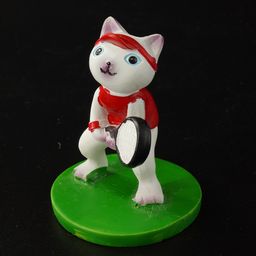}
        \end{subfigure} &
        \begin{subfigure}[t]{\cellwidth}
            \centering
            \includegraphics[height=\cellheight]{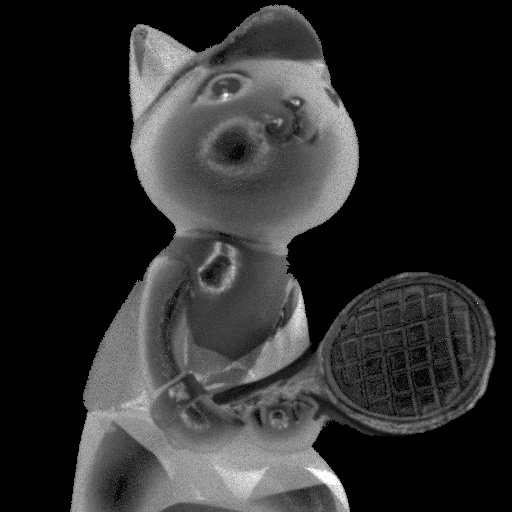}
        \end{subfigure} &
        \begin{subfigure}[t]{\cellwidth}
            \centering
            \includegraphics[height=\cellheight]{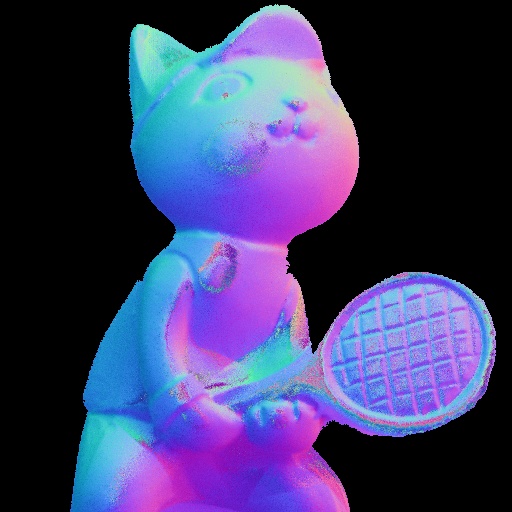}
        \end{subfigure} &
        \begin{subfigure}[t]{\cellwidth}
            \centering
            \includegraphics[height=\cellheight]{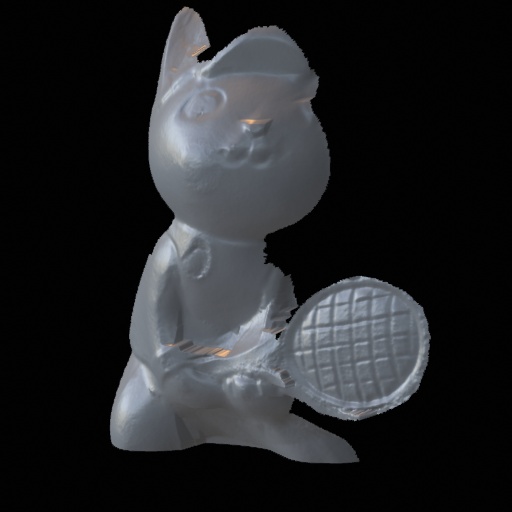}
        \end{subfigure} &
        \begin{subfigure}[t]{\cellwidth}
            \centering
            \includegraphics[height=\cellheight]{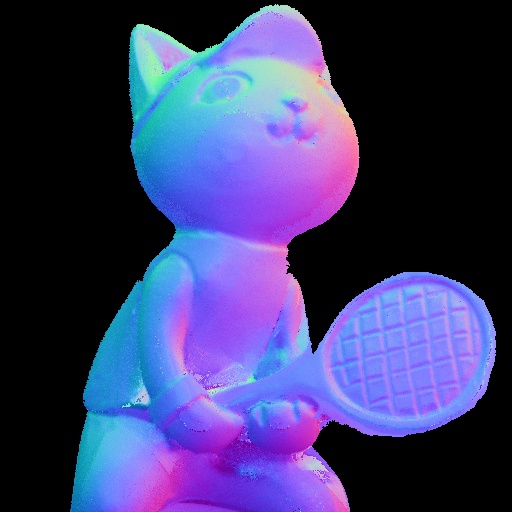}
        \end{subfigure} &
        \begin{subfigure}[t]{\cellwidth}
            \centering
            \includegraphics[height=\cellheight]{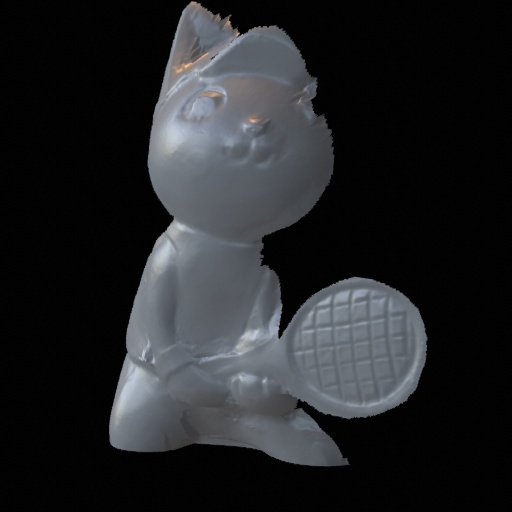}
        \end{subfigure} \\
        %%%%%%%%%%%%%%%%%%%%%%%%%%%%%%%%%%%%%%%%%%%%%%%%%%
    
        %%%%%%%%%%%%%%%%%%%%%%%%%%%%%%%%%%%%%%%%%%%%%%%%%%
        \rotatebox[origin=l]{90}{\medalfrog} &
        \begin{subfigure}[t]{\cellwidth}
            \centering
            \includegraphics[height=\cellheight]{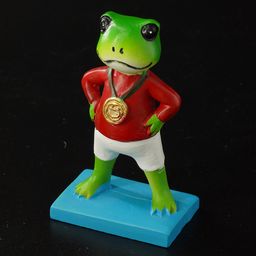}
        \end{subfigure} &
        \begin{subfigure}[t]{\cellwidth}
            \centering
            \includegraphics[height=\cellheight]{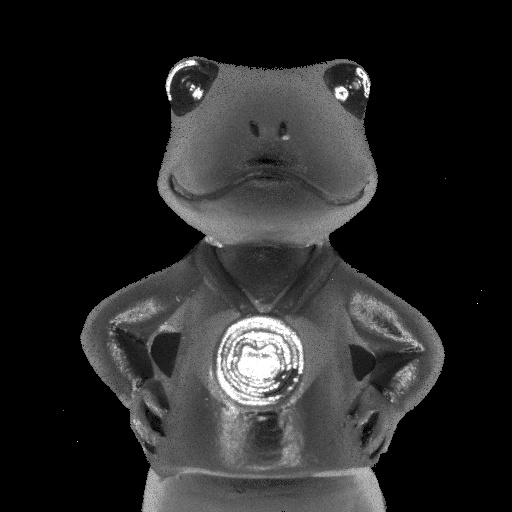}
        \end{subfigure} &
        \begin{subfigure}[t]{\cellwidth}
            \centering
            \includegraphics[height=\cellheight]{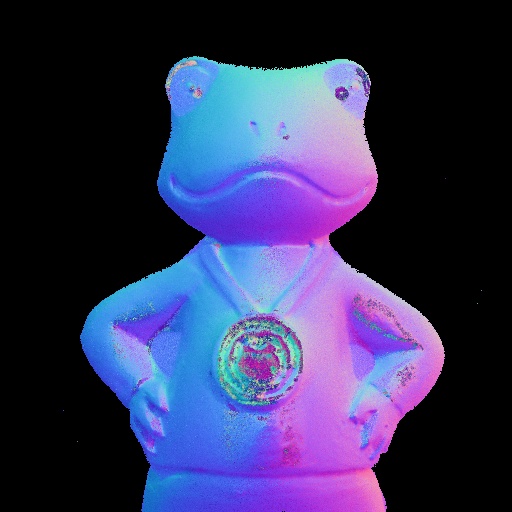}
        \end{subfigure} &
        \begin{subfigure}[t]{\cellwidth}
            \centering
            \includegraphics[height=\cellheight]{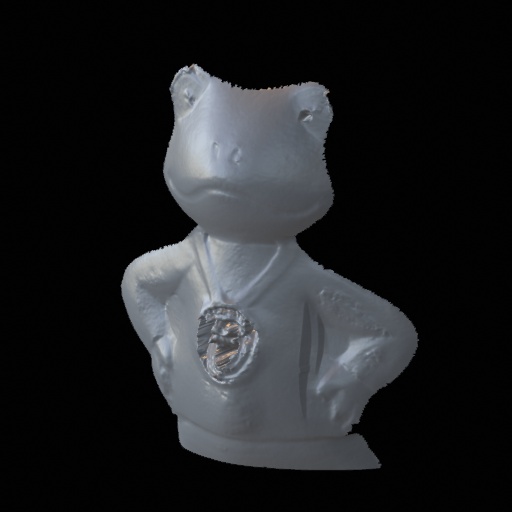}
        \end{subfigure} &
        \begin{subfigure}[t]{\cellwidth}
            \centering
            \includegraphics[height=\cellheight]{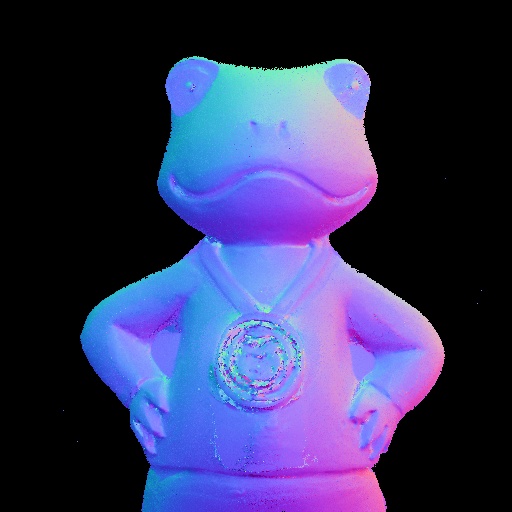}
        \end{subfigure} &
        \begin{subfigure}[t]{\cellwidth}
            \centering
            \includegraphics[height=\cellheight]{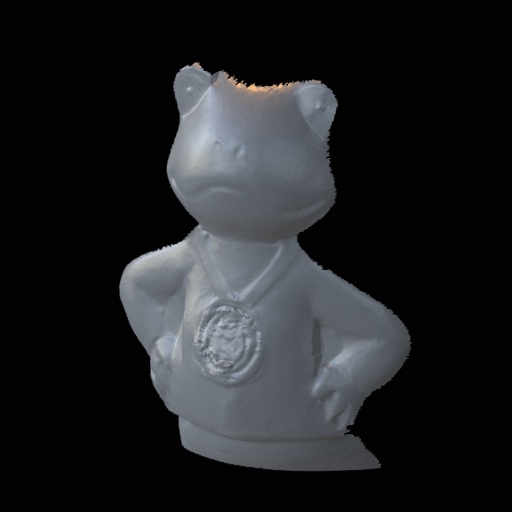}
        \end{subfigure} \\
        %%%%%%%%%%%%%%%%%%%%%%%%%%%%%%%%%%%%%%%%%%%%%%%%%%

    \end{tabular*}
    
    \caption{
    Additional, qualitative evaluation with the practical objects. From left to right: photograph, event accumulation image from events, normal map recovered by EventPS-FCN~\cite{EventPS} and its depth~\cite{bini2022cao}, and those by ours.}
    \label{fig:additional eval practical}
\end{figure}

Finally, we evaluate the proposed method on $12$ more practical objects are targeted for qualitative assessment. The experimental results are shown in \cref{fig:additional eval practical}. 
The fine surface structures of objects such as \owl, \cactusB, and \cactusC are successfully recovered. Most of the practical objects exhibit specularity on their surface. For instance, \whitedog and \satwhitedog are pottery objects with strong specularity, yet their normal maps appear well-recovered. \bear also has specularity with a wider lobe. It seems that the texture colors do not significantly affect the normal map, but the impact of specularity remains noticeable. This discrepancy might be due to the mask margin not aligning well with this particular case, suggesting a need for adjustment tailored to the target. Additionally, \panda, being a rubber object, exhibits subsurface scattering, presenting a challenging non-Lambertian effect that warrants consideration in future work.

Since the ground truth shapes of these practical objects are unknown, we conduct a visual evaluation of depth structures reconstructed from the normal maps. Therefore, in this paper, we employ a state-of-the-art method for reconstructing depth from a normal map, as proposed by Cao~\textit{et}~\textit{al.}~\cite{bini2022cao}. The code is accessible on GitHub~\footnote{https://github.com/xucao-42/bilateral\_normal\_integration}. Upon inputting the recovered normal map, its spatial foreground mask image, and the camera intrinsics, a depth structure is generated as polygons (PLY file). The depth images in \cref{fig:additional eval practical} are rendered from these polygons with a MatCap (ceramic dark) in Blender~\cite{Blender}.

\subsection{Extension to non-Lambertian BRDF}
Our PS-EIP assumes an event generation model based on Lambertian diffuse reflection. However, the definition of EIP itself is more general, allowing for extensions to models other than Lambertian diffuse reflection (e.g., microfacet BRDF such as Cook-Torrance). Even if a non-Lambertian model is employed, the optimization process for EIP can still be achieved using the same methodology. Although we have not yet explored this, we plan to examine it in future work.

\end{document}